\documentclass[11pt,a4paper]{article}
\usepackage[dvipsnames]{xcolor}
\usepackage[hyperref]{emnlp2020}
\usepackage{times}
\usepackage{latexsym}
\usepackage{amssymb}
\usepackage{amsmath}
\usepackage{romannum}
\usepackage{graphicx}
\usepackage{booktabs}
\usepackage{subcaption}
\usepackage{multirow}
\usepackage{adjustbox}
\usepackage{float}
\usepackage{xargs} 
\usepackage{colortbl}
\usepackage[linesnumbered,ruled]{algorithm2e}
\usepackage{IEEEtrantools}
\usepackage{soul}
\usepackage{mathabx}
\usepackage{tikz}

\usepackage{microtype}

\aclfinalcopy 

\newcommand{\F}{\ensuremath{\text{F}_1}}
\renewcommand{\vec}[1]{\mathbf{#1}}

\definecolor{tableGreen}{rgb}{0.32,0.71,0.54}
\definecolor{tableRed}{rgb}{0.84,0.33,0.30}
\definecolor{lightgray}{gray}{0.9}

\newcommand{\ffbox}[1]{%
  {%
  \setlength{\fboxrule}{0pt}%
   \setlength{\fboxsep}{-2\fboxrule}%
   \fbox{\hspace{1pt}\strut#1\hspace{1pt}}%
  }%
}
\newcommand*\circled[1]{\tikz[baseline=(char.base)]{
            \node[shape=circle,draw,align=center, inner sep=0pt, outer sep=0pt] (char) {\ffbox{#1}};}}

\title{{F}1 is {N}ot {E}nough! {M}odels and {E}valuation\\
{T}owards {U}ser-{C}entered {E}xplainable {Q}uestion {A}nswering}

\author{Hendrik Schuff$^{1,2}$ \hspace*{1.5cm} Heike Adel$^1$\hspace*{1.5cm} Ngoc Thang Vu$^2$\\
    $^1$Bosch Center for Artificial Intelligence, Renningen, Germany\\
    $^2$Institut für Maschinelle Sprachverarbeitung, University of Stuttgart\\
    \texttt{\{Hendrik.Schuff,Heike.Adel\}@de.bosch.com}\\
    \texttt{Thang.Vu@ims.uni-stuttgart.de}
}

\date{}

\begin{document}
\pagenumbering{arabic}

\maketitle

\begin{abstract}
Explainable question answering systems predict an answer together with an explanation showing why the answer has been selected.
The goal is to enable users to assess the correctness of the system and understand its reasoning process.
However, we show that current models and evaluation settings have shortcomings regarding the coupling of answer and explanation which might cause serious issues in user experience.
As a remedy, we propose a hierarchical model and a new regularization term to strengthen the answer-explanation coupling as well as two 
evaluation scores to quantify the coupling. 
We conduct experiments on the \textsc{HotpotQA} benchmark data set and perform a user study.
The user study shows that our models increase the ability of the users to judge the correctness of the system and that 
scores like \F\ are not enough to estimate the usefulness of a model in a practical setting with human users.
Our scores are better aligned with user experience, making them promising candidates for model selection.

\end{abstract}

\section{Introduction}\label{sec:introduction}
Understanding the decisions of deep learning models is of utmost importance, especially when they are deployed in critical domains, such as medicine or finance \cite{ribeiro2016should}.
In natural language processing (NLP), a variety of tasks have been addressed regarding explainability of neural networks, such as textual entailment \citep{camburu2018snli}, sentiment classification \citep{clos2017towards}, machine translation \cite{stahlberg2018operation} and question answering \citep{yang2018hotpotqa}.
In this paper, we address question answering (QA) due to its proximity to users in real-life settings, for instance, in the context of personal assistants.

\begin{figure}[t]
    \centering
    \includegraphics[width=.85\columnwidth]{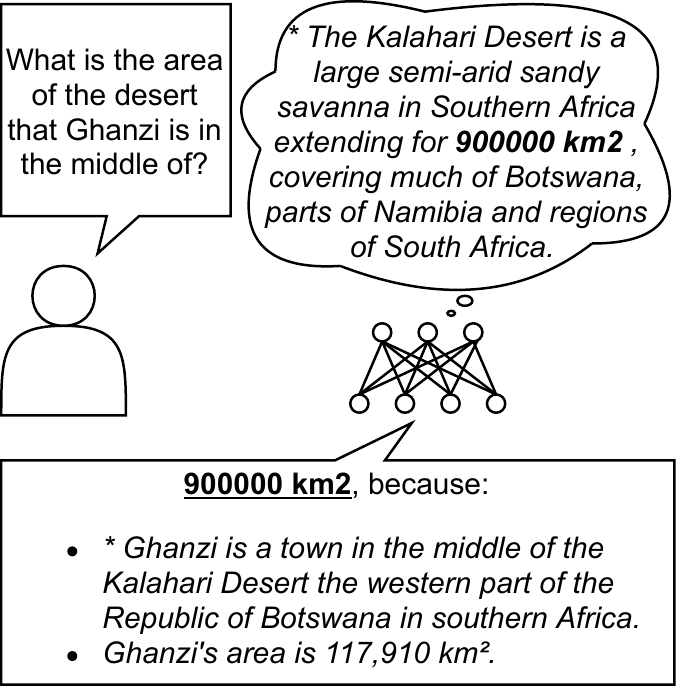}
    \caption{Example output of a representative XQA system \cite{yang2018hotpotqa} that would receive an answer-\F\ of 1 and an explanation-\F\ of 0.5 although the explanation provides no value to the user since the actual answer evidence (shown in cloud) is not included in the explanation (asterisks mark ground truth explanation).}
    \label{fig:motivating_example}
\end{figure}

Explainable question answering (XQA) is the task of (\romannum{1}) answering a question and (\romannum{2}) providing an explanation that enables the user to understand \textit{why} the answer was selected, 
e.g., by pointing to the facts that are needed for answering the question.
Compared to approaches that output importance weights or analyze gradients \cite{Simonyan14a,ribeiro2016should,lundberg2017unified,DBLP:conf/icml/SundararajanTY17}, this has the advantage that the explanations are intuitively assessible even by lay users without machine learning background.

A \emph{good explanation} (i.e., one that is helpful for the user) should therefore satisfy the following requirements:
(i) It should contain all information that the model used to predict the answer for the question. This is necessary so that the user can reconstruct the model's reasoning process.
(ii) It should not include additional information that it did not use for predicting the answer. Otherwise, the explanation will confuse the users rather than help them.
Note that these requirements do not only hold for correct model decisions but are also valid for explaining wrong model answers so that the user can assess the correctness of the answers.
Previous work on XQA mostly focuses on developing models that predict the correct answer and, independent of this, the correct explanation \cite{yang2018hotpotqa,qi2019answering,shao2020graph}. This can lead to model outputs in which the explanations do not sufficiently relate to the answers. Consider the example provided in Figure \ref{fig:motivating_example}.
The model gives the correct answer to the question and provides an explanation consisting of one out of two relevant facts. However, the most important relevant fact (in which the answer actually appears) is not part of the explanation. As a result, the user cannot assess whether the model answer is correct or not and, thus, cannot trust the system.
To strengthen the \emph{coupling of answer and explanation prediction} in the model architecture and during training, we propose two novel approaches in this paper: 
(i) a hierarchical neural network architecture for XQA that ensures that only information included in the explanation is used to predict the answer to the question, and
(ii) a regularization term for the loss function that explicitly couples answer and explanation prediction during training.

A \emph{good evaluation measure} should score explanations by satisfying the following requirements:
(i) It should reward explanations that are coupled to the answers of the model.
(ii) It should punish explanations that are unrelated to the answers of the model.
(iii) It should be correlated to user experience.
Since explanations cannot only empower the user to assess the correctness of a system \cite{biran2017human, kim2016examples}
but also improve user satisfaction and confidence \cite{sinha2002role, biran2017human} and, thus, increase the acceptance of automatic systems \cite{herlocker2000explaining, cramer2008effects}, this aspect
is very important when evaluating models that should be applied in real-life scenarios. 
In most recent works, evaluation of XQA models focuses on optimizing \F-scores of answers and explanations (a collection of so-called supporting or relevant facts) \cite{yang2018hotpotqa}.
However, \F-scores only assess model outputs with respect to ground-truth annotations which only contain explanations for the correct answer.
Thus, they fail to quantify the coherence between answer and explanation, especially when the predicted model answer is wrong.
The example model output in Figure \ref{fig:motivating_example} leads to an answer-\F-score of 1 and an explanation-\F-score of 0.5 although the explanation is useless for the user as described before.
To \emph{quantify the model's answer-explanation coupling}, we propose two novel evaluation scores:
(i) \textsc{FaRM} which tracks prediction changes when removing facts,
and (ii) \textsc{LocA} which assesses whether the answer is contained in the explanation or not.
Both scores do not require ground-truth annotations.

To summarize, we make contributions in two directions in this paper:
For \textit{modeling},
(i) we propose a hierarchical neural network architecture as well as
(ii) a regularization term for the loss function of XQA systems.
For \textit{evaluation},
(iii) we propose two scores that are able to quantify a model's answer-explanation coupling without relying on ground-truth annotations.
(iv) To investigate the relation between different evaluation scores and user experience, we conduct a \textit{user study}.
The results show that our proposed models increase the ability of the user to judge the correctness of an answer and that our scores are stronger predictors of human behavior than standard scores like \F.
(v) For reproducibility and future research, we will release code for our methods and for computing the evaluation scores as well as the user study data.\footnote{\url{https://github.com/boschresearch/f1-is-not-enough}}

\section{Related Work}\label{sec:related_work}
In the context of XQA, \citet{yang2018hotpotqa} present the \textsc{HotpotQA} data set which we also use for the experiments in this paper. In addition to questions and answers, it contains explanations in the form of relevant sentences from Wikipedia articles.

Most state-of-the-art models for \textsc{HotpotQA} extend the BiDaf++ architecture \cite{clark2017simple,seo2016bidirectional}, e.g., \cite{yang2018hotpotqa,qi2019answering,nishida2019answering,DBLP:journals/corr/abs-1911-02170, qiu-etal-2019-dynamically, shao2020graph}.
Other approaches are based on question decomposition \cite{min-etal-2019-multi,DBLP:journals/corr/abs-2002-09758}, 
graph/hierarchical structures \cite{DBLP:journals/corr/abs-1911-00484,DBLP:journals/corr/abs-1911-03631,DBLP:conf/iclr/AsaiHHSX20},
virtual knowledge bases \cite{DBLP:conf/iclr/DhingraZBNSC20} or transformer models \cite{DBLP:conf/iclr/ZhaoXRSBT20}.

So far, all of the research work on \textsc{HotpotQA} focuses on reaching higher \F-scores. In contrast, we question whether this actually aligns with user experience. 
To the best of our knowledge, only \citet{chen2018multi} additionally conduct a human evaluation.
This confirms the observation of \citet{adadi2018peeking} that only
very few
papers related to explainable AI
address (human) evaluation of explainability.
Despite the large body of research in the field of human computer interaction, \citet{abdul2018trends} show that there is a lack of collaboration and transfer of results to machine learning communities.

Another line of research our work relates to is the criticism of automatic evaluation scores.
One frequently questioned score is BLEU \cite{papineni-etal-2002-bleu}, which was shown to only correlate weakly with human judgements in tasks like machine translation \cite{callison-burch-etal-2006-evaluating}, storytelling \cite{wang-etal-2018-metrics} and dialogue response generation \cite{DBLP:conf/emnlp/LiuLSNCP16}. 
\F\ has been criticized from various perspectives including theoretical considerations and concrete applications \cite{DBLP:journals/sac/HandC18, chicco2020advantages, DBLP:conf/ausai/SokolovaJS06}.
\citet{qian-etal-2016-new} show that modifying \F-scores based on insights from psychometrics improves their correlation with human evaluations.
In this paper we criticize the usage of \F\ as a measure of explainability in XQA and show in a user study that it is not related to user experience.

\section{Methods for XQA}\label{sec:methods}
We built upon the model by \newcite{qi2019answering} as it is an improved version of the BiDaf++ model, which is used in numerous state-of-the-art XQA models \cite{yang2018hotpotqa,qi2019answering,nishida2019answering,DBLP:journals/corr/abs-1911-02170, qiu-etal-2019-dynamically} including the best-scoring publication \cite{shao2020graph}. 
It consists of a question and context encoding part with self-attention, followed by two prediction heads: a prediction of relevant facts (i.e., the explanation) and a prediction of the answer to the question. The two heads are trained in a multi-task fashion based on the sum of their respective losses.
First, we analyze the outputs of the model, revealing severe weaknesses in answer-explanation coupling.
To address those weaknesses, we then propose (\romannum{1}) a novel neural network architecture that \textit{selects} and \textit{forgets} facts, and (\romannum{2}) a novel answer-explanation coupling regularization term for the loss function.

\subsection{Limitations of Current Models}\label{sec:weaknesses}
We manually analyze outputs of the models by \newcite{qi2019answering} and \newcite{yang2018hotpotqa} and identify the following two problems.

\paragraph{Silent Facts.}
The models make use of facts without including them into their explanations (cf., Figure \ref{fig:motivating_example}).
As a result, the predicted answer does not occur in the explanation, leaving the user uninformed about where it came from.

\paragraph{Unused Facts.}
The models predict facts to be relevant without any relation to the predicted answer.
The second fact of the explanation in Figure \ref{fig:motivating_example} is an example for this. We also found examples where the facts predicted to be relevant do not even contain the entities from the question.

\subsection{Select \& Forget Architecture}
\begin{figure}[t]
    \centering
        \includegraphics[width=.9\columnwidth]{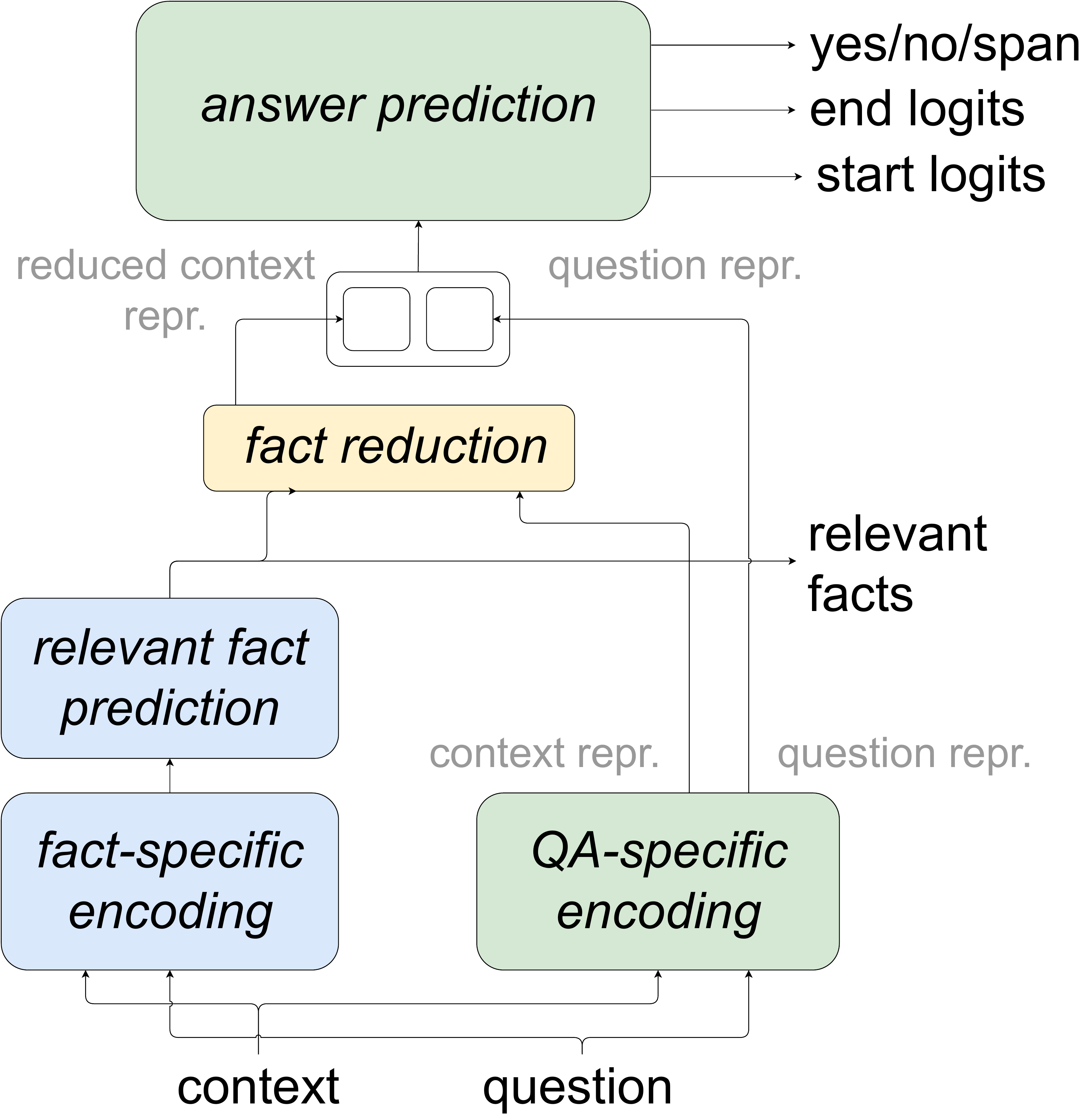}
    \caption{``Select and Forget'' architecture with task-specific encodings and fact reduction.}\label{fig:HA_arch}
\end{figure}

To explicitly ensure that the model only uses information from facts it predicts to be relevant for the answer selection, we propose a hierarchical model that first \textit{selects} facts which are relevant to answer the question and then \textit{forgets} about all other facts (see Figure~\ref{fig:HA_arch}).
We use recurrent and self-attention layers to create encodings of the question and the context.
In particular, we create two different encodings: one that will be used for predicting the relevance of the facts (fact-specific encoding)
and one that will be used for predicting the answer to the question (QA-specific encoding).
Based on the fact-specific encoding, the model first predicts which facts are relevant to answer the question.
Next, we reduce the QA-specific encoding of the context based on the relevance predictions.
In particular, we mask all facts that were not predicted to be relevant by zeroing out their encodings.
The reduced context representation is concatenated with the QA-specific question encoding and passed on to the answer prediction, which we implement in the same way as \citet{qi2019answering}.
Thus, the answer prediction now only receives encodings of facts that the model has predicted to be relevant.
It predicts the type of the answer (yes/no/text span), as well as the start and end positions of the answer span within the context.

\subsection{Answer-Fact Coupling Regularizer}\label{sec:regularizer}
Our second method addresses the coupling of answer and explanation prediction by modifying the loss function.
The loss function used by \newcite{qi2019answering} is the sum of four cross entropy losses concerning (\romannum{1}) the answer type (yes/no/span) distribution, (\romannum{2}) the answer start token distribution, (\romannum{3}) the answer end token distribution, and (\romannum{4}) the fact relevance distributions.
All terms are optimized to be close to their respective ground truth annotations.
This is the desired effect in many, but, as our examples in Section \ref{sec:weaknesses} show, not in all situations.
The loss function especially encourages the model to predict the ground truth explanation rather than an explanation that explains the predicted answer.

In order to reward a coupling between answer and explanation, we propose to add the following regularization term to the loss function:
\begin{IEEEeqnarray}{lCl}
  J_{\text{reg}} &=& \underbrace{p_{\text{a}} \cdot (\overbrace{p_{\text{e}} \cdot 0}^{\text{GT expl.}} + \overbrace{(1-p_{\text{e}}) \cdot c_1}^{\text{non-GT expl.}})}_{\text{correct answer}} \nonumber \\
  && + \underbrace{(1-p_{\text{a}}) \cdot (\overbrace{p_{\text{e}} \cdot c_2}^{\text{GT expl.}} + \overbrace{(1-p_{\text{e}}) \cdot c_3}^{\text{non-GT expl.}})}_{\text{wrong answer}}\label{eq:regularizer}
\end{IEEEeqnarray}
with $p_{\text{a}}$ corresponding to the probability of the model for the correct answer span and $p_{\text{e}}$ denoting the probability of the model for the ground truth relevant facts.
The term can be broken down into four cases: (\romannum{1}) correct answer and ground truth explanation, (\romannum{2}) correct answer but non-ground truth explanation, (\romannum{3}) incorrect answer but ground truth explanation and (\romannum{4}) incorrect answer and non-ground truth explanation.
Each case corresponds to a constant cost of $0$, $c_1$, $c_2$ and $c_3$, respectively, with $c_1, c_2, c_3$ being hyperparameters.
The resulting cost $J_{\text{reg}}$ is the sum of the four individual costs weighted with their respective probabilities.

In particular, $p_{\text{a}}$ is defined as the product of the probabilities assigned to start and end token positions of the answer span.
For a data set instance with a context containing $N$ facts, we define $\vec{s_t}\in \{0,1\}^N$ as the ground truth annotations for the relevant facts.
Accordingly, we denote the model's relevance probability estimates with $\vec{s_p}\in [0,1]^N$.
Based on this, we define $p_{\text{e}}$ as
$		p_{\text{e}}^{*} = \prod_{i \in F} \vec{s_p}_i$
with $F = \{i \in \{1, ..., N\}: \vec{s_t}_i=1\}$ denoting the indices that correspond to ground truth facts.
This corresponds to the joint probability of selecting the ground truth facts assuming the single selection probabilities to be independent.
For our experiments, we adapt this definition to a numerically more stable term $p_{\text{e}}^{+}$ by replacing the product with a sum as this led to slightly better results on the development set.

\section{New Evaluation Scores for XQA}\label{sec:metrics}
In this section, we motivate that standard scores like \F\ are not enough to score XQA systems by presenting their limitations in those settings.
To be able to quantify the degree to which a model is affected by those limitations, we propose two scores that go beyond standard scores:
the \textit{fact-removal score} and the \textit{answer-location score}.
Both scores can be calculated without any assumptions on model architecture and no need for ground truth annotations for answers or supporting facts.

\subsection{Limitations of Current Evaluation Scores}
Current evaluation of XQA is focused on three scores: (\romannum{1}) answer-\F, which is based on the token overlap between the predicted and the ground truth answer, (\romannum{2}) SP-\F, which calculates \F\ based on the overlap of predicted and ground truth relevant (``supporting'') facts and (\romannum{3}) joint-\F, which is based on the definitions of joint precision and joint recall as the products of answer and SP precision and recall as described in \citet{yang2018hotpotqa}.
For \textsc{HotpotQA}, models are ranked based on joint-\F.
We argue that this creates a false incentive that potentially hinders the development of truly usable models for the following reasons.

\paragraph{No Empirical Evidence.}
There is no empirical evidence that joint-\F\ is related to user performance or experience regarding XQA.
\paragraph{Rewarding Poor Explanations.}
Figure~\ref{fig:motivating_example} shows an example prediction that is rewarded with a joint-\F\ of 0.5 although its explanation provides no value to the user.
The reward stems from the overlap of the explanation with the ground truth but does not consider that the predicted answer is not contained in any of the predicted relevant facts.

\paragraph{Punishing Good Explanations.}
Consider a model output in which the predicted answer is wrong but the explanation perfectly explains this wrong answer, showing to the user why the model has selected it. Standard \F-scores compare the model output to the ground truth annotations and will, therefore, score both the answer and the explanation with an \F\ of 0. 
However, we argue that an explanation should be evaluated with a score higher than 0 if it is able to explain the reasoning process of the model to the user and, thus, lets the user identify the failure of the model.

\subsection{Fact-Removal Score (\textsc{FaRM})}
Ideally, the explanations of the model include all facts that the model uses within its reasoning chain
but no additional facts beyond that.
Note that  
even for a wrong model answer, this assumption should hold so that the relevant facts provide explanations for the (wrongly) predicted answer.
To quantify the degree of answer-explanation coupling, we propose to iteratively remove parts (individual facts) of the explanation, re-evaluate the model using the reduced context and track how many of the model's answers change.
For a model with perfect coupling of answer and explanation, the answer will change with the first fact being removed (assuming no redundancy) but will not change when removing irrelevant facts not belonging to the explanation.
We remove facts in order of decreasing predicted relevance as more relevant facts should influence the model's reasoning process the strongest.

In the following, we denote an instance of the data set by $e \in E$ with its corresponding question $e_{\text{ques}}$ and context $e_{\text{con}}$.
We use $\mathrm{answer}(\cdot, \cdot)$ to denote the answer that a model predicts for a given question and context.
The functions $\mathrm{reduce}_{\text{rel}}(\cdot, k)$ ($\mathrm{reduce}_{\text{irr}} (\cdot, k)$) return a context from which up to $k$ facts the model predicts to be relevant (irrelevant) have been removed.\footnote{If the number of facts predicted as (ir)relevant is less or equal to $k$, we remove all (ir)relevant facts from the context.}
We re-evaluate the model on this reduced context and calculate the fraction of changed answers $c_{\text{rel}}(k)$ and $c_{\text{irr}}(k)$, respectively.
\begin{IEEEeqnarray}{lCl}
  a(e) &=& \mathrm{answer}(e_{\text{ques}}, e_{\text{con}})\\
  \hat{a}_{\text{rel},k}(e) &=& \mathrm{answer}(e_{\text{ques}}, \mathrm{reduce}_{\text{rel}}(e_{\text{con}}, k))\\
  \hat{a}_{\text{irr},k}(e) &=& \mathrm{answer}(e_{\text{ques}}, \mathrm{reduce}_{\text{irr}}(e_{\text{con}}, k))\\
  c_{\text{rel}}(k) &=& \frac{|\{e \in E: a(e) \neq \hat{a}_{\text{rel},k}(e)\}|}{|E|}\\
  c_{\text{irr}}(k) &=& \frac{|\{e \in E: a(e) \neq \hat{a}_{\text{irr},k}(e)\}|}{|E|}
\label{eq:farm}
\end{IEEEeqnarray}
Finally, we condense $c_{\text{rel}}(k)$ and $c_{\text{irr}}(k)$ into a single fact-removal score:
\begin{IEEEeqnarray}{lCr}
  \textsc{FaRM}(k) &=& \frac{c_{\text{rel}}(k)}{1 + c_{\text{irr}}(k)} \in [0,1]
\label{eq:FaRM}
\end{IEEEeqnarray}

$\textsc{FaRM}(k)$ ranges between zero and one and a higher score corresponds to a better explanation.

\subsection{Answer-Location Score (\textsc{LocA})}
A second important indicator for the degree of a model's answer-explanation coupling is the location of the answer span:
As shown in Figure~\ref{fig:motivating_example}, the models can predict answers that are located outside the facts they predict to be relevant, i.e., outside the explanation. This is confusing for a user.
Therefore, we consider the fractions of answer spans that are inside the explanation of the model and the fraction of answer spans that are outside.
For an ideal model, all answer spans would be located inside the explanation.
We use $I$ and $O$ to denote the number of answers inside/outside of the set of facts predicted as relevant.
$A$ denotes the total number of answers.\footnote{In \textsc{HotpotQA}, answers can stem from article titles although titles are never used as relevant facts. Thus, $A > I + O$ is possible. Our score is still applicable in this case.}
Based on these counts, we propose the answer-location score that we define as
\begin{IEEEeqnarray}{lCl}
  \textsc{LocA} &=& \frac{\frac{I}{A}}{1 + \frac{O}{A}} =  \frac{I}{A + O} \in [0,1]. 
\label{eq:combined_metric_location}
\end{IEEEeqnarray}
The \textsc{LocA} score ranges between zero and one, with larger values indicating better answer-explanation coupling.

\section{Experiments and Results}\label{sec:experiments}
In this section, we describe the dataset we used in our experiments as well as our results. 
More details for reproducibility, including hyperparameters, are provided in the appendix.

\subsection{Dataset}
The \textsc{HotpotQA} data set is a multi-hop open-domain explainable question answering data set containing 113k questions with crowd-sourced annotations.
Each instance of the training data contains a question, a context consisting of the first paragraph of ten Wikipedia articles, the annotated answer and an explanation in the form of a selection of relevant sentences from the context.
As \textsc{HotpotQA} was designed as a multi-hop data set, finding the answer to a question requires combining information from two different articles. The eight other articles are \textit{distracting} the system.\footnote{The data set also contains a \textit{full wiki} test set in which the context spans all collected Wikipedia articles. We focus on the distractor setting in this paper. The data set can be downloaded from \url{https://hotpotqa.github.io/}.}

\begin{table}
\centering
\resizebox{\columnwidth}{!}{%
\begin{tabular}{p{0.1cm}l|c|rrrr}
\toprule
& Score & Qi-2019 & S\&F & reg. & S\&F+reg. \\ \midrule
\multirow{12}{*}{\rotatebox[origin=c]{90}{Standard Scores}} & Answer-EM& 49.48& 46.09& \textbf{49.67}& 46.44\\
 & Answer-\F& \textbf{63.76}& 59.99& 63.56& 60.60\\
 & Answer-P & 66.26& 62.41& \textbf{66.27}& 62.77\\
 & Answer-R & \textbf{65.52}& 61.61& 65.06& 62.53\\
 & SP-EM& 39.81& \textbf{42.16}& 25.98& 30.45\\
 & SP-\F& 79.34& \textbf{80.07}& 75.60& 77.53\\
 & SP-P & 78.01& \textbf{78.84}& 66.79& 70.09\\
 & SP-R & 85.26& 85.45& \textbf{93.26}& 92.03\\
 & Joint-EM& 22.28& \textbf{22.78}& 14.56& 16.38\\
 & Joint-\F& \textbf{52.51}& 50.71& 49.66& 48.99\\
 & Joint-P & \textbf{53.33}& 51.40& 45.64& 45.74\\
 & Joint-R & 57.92& 55.61& \textbf{62.09}& 59.52\\
 \midrule
 \multirow{6}{*}{\rotatebox[origin=c]{90}{Proposed Scores}} & $\textsc{FaRM}(4)$& 66.20& 75.54& 73.32& \textbf{76.64}\\
 & $\drsh$ $c_{\text{rel}}(4)$& 77.06& \textbf{86.05} & 81.69& 84.58\\
 & $\drsh$ $c_{\text{irrel}}(4)$& 16.39& 13.91& 11.41& \textbf{10.36}\\
 & \textsc{LocA}& 60.49& 70.60& 67.92& \textbf{75.56}\\
 & $\drsh$ I & 67.48& 71.68& 72.60& \textbf{76.32}\\
 & $\drsh$ O & 11.55& 1.53& 6.89& \textbf{1.01}\\
 \bottomrule

\end{tabular}%
}
\caption{Comparison of our methods to \newcite{qi2019answering} regarding evaluation scores from related work and our proposed scores on the distractor dev set (SP: supporting facts).
All values in \%.}\label{tab:comp_solutions_bl}
\end{table}

\subsection{Experimental Results}
In our experiments, we assess the effects of our Select \& Forget architecture (S\&F) and the regularization term (reg.).
Table~\ref{tab:comp_solutions_bl} shows our approaches in comparison to the model by \newcite{qi2019answering}.\footnote{We 
retrain their model using the implementation and preprocessing provided at \url{https://github.com/qipeng/golden-retriever}.}
While our S\&F architecture performs comparable in standard scores like answer-exact-match (Answer-EM), answer-\F, joint-EM and joint-\F\ (for some of them slightly better, for some of them slightly lower), the regularization term increases the recall of the relevant fact prediction considerably.

\begin{figure*}
    \centering
    \begin{subfigure}[t]{.25\textwidth}
        \centering
    \includegraphics[width=\textwidth]{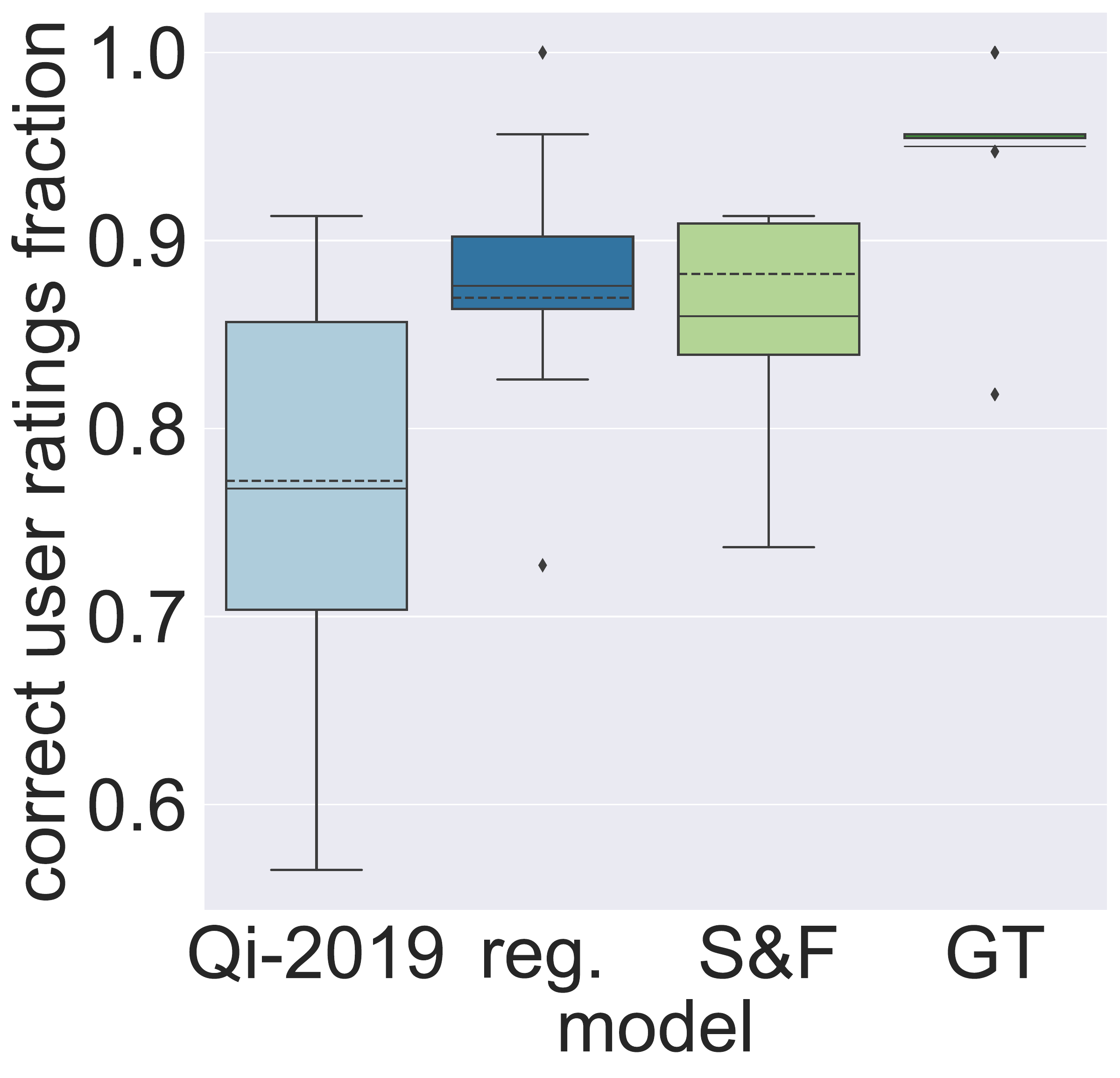}
    \caption{Correct user ratings.}\label{fig:correctness_means_box}
    \end{subfigure}%
    \begin{subfigure}[t]{.25\textwidth}
    \centering
    \includegraphics[width=\textwidth]{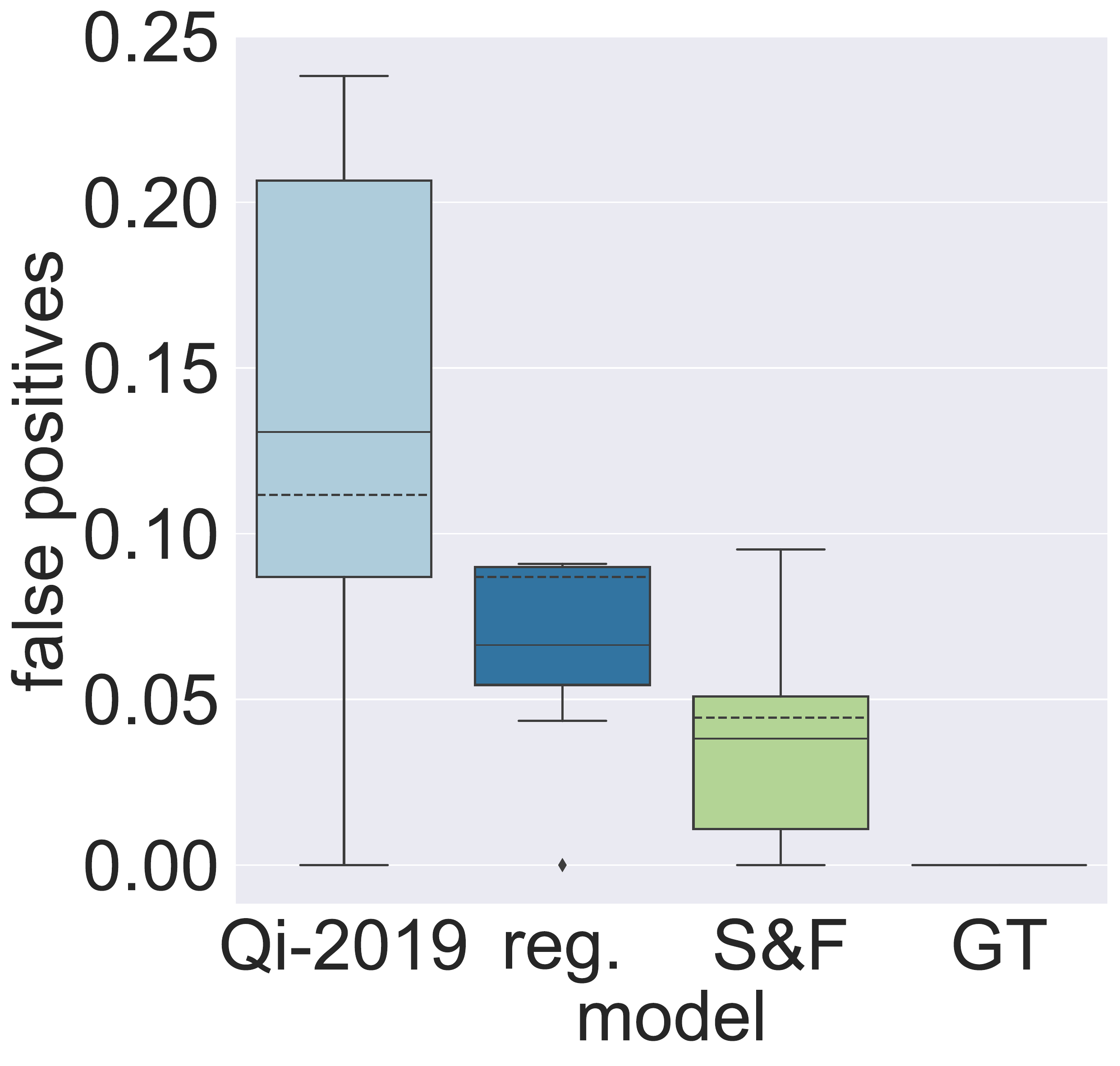}
    \caption{False positive ratio.}\label{fig:FP}
    \end{subfigure}%
    \begin{subfigure}[t]{.25\textwidth}
    \centering
    \includegraphics[width=\textwidth]{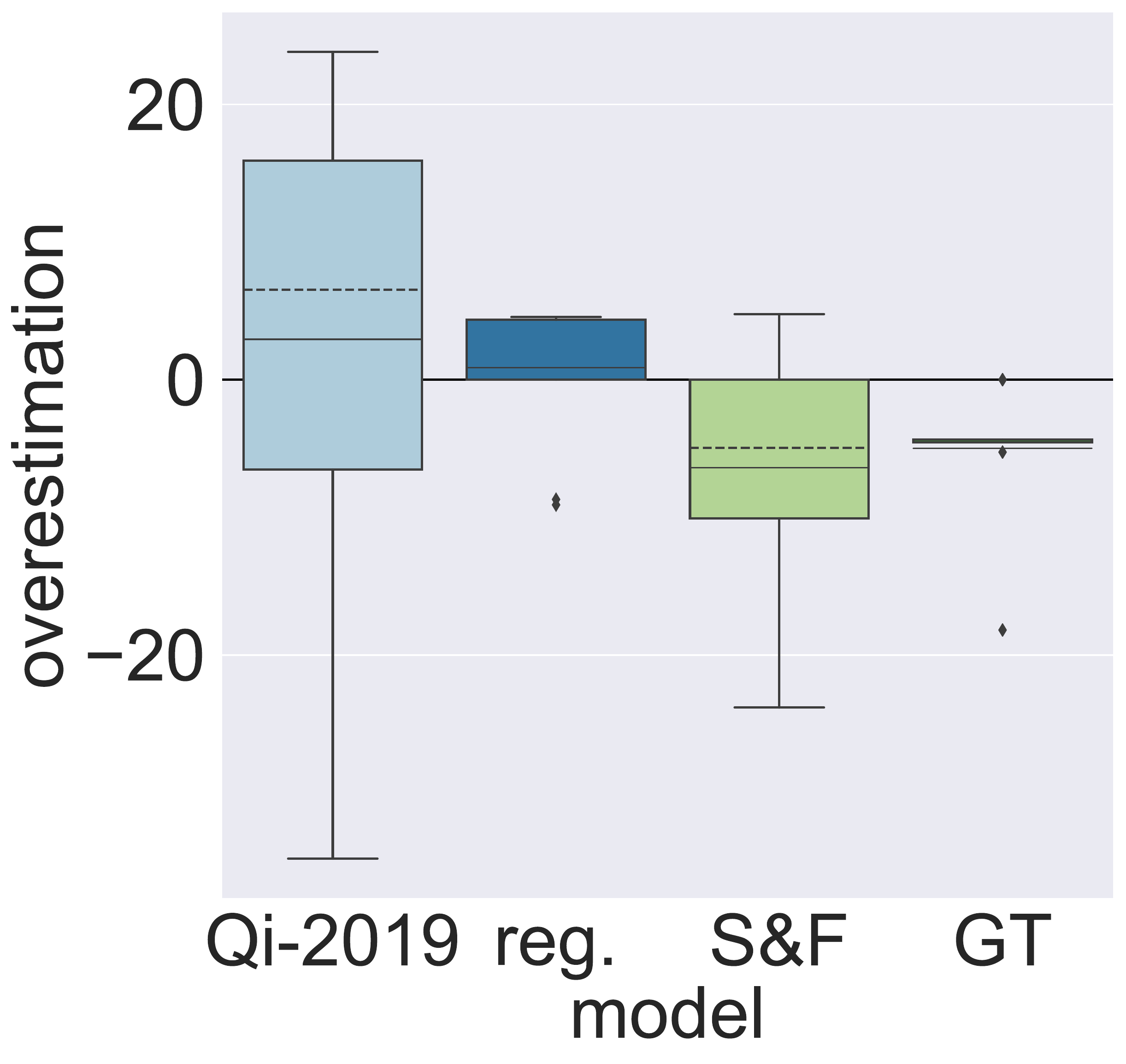}
    \caption{Overestimation in percent.}\label{fig:overestimation}
    \end{subfigure}%
    \begin{subfigure}[t]{.25\textwidth}
    \centering
    \includegraphics[width=\textwidth]{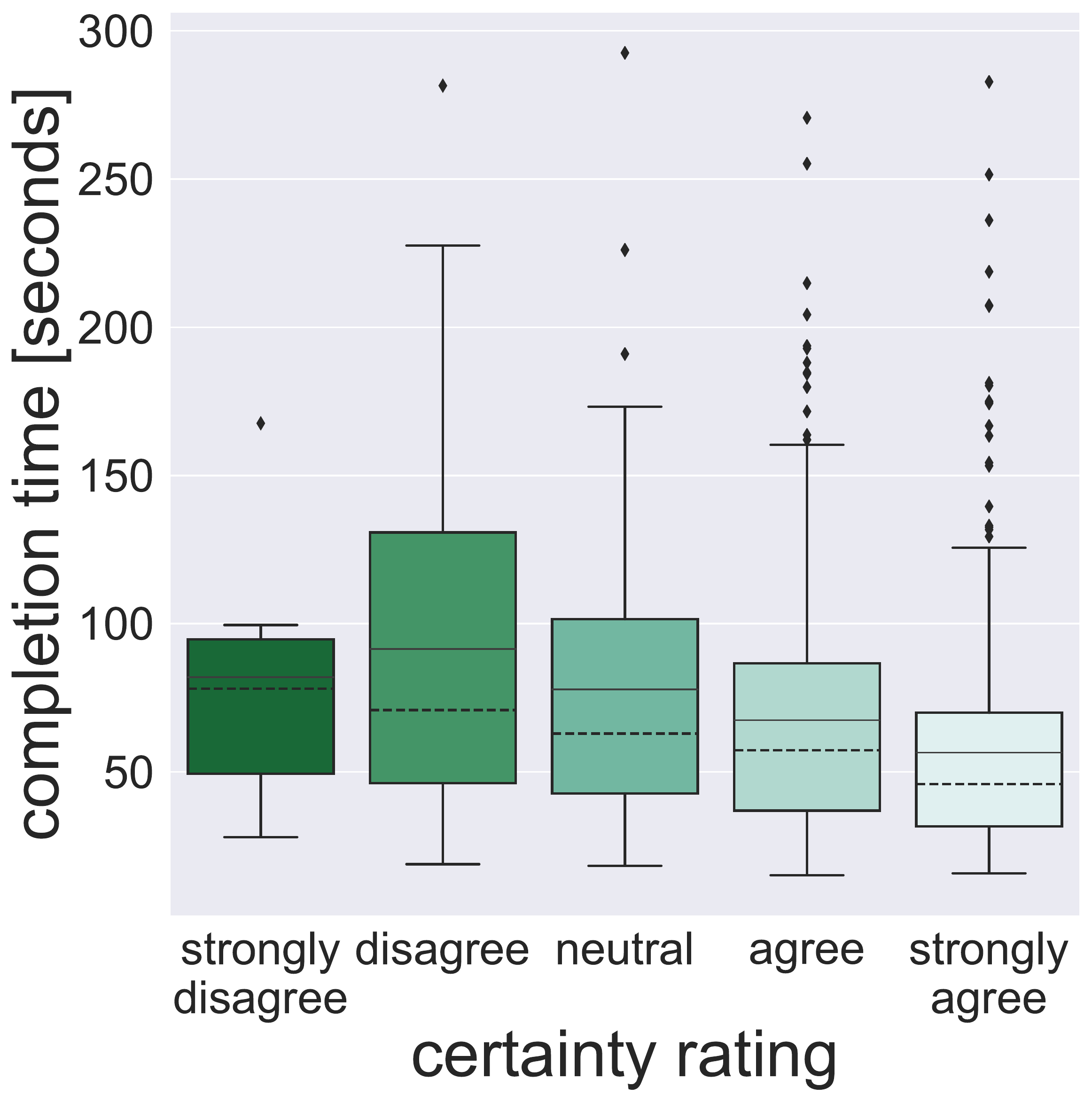}
        \caption{Completion time$\sim$certainty.}\label{fig:ct_of_certainty}
    \end{subfigure}%
    \caption{Boxplots showing results from the user study. Boxes mark quartiles, whiskers mark 1.5 inter-quartile ranges, outliers are plotted separately. Horizontal solid/dashed lines within boxes mark means and medians, resp.}
\end{figure*}

In terms of our proposed scores for measuring answer-explanation-coupling, all our three models clearly outperform the baseline model (lower part of Table~\ref{tab:comp_solutions_bl}).
In the first three rows, we report the models' \textsc{FaRM} scores and the fractions of changed answers for $k=4$, i.e., when a maximum of four facts are removed.
We choose $k=4$ as this is the highest number of facts within an explanation in the ground truth annotations of the \textsc{HotpotQA} data.
The last three rows show the \textsc{LocA} scores and the respective fractions of answers inside and outside facts predicted as relevant.

The different behavior of models regarding joint-\F\ vs.\ \textsc{FaRM} and \textsc{LocA} raises the question which scores are better suited to quantify explainability in a real-life setting with human users. 
To answer this, we conduct a user study in Section \ref{sec:human_eval}.

\section{Human Evaluation}\label{sec:human_eval}
We conduct a user study to investigate whether standard scores like \F\ or our proposed scores are better suited to predict user behavior and performance. Moreover, the study provides another way to compare our proposed methods to the model by \newcite{qi2019answering} and the ground truth explanations.
In contrast to the human evaluation from \citet{chen2018multi}, we evaluate explanations in the context of the model answer, ask participants to rate the predictions along multiple dimensions and collect responses from 40 instead of 3 subjects.

\subsection{Choice of Models}
We choose to compare the model proposed by \citet{qi2019answering} (called ``Qi-2019'' in the following) as a representative of the commonly used BiDaf++ architecture \cite{clark2017simple} in XQA, our proposed Select \& Forget architecture (S\&F) and our proposed regularization term (reg.).
In addition, we include the ground truth (GT) annotations to set an upper bound.
Although the combination of regularization and the S\&F architecture reaches promising performance in Table \ref{tab:comp_solutions_bl}, we assess the effects of our methods in isolation here and leave the evaluation of the combination to future work. 

\subsection{Study Design}
We make use of a unifactorial between-subject design in which each model constitutes one condition.
We randomly sample a set of 25 questions from the \textsc{HotpotQA} dev set and collect the model answer and explanation predictions (or annotations for GT) for each condition.
For each answer prediction, we manually assess whether it is equivalent to the ground truth answer.\footnote{For example, we consider the answer prediction ``Firth of Clyde, Scotland'' to be equivalent to the ground truth answer ``Firth of Clyde''.}
Each participant sees the 25 questions and the answers and explanations of one model in a random permutation.
For each question, we ask the participants to rate whether the model answer is correct.
In addition, we ask for multiple self-reports to assess, e.g., the trust of the user in the system.
In particular, we track the variables discussed in the following subsection.

\subsection{Dependent Variables}
\label{sec:dependent-variables}
We derive multiple dependent variables from the participants' ratings, namely completion time 
\citep{lim2009and, lage2019human},
several performance variables indicating how well they judged the correctness of the model (fraction of correct ratings, false positive ratio (FP), false negative ratio (FN), true positive ratio (TP), true negative ratio (TN), precision (P), recall (R) and \F\ values), agreement (fraction of model predictions that the users rate as correct \citep{bussone2015role}), and
overestimation (difference between agreement and true model accuracy \citep{nourani2019effects}).

Furthermore, we collect the following variables in self-reports with five-point Likert scales:
certainty of the participants \citep{greis2017detecting}, completeness and helpfulness of the explanations \citep{nourani2019effects}, trust of the participants in the model \citep{bussone2015role}, and satisfaction \citep{kulesza2012tell,greis2017input}.

All the questions and screenshots of our study are given in the appendix.

\begin{table*}
\centering
\resizebox{\textwidth}{!}{%
\begin{tabular}{l!{\vrule width 1pt}cccccccccccc!{\vrule width 1pt}cccccccc}
\toprule
		& \multicolumn{12}{c}{Standard Scores} &  \multicolumn{7}{c}{Proposed Scores}    \\
		\cmidrule(lr){2-13}
		\cmidrule(lr){14-20} 
		& \multicolumn{4}{c}{Answer} &  \multicolumn{4}{c}{Supporting Facts}  &  \multicolumn{4}{c}{Joint} & \multicolumn{2}{c}{Answer Changes} & \multicolumn{2}{c}{\textsc{FaRM}} & \multicolumn{2}{c}{\%-in-fact} & \textsc{LocA} \\
		\cmidrule(lr){2-5}
		\cmidrule(lr){6-9}
		\cmidrule(lr){10-13}
		\cmidrule(lr){14-15}
		\cmidrule(lr){16-17}
		\cmidrule(lr){18-19}
		\cmidrule(lr){20-20}
Human eval. & EM & \F\ & P & R & EM & \F\ & P & R & EM & \F\ & P & R & rel. & irrel. & $\textsc{FaRM}(1)$ & $\textsc{FaRM}(4)$ & rel. & irrel. & $\textsc{LocA}$\\\midrule 
\rowcolor{lightgray} correct decision  &   &   &   &   &   &   &   & \cellcolor{tableGreen}{\circled{+}} &  & \cellcolor{tableRed}{-}& \cellcolor{tableRed}{-} &   &   & \cellcolor{tableGreen}{\circled{+}} & \cellcolor{tableGreen}{\circled{+}} &   & \cellcolor{tableGreen}{\circled{+}} &   &  \\
\rowcolor{white} overestimation  &   & \cellcolor{tableRed}{+} &   & \cellcolor{tableRed}{+} &   &   &   &   &   &   &   &   & \cellcolor{tableGreen}{\circled{-}} &   &   & \cellcolor{tableGreen}{\circled{-}} &   & \cellcolor{tableGreen}{\circled{-}} & \cellcolor{tableGreen}{\circled{-}}\\ 
\rowcolor{lightgray} completion time  & \cellcolor{tableRed}{+} &   & \cellcolor{tableRed}{+} &   & \cellcolor{tableGreen}{\circled{-}} & \cellcolor{tableGreen}{\circled{-}} & \cellcolor{tableGreen}{\circled{-}} &   & \cellcolor{tableGreen}{\circled{-}} &   &   & \cellcolor{tableRed}{+} &   &   &   &   &   &   &  \\ 
\rowcolor{white} human-FP  &   & \cellcolor{tableRed}{+} &   & \cellcolor{tableRed}{+} &   &   &   &   &   &   &   &   & \cellcolor{tableGreen}{\circled{-}} &   &   & \cellcolor{tableGreen}{\circled{-}} &   & \cellcolor{tableGreen}{\circled{-}} & \cellcolor{tableGreen}{\circled{-}}\\ 
\rowcolor{lightgray} human-TP  &   &   &   &   &   &   &   & \cellcolor{tableGreen}{\circled{+}} &  & \cellcolor{tableRed}{-}& \cellcolor{tableRed}{-} &   &   & \cellcolor{tableGreen}{\circled{+}} & \cellcolor{tableGreen}{\circled{+}} &   & \cellcolor{tableGreen}{\circled{+}} &   &  \\ 
human-FN  & \cellcolor{tableGreen}{\circled{-}} &   & \cellcolor{tableGreen}{\circled{-}} &   & \cellcolor{tableRed}{+} & \cellcolor{tableRed}{+} & \cellcolor{tableRed}{+} &   & \cellcolor{tableRed}{+} &   &   & \cellcolor{tableGreen}{\circled{-}} &   &   &   &   &   &   &  \\
\rowcolor{lightgray} human-TN  &   &   &   &   &   &   &   & \cellcolor{tableGreen}{\circled{+}} &  & \cellcolor{tableRed}{-}& \cellcolor{tableRed}{-} &   &   & \cellcolor{tableGreen}{\circled{+}} & \cellcolor{tableGreen}{\circled{+}} &   & \cellcolor{tableGreen}{\circled{+}} &   &  \\
\rowcolor{white} human-P  &  & \cellcolor{tableRed}{-} &  & \cellcolor{tableRed}{-} &   &   &   &   &   &   &   &   & \cellcolor{tableGreen}{\circled{+}} &   &   & \cellcolor{tableGreen}{\circled{+}} &   & \cellcolor{tableGreen}{\circled{+}} & \cellcolor{tableGreen}{\circled{+}}\\ 
\rowcolor{lightgray} human-R  &   &   &   &   &   &   &   & \cellcolor{tableGreen}{\circled{+}} &  & \cellcolor{tableRed}{-}& \cellcolor{tableRed}{-} &   &   & \cellcolor{tableGreen}{\circled{+}} & \cellcolor{tableGreen}{\circled{+}} &   & \cellcolor{tableGreen}{\circled{+}} &   &  \\ 
\rowcolor{white} human-\F\  &   &   &   &   &   &   &   & \cellcolor{tableGreen}{\circled{+}} &  & \cellcolor{tableRed}{-}& \cellcolor{tableRed}{-} &   &   & \cellcolor{tableGreen}{\circled{+}} & \cellcolor{tableGreen}{\circled{+}} &   & \cellcolor{tableGreen}{\circled{+}} &   &  \\ \bottomrule

\end{tabular}
}
\caption{The table shows whether sorting the conditions by a human score (rows) and an automatized score (columns) results in the same order (+), the inverse order (-) or a different order (blank cell). Green (\textcolor{tableGreen}{$\blacksquare$}) cells with circles mark desirable relations, red (\textcolor{tableRed}{$\blacksquare$}) cells without circles mark undesirable relations.}\label{tab:ml_human}
\end{table*}

\subsection{Participants and Data Cleaning}
We collect the ratings of 40 participants (16 female, 24 male) with a mean age of 26.6 years ($SD=3.4$).
We filter out all responses with a completion time smaller than 15 seconds or larger than 5 minutes as this indicates that the participant did not read the whole explanation or was interrupted during the study.
We further asked them whether they knew the answer before and exclude the responses to those questions from our evaluation.
In total, we discard 12.10\% of the responses.

\subsection{Results}
In this section, we summarize the main results of the user study. For better overview, we do not include evaluations on every variable from Section \ref{sec:dependent-variables} but show them in the appendix.

Figure \ref{fig:correctness_means_box} shows the fraction of correct user ratings of model correctness. 
The correctness of our proposed models can be better judged than the Qi-2019 model:
The regularized model and the S\&F model increase the fraction by 10.79\% and 9.17\%, respectively, compared to Qi-2019.
The GT comparison shows an upper bound.

Among the performance variables, the false positive ratio deserves particular attention as a false positive corresponds to a user thinking the model answer is correct while it is not.
Such an error can be dangerous in safety-critical domains.
Figure~\ref{fig:FP} shows that the fraction of FPs is decreased by 6.43\% by the regularized model and by 9.25\% by the S\&F model compared to Qi-2019.
The ground truth has zero false positives by definition.

A similar effect can be seen when evaluating overestimation: Both our models alleviate overestimation as shown in Figure~\ref{fig:overestimation}.
While participants overestimate the model accuracy of Qi-2019 by 2.93\% on average, the regularized model only leads to 0.87\% overestimation.
The S\&F model is even underestimated by 6.40\% on average.
While an ideal model would lead to neither over- nor underestimation, underestimation can be preferable to overestimation if the model is deployed into high-risk environments, such as medical contexts.
In general, a reduction in overestimation can --- besides enhanced fact selection --- also be linked to an improved answer accuracy as better-performing models naturally leave more room for underestimation.

Finally, we consider relations within the variables from Section \ref{sec:dependent-variables}.
Figure~\ref{fig:ct_of_certainty} shows that mean completion time monotonously decreases with increasing user certainty 
(with the exception of ``strongly disagree'').\footnote{The low completion time for ``strongly disagree'' could indicate that the users could not find any relation at all between answer and explanation.}
This confirms the findings of \citet{greis2017detecting} who investigate the effect of user uncertainty on behavioral measurements.

\begin{figure}[h]
	    \centering
        \includegraphics[width=0.9\columnwidth]{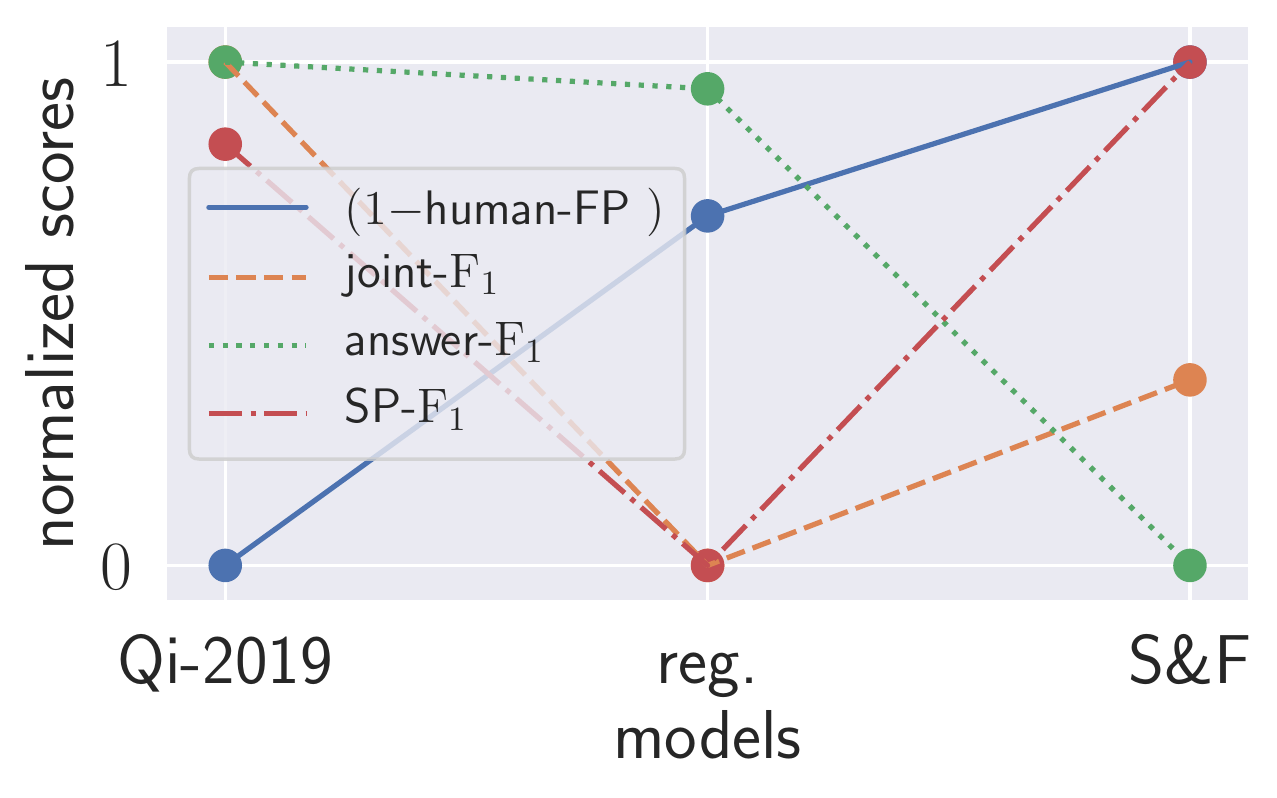}
        \includegraphics[width=0.9\columnwidth]{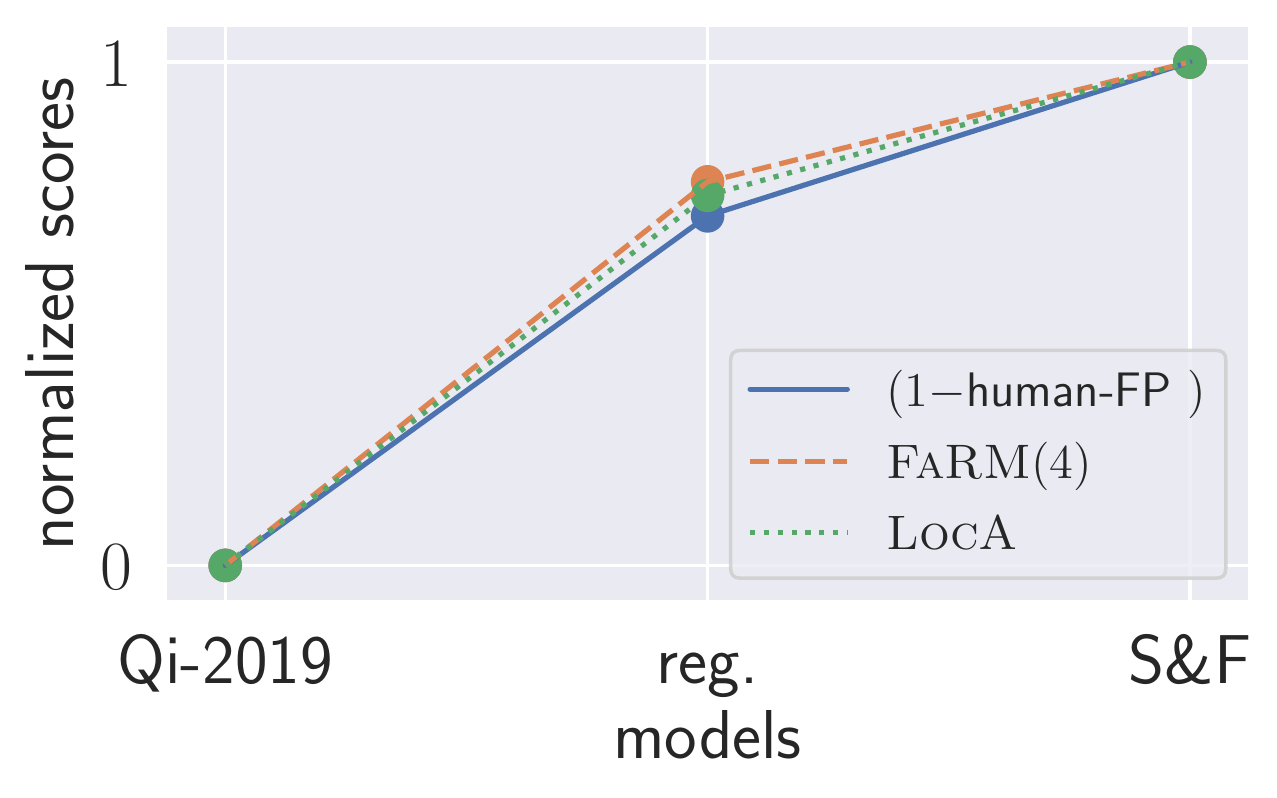}%
        \caption{Model score comparisons between human-false positives and model scores. All scores are normalized to $[0,1]$.
        We show $(1-$human-FP$)$ as less FPs are better.
        The upper figure shows that \F\ poorly correlates to human performance. The lower figure shows a much stronger correlation for our proposed scores.}\label{fig:contrast_f1_ours}
\end{figure}

\subsection{Correlation with Evaluation Scores}
Finally, we investigate the correlation of human ratings with model evaluation scores.
We rank the models by (\romannum{1}) human measures obtained in the user study and (\romannum{2}) model evaluation scores.
In Table~\ref{tab:ml_human}, a cell is marked with a ``+'' if the ranking with respect to the human measure and the model score is identical (e.g., the ranks regarding human-FPs and answer-\F\ are identical).
If the ranks are exactly reversed, we mark the cell with a ``-''.
All other cells are left empty.
``+'' and ``-'' both indicate a perfect correlation and do not imply one being preferable over the other.
Next, we consider whether selecting a model based on the different model scores would result in a desired change in human evaluation scores or not.
This depends on whether a high score (e.g., \F) or a low score (e.g., the fraction of answers outside the predicted relevant facts) is aimed for.
We indicate desired model selection with green circled cells and undesired model selection with red cells
(e.g., choosing a model with a higher answer-\F\ would result in a model with more human-FPs. This is not desired.)
All \F-scores show at least one undesirable rank relation.
Notably, joint-\F\ is among the least aligned scores.
In contrast, our scores have only desirable relations.
In particular, 
$\textsc{FaRM(4)}$ and \textsc{LocA}
lead to a model ranking that is inverse to the ranking by human-overestimation and human-FPs.
This is also confirmed in Figure~\ref{fig:contrast_f1_ours}, which shows how the human-FP ratio varies in comparison to the three \F\ scores (upper plot: no correlation) and to our proposed scores (lower plot: correlated).
See appendix for other dependent variables.

To sum up, our results indicate that (\romannum{1}) \F\ is not suited to quantify the explanatory power of a model
and (\romannum{2}) our proposed scores predict user behavior better than standard scores, opening the possibility of using them for model selection.

\section{Conclusion}\label{sec:conclusion}
In this paper, we investigated explainable question answering, revealing that existing models lack an explicit coupling of answers and explanations and that evaluation scores used in related work fail to quantify that. This highly impairs their applicability in real-life scenarios with human users.
As a remedy, we addressed both modeling and evaluation, proposing a hierarchical neural architecture, a regularization term, as well as two new evaluation scores.
Our user study showed that our models help the users assess their correctness and that our proposed evaluation scores are better correlated with user experience than standard measures like \F.

\section*{Acknowledgement}
We thank the members of the BCAI NLP\&KRR research group and the anonymous reviewers for their helpful comments.
Ngoc Thang Vu is funded by Carl Zeiss Foundation.

\bibliography{references}
\bibliographystyle{acl_natbib}

\appendix
\section*{Appendix}\label{sec:appendix}
\section{\textsc{HotpotQA} Data Set}
We conduct our analysis and experiments on the explainable multi-hop reasoning question answering data set \textsc{HotpotQA} \cite{yang2018hotpotqa}.
The data set contains 113k questions with crowd-sourced annotations for answers and explanations.
Explanations correspond to a selection of supporting facts (i.e., sentences) from Wikipedia articles.
The data set contains 90,564 training instances as well as 7405 development instances.
The test set is split into two separate test sets with 7405 instances each: the full-wiki test set and the distractor test set.
Models evaluated on the full-wiki test data need to retrieve relevant articles from a given set of Wikipedia articles (therefore \textit{full wiki}) while the distractor test set provides models with ten articles of which two contain the supporting facts and the other eight articles are \textit{distracting} the system.
Both test sets are not publicly available.
\F\ values on the test sets can be obtained by submitting a model to the leaderboard.
As our proposed \textsc{FaRM} and \textsc{LocA} scores need to access context information, we evaluate all tested models on the distractor development set while training them on the provided training data.
The training and development data as well as the leaderboard can be found online.\footnote{\url{https://hotpotqa.github.io/}}

\section{S\&F Workflow Example}
\begin{figure}[H]
    \centering
        \includegraphics[width=\columnwidth]{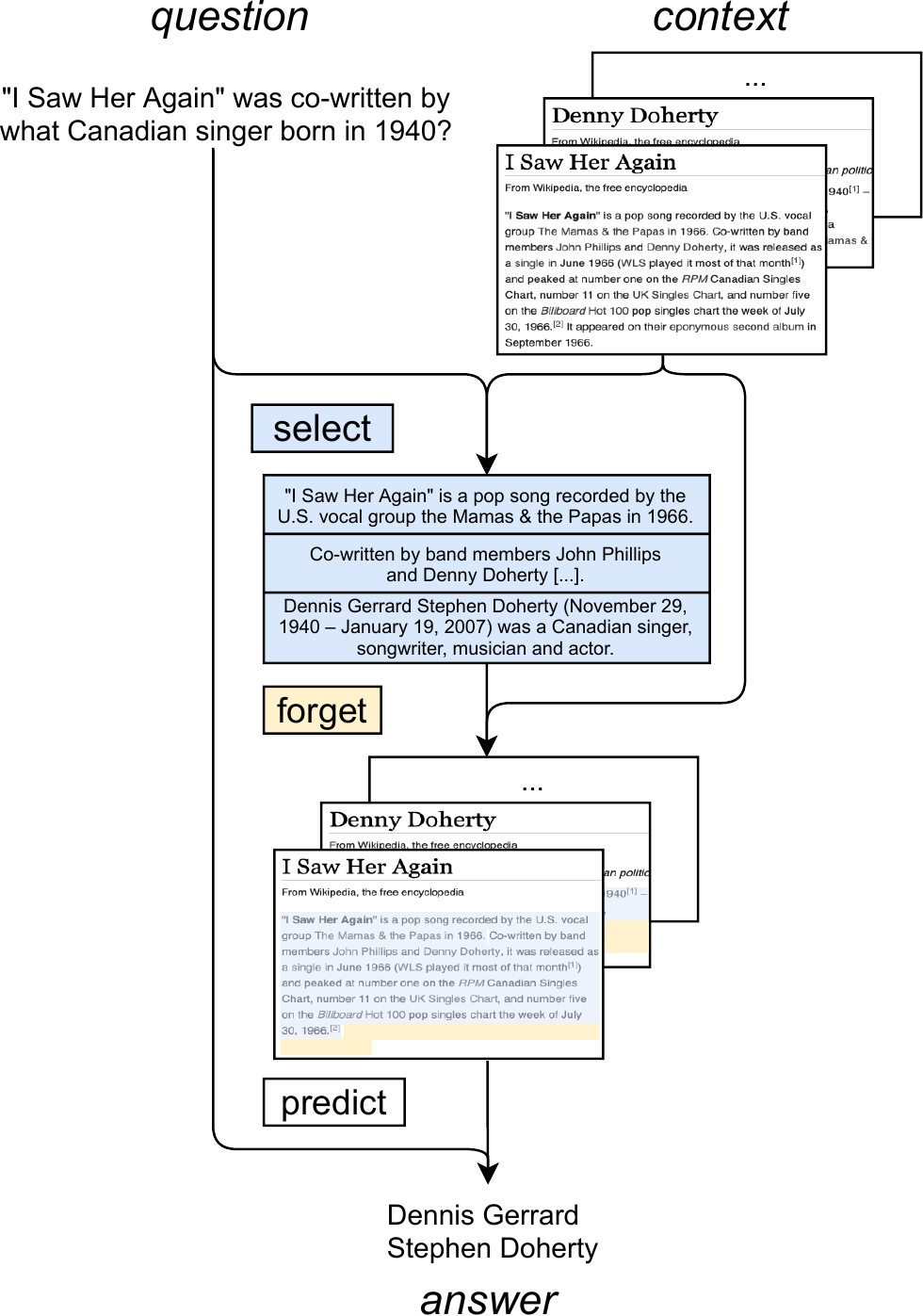}
    \caption{Exemplary workflow of the S\&F model.}\label{fig:s_and_F_example}
\end{figure}

Figure~\ref{fig:s_and_F_example} shows an exemplary workflow of the S\&F architecture.
First, relevant facts are selected.
Second, all other facts are masked and third, the answer is predicted based on the masked context.

\section{Training Details and Hyperparameters}\label{apx_sec:hyperparameters}
We use the same preprocessing, hyperparameters and early stopping procedure as \citet{qi2019answering} to train our models.\footnote{\url{https://github.com/qipeng/golden-retriever}}
The Qi-2019 model as well as our regularized adaption contain 99M parameters, our S\&F model and its regularized version contain 100M parameters each.
The additional hyperparameters of our regularization term are optimized with random search using 100 runs.
In particular, we sample $c_i \sim U([0.0, 5.0]), i \in \left\{1,2,3\right\}$ and select $p_{\text{e}}^{*}$ and $p_{\text{e}}^{+}$ with equal probability. 
We select the best model based on the percentage of answer spans inside facts predicted as relevant to ensure decent answer-explanation coupling while not directly optimizing on the \textsc{LocA}.
Based on our hyperparameter search, the best regularization parameters for the model proposed by \citet{qi2019answering} are $c_1 = 4.96$, $c_2 = 2.02$ and $c_1 = 3.10$.
The best parameters for the regularized S\&F model are $c_1 = 1.18$, $c_2 = 0.24$ and $c_1 = 1.61$.
We trained all models on Nvidia Tesla V100 GPUs.

\section{Comparison of BiDaf++ Versions}
We compare the BiDaf++ model used as the official \textsc{HopotQA} baseline of \citet{yang2018hotpotqa} to the modified model by \citet{qi2019answering} in Table~\ref{tab:comp_rw}.
As the modified model outperforms the Yang-2018 model on all metrics, we use the model proposed by \citet{qi2019answering} throughout our experiments as well as the user study.
\begin{table}
\centering
\footnotesize
\begin{tabular}{p{0.1cm}l|cc}
\toprule
& Metric & Yang-2018 & Qi-2019 \\ \midrule
\multirow{12}{*}{\rotatebox[origin=c]{90}{Standard Scores}} & Answer-EM& 43.74& \textbf{49.48}\\
& Answer-\F& 57.29& \textbf{63.76}\\
& Answer-P& 59.76& \textbf{66.26}\\
& Answer-R& 58.74& \textbf{65.52}\\
& SP-EM& 24.54& \textbf{39.81}\\
& SP-\F& 68.02& \textbf{79.34}\\
& SP-P& 69.86& \textbf{78.01}\\
& SP-R& 72.90& \textbf{85.26}\\
& Joint-EM& 12.83& \textbf{22.28}\\
& Joint-\F& 41.12& \textbf{52.51}\\
& Joint-P& 43.43& \textbf{53.33}\\
& Joint-R& 45.44& \textbf{57.92}\\
\midrule
\multirow{6}{*}{\rotatebox[origin=c]{90}{Proposed Scores}} & $\textsc{FaRM}(4)$ & 50.08& \textbf{66.20}\\
& $\drsh$ $c_{\text{rel}}(4)$& 62.50& \textbf{77.06}\\
& $\drsh$ $c_{\text{irrel}}(4)$& 24.81& \textbf{16.39}\\
& \textsc{LocA}& 44.78& \textbf{60.49}\\
& $\drsh$ I & 54.90& \textbf{67.48}\\
& $\drsh$ O & 22.60& \textbf{11.55}\\
\bottomrule

\end{tabular}%
\caption{Comparison of the \textsc{HotpotQA} baseline model by \citet{yang2018hotpotqa} and the modified model by \citet{qi2019answering}. The modified model outperforms the baseline on all scores.
All values in \%.}\label{tab:comp_rw}
\end{table}

\section{Hierarchical Error Propagation}
As the S\&F model first selects the supporting facts and then predicts the answer based on the selected subset, we evaluate how errors in the fact selection effect the answer prediction and compare it to the Qi-2019 model.
For both models, we compare the fraction of predictions with correct (exact match) answers among predictions that (\romannum{1}) contain all ground truth facts or (\romannum{2}) contain no ground truth facts.
We observe that for Qi-2019 the fraction of correct answers drops from 51.5\% for predictions with all ground truth facts to 25.8\% for predictions with no ground truth facts.
For the S\&F model, we observe a drop from 50.4\% to 20.6\% respectively.
This confirms our expectations of an increased error propagation in the S\&F model.
However, only 1.8\% of the S\&F fact predictions contain no ground truth fact at all, whereas for Qi-2019 this case occurs for 2.1\% of the predictions.

\section{Answer Changes per Removal Step}
\begin{figure}[t]
    \centering
        \includegraphics[width=\columnwidth]{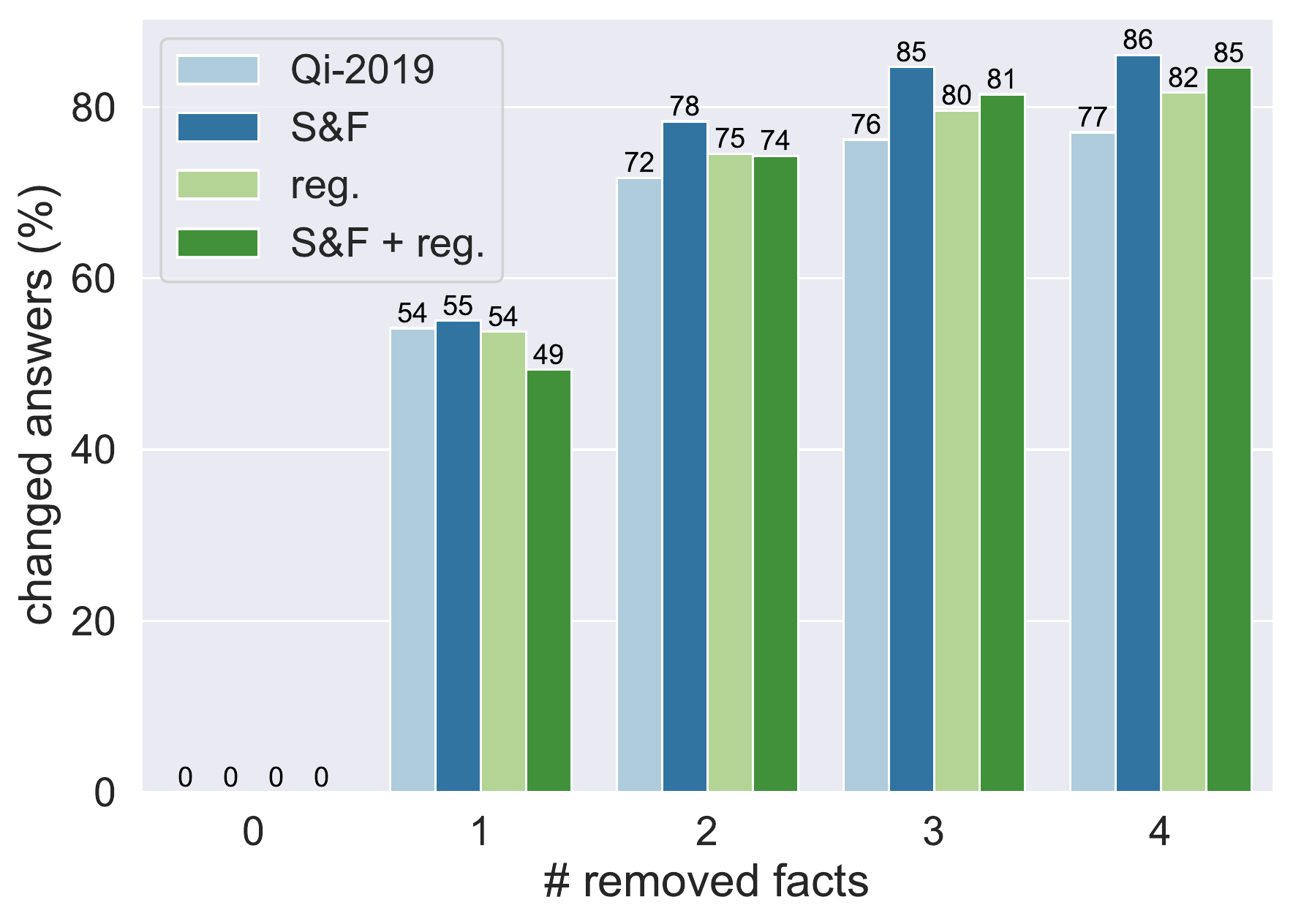}
        \includegraphics[width=\columnwidth]{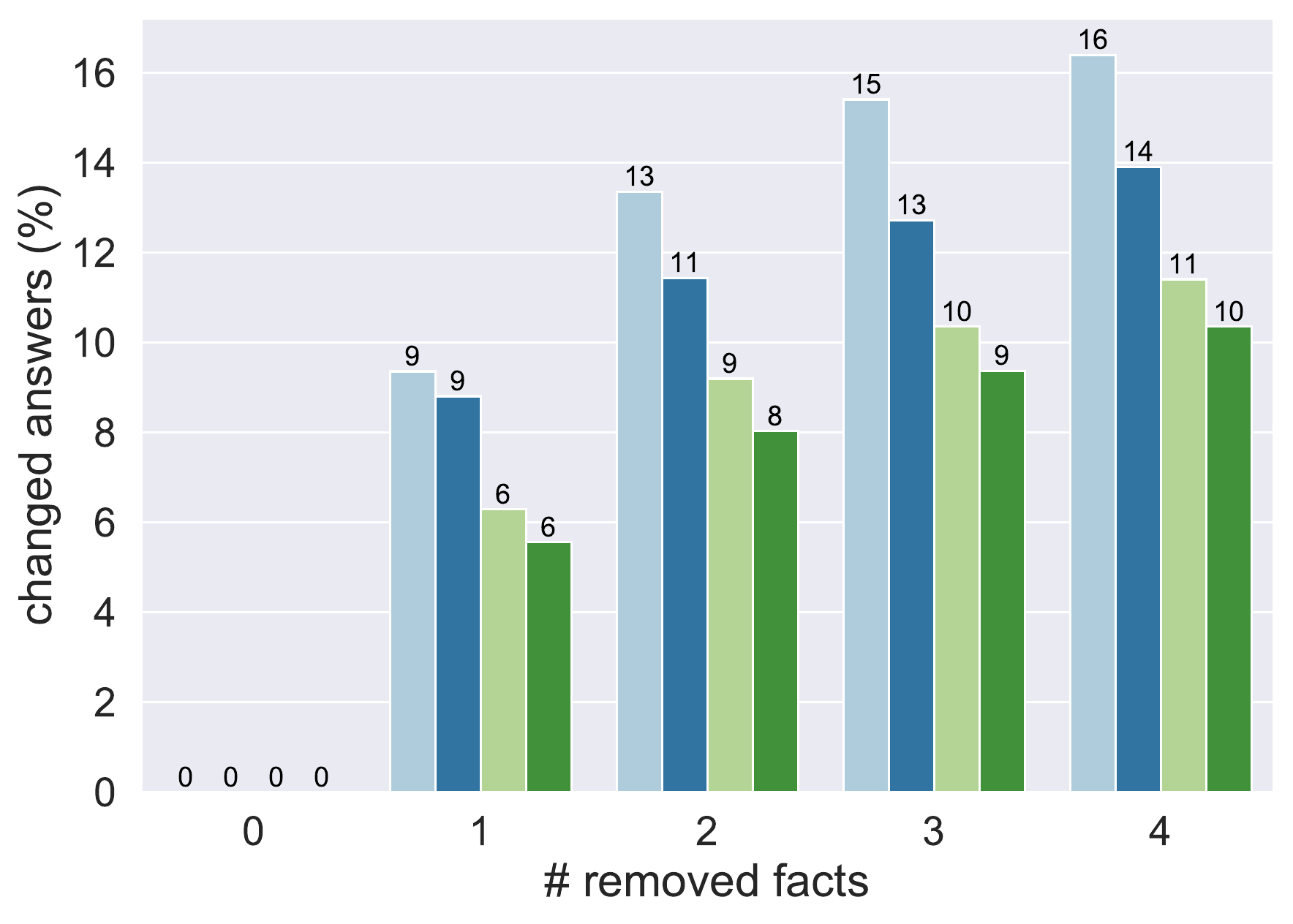}
    \caption{Answer changes $c_\text{rel}(k)$ and  $c_\text{irr}(k)$ when removing facts predicted as relevant (top) and irrelevant (bottom).}
    \label{fig:flips}
\end{figure}

Figure~\ref{fig:flips} shows the values of $c_\text{rel}(k)$ (top), i.e., the fraction of changed answers when removing $k$ facts predicted as relevant, and $c_\text{irr}(k)$ (bottom), i.e., the fraction of changed answers when removing $k$ facts predicted as irrelevant, for $k \in \{0, ..., 4\}$.

\section{User Study Design Details}
\begin{figure*}
    \centering
        \includegraphics[trim=0.0cm 11.0cm 0.0cm 0.0cm, clip, width=\textwidth]{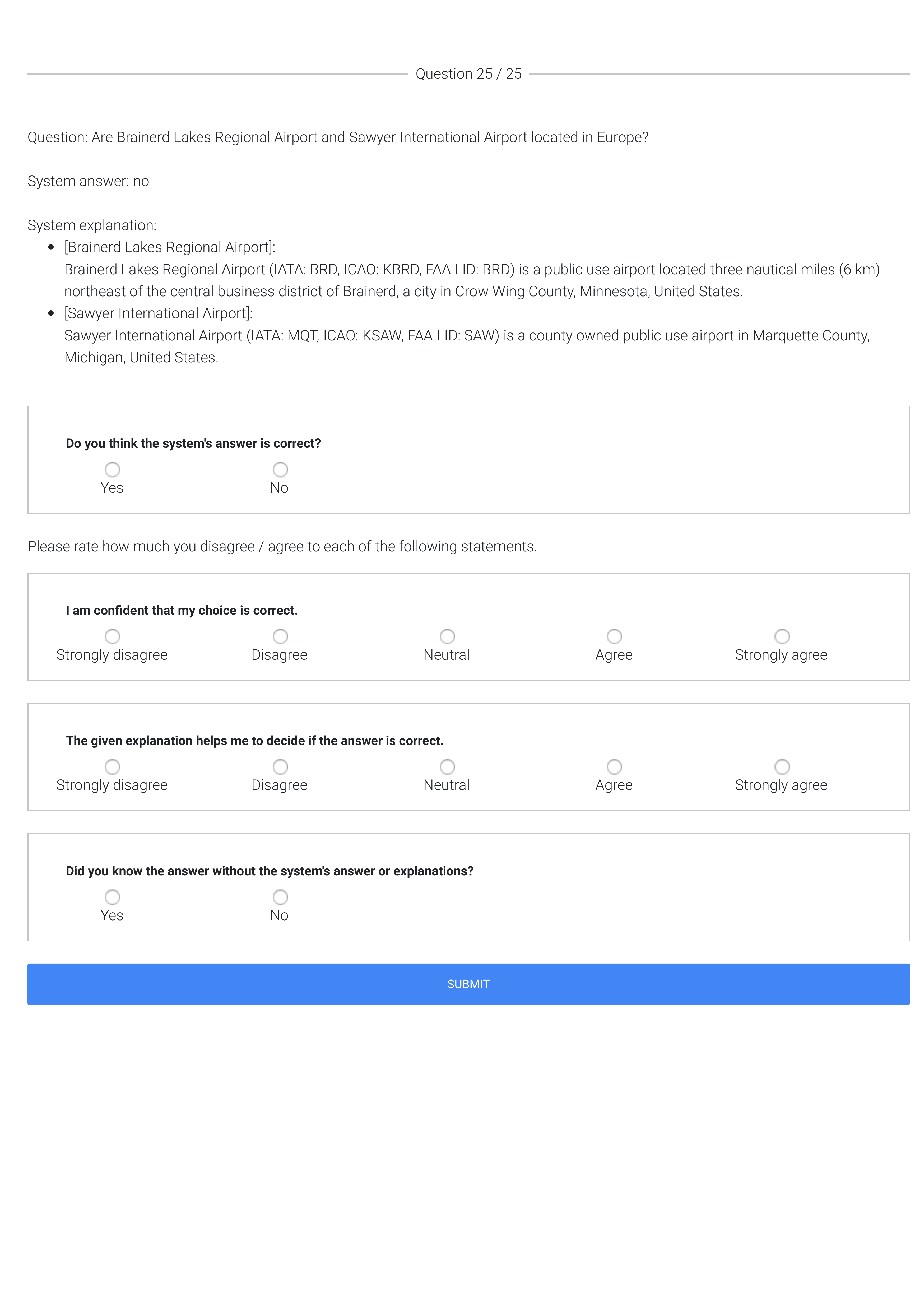}
    \caption{Screenshot of the question rating interface.}\label{fig:screenshot_1}
\end{figure*}

\begin{figure*}
    \centering
        \includegraphics[trim=0.0cm 19.5cm 0.0cm 0.0cm, clip, width=\textwidth]{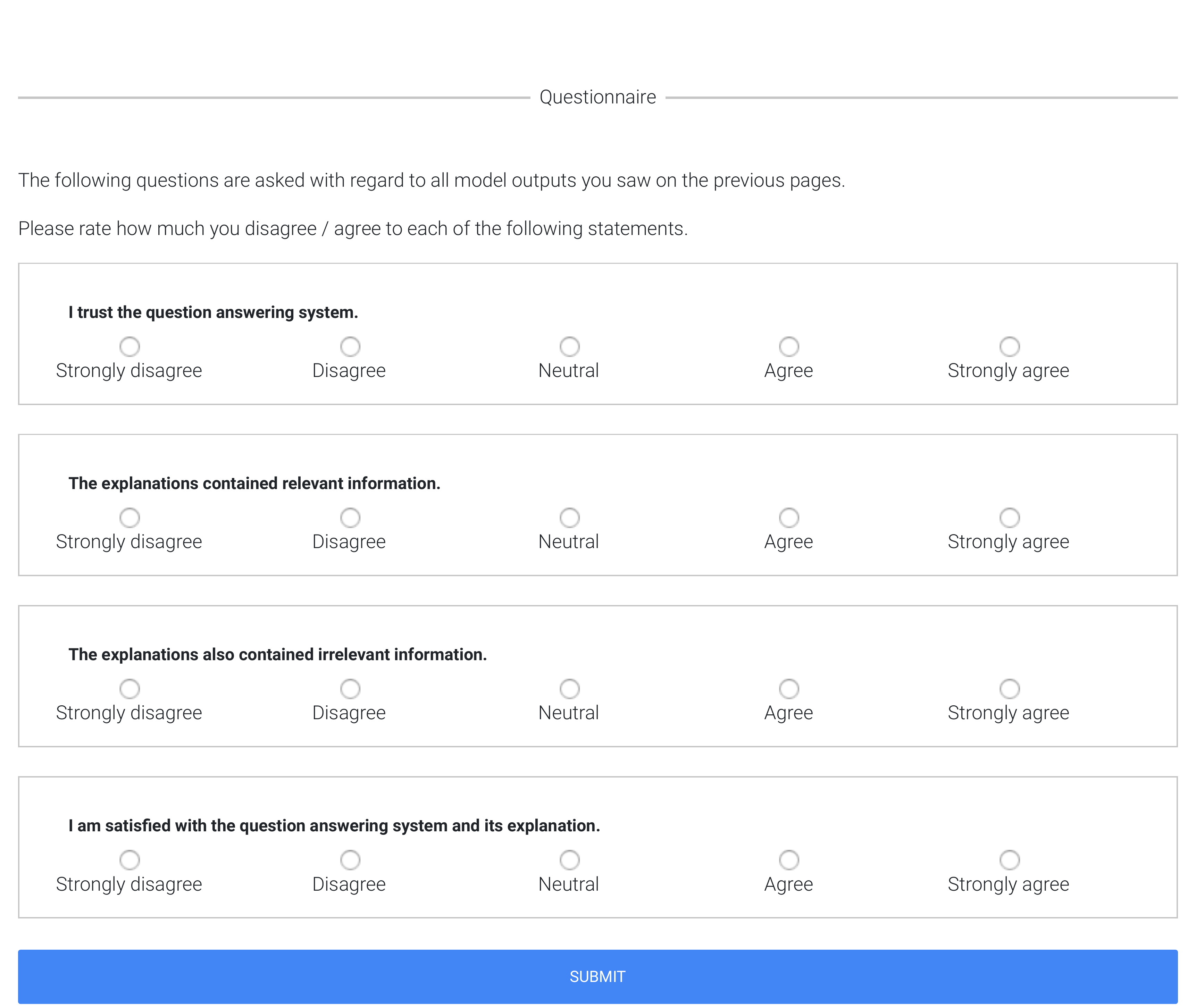}
    \caption{Screenshot of the post-questionnaire.}\label{fig:screenshot_2}
\end{figure*}

All questions and statements that we ask participants to answer/rate are listed in Table~\ref{tab:study_questions}.
\begin{table*}
\centering
\begin{tabular}{c|l|c}
\toprule
Context & Question/Statement & Answer Range \\
\midrule
\parbox[t]{2mm}{\multirow{4}{*}{\rotatebox[origin=c]{90}{\centering \small Each Question}}}
& Do you think the  system's answer is correct? & yes/no\\
& Did you know the answer without the system's answer or explanations? & yes/no\\
& I am confident that my choice is correct. & 5-point Likert\\
& The given explanation helps me to decide if the answer is correct. & 5-point Likert\\
\midrule
\parbox[t]{2mm}{\multirow{4}{*}{\rotatebox[origin=c]{90}{\centering \small Post Survey}}}
& I trust the question answering system. & 5-point Likert\\
& The explanations contained relevant information. & 5-point Likert\\
& The explanations also contained irrelevant information. & 5-point Likert\\
& I am satisfied with the question answering system and its explanations. & 5-point Likert\\
\bottomrule
\end{tabular}%
\caption{Questions and statements shown to the participants for (a) each question (upper part) and (b) in the post questionnaire (lower part). Statements were presented along with the prompt ``Please rate how much you disagree/agree to each of the following statements''.}\label{tab:study_questions}
\end{table*}
Figure~\ref{fig:screenshot_1} and Figure~\ref{fig:screenshot_2} show screenshots of the study interface for the question rating and the post-questionnaire.

\section{Detailed User Study Results}
\begin{figure*}
    \centering
    \begin{subfigure}[t]{.25\textwidth}
    \centering
    \includegraphics[width=\textwidth]{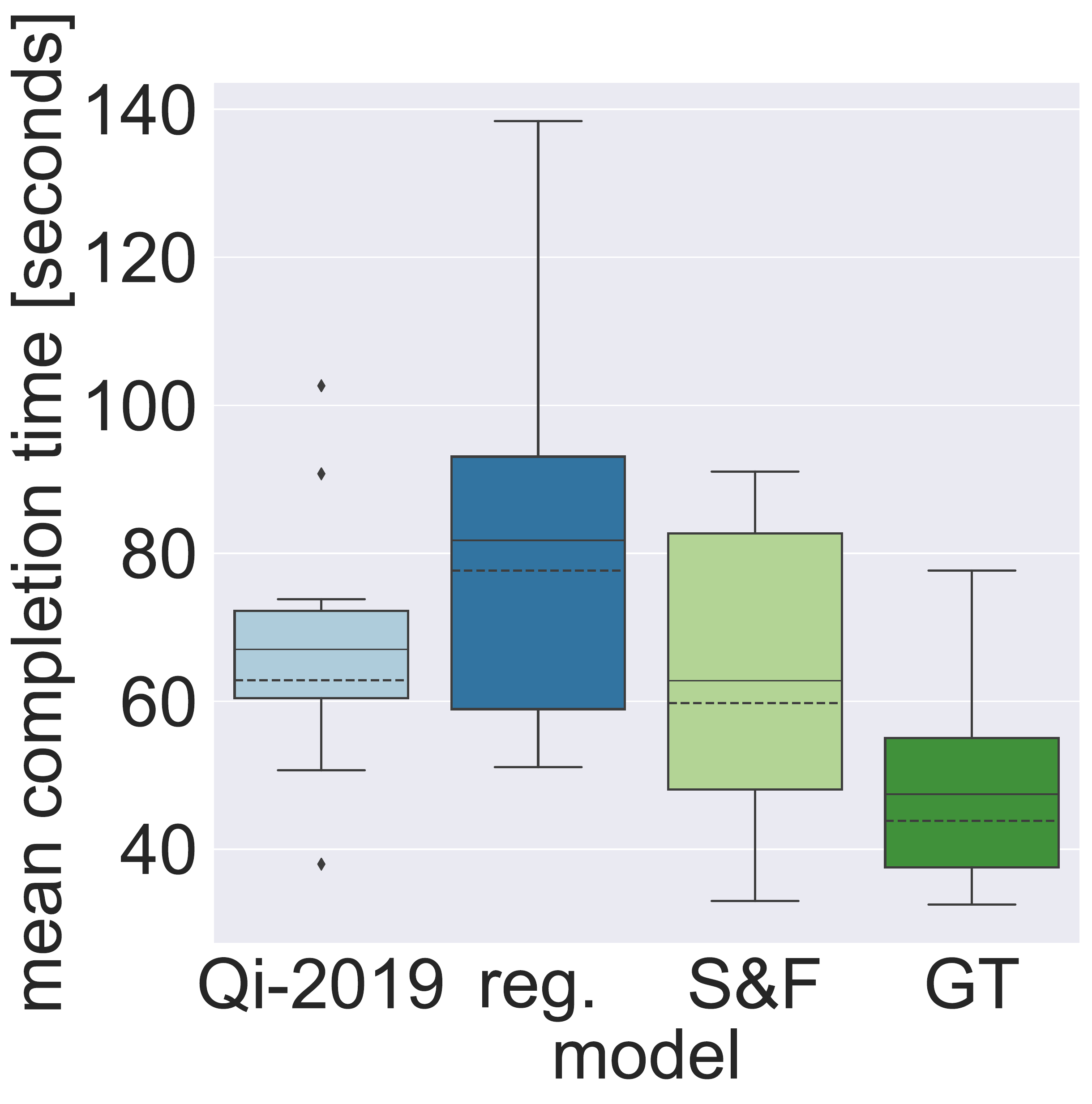}
    \caption{Mean completion times.}
    \end{subfigure}%
    \begin{subfigure}[t]{.25\textwidth}
    \centering
    \includegraphics[width=\textwidth]{figures/rating_is_correct_fraction_box.pdf}
    \caption{Fraction of correct decisions.}
    \end{subfigure}%
    \begin{subfigure}[t]{.25\textwidth}
    \centering
    \includegraphics[width=\textwidth]{figures/overestimation_box.pdf}
    \caption{Overestimation in percent.}
    \end{subfigure}%
    \begin{subfigure}[t]{.25\textwidth}
    \centering
    \includegraphics[width=\textwidth]{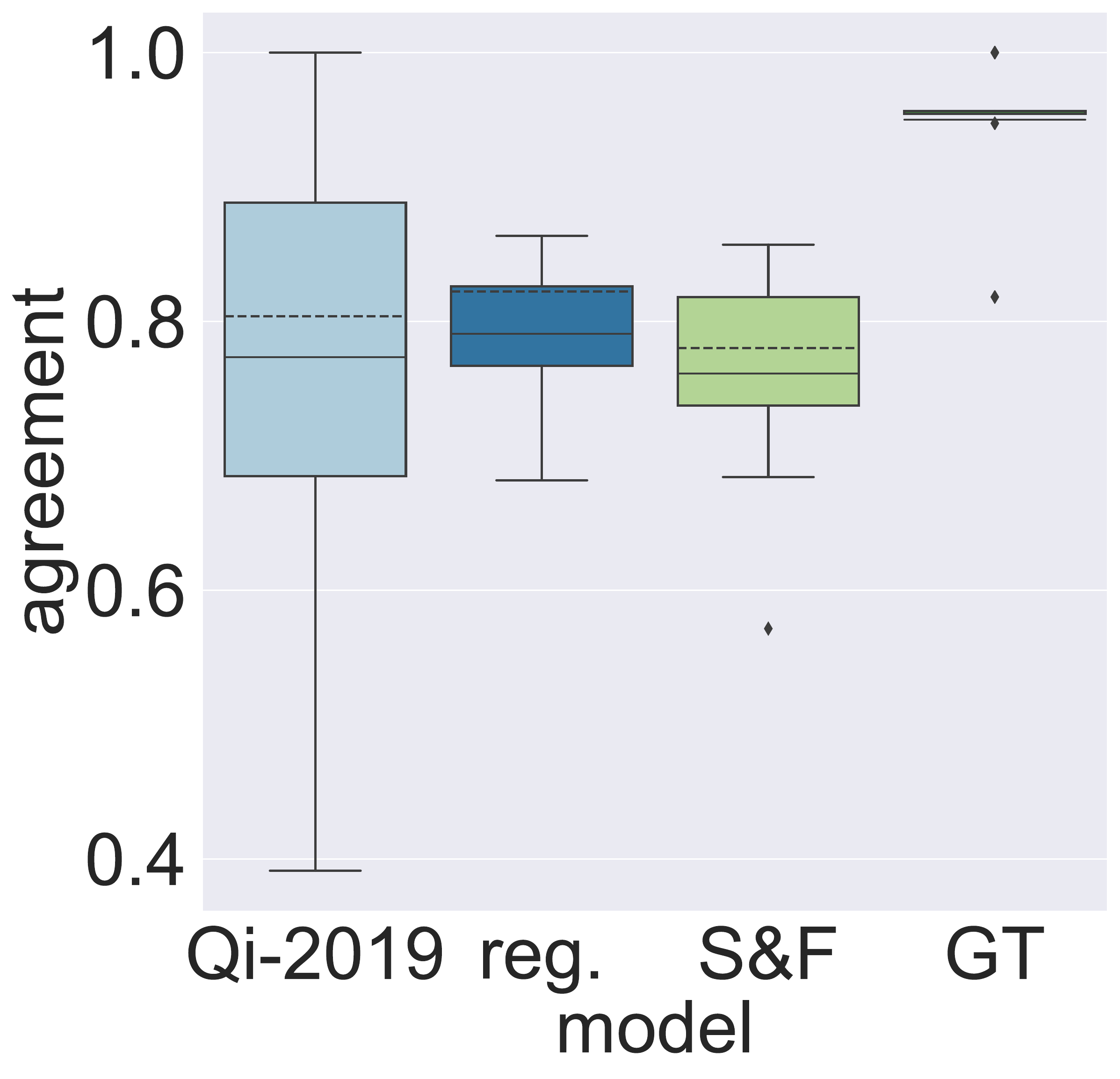}
    \caption{Agreement ratio.}
    \end{subfigure}
    ~
    \begin{subfigure}[t]{.25\textwidth}
    \centering
    \includegraphics[width=\textwidth]{figures/FP_box.pdf}
    \caption{False positive ratio.}
    \end{subfigure}%
    \begin{subfigure}[t]{.25\textwidth}
    \centering
    \includegraphics[width=\textwidth]{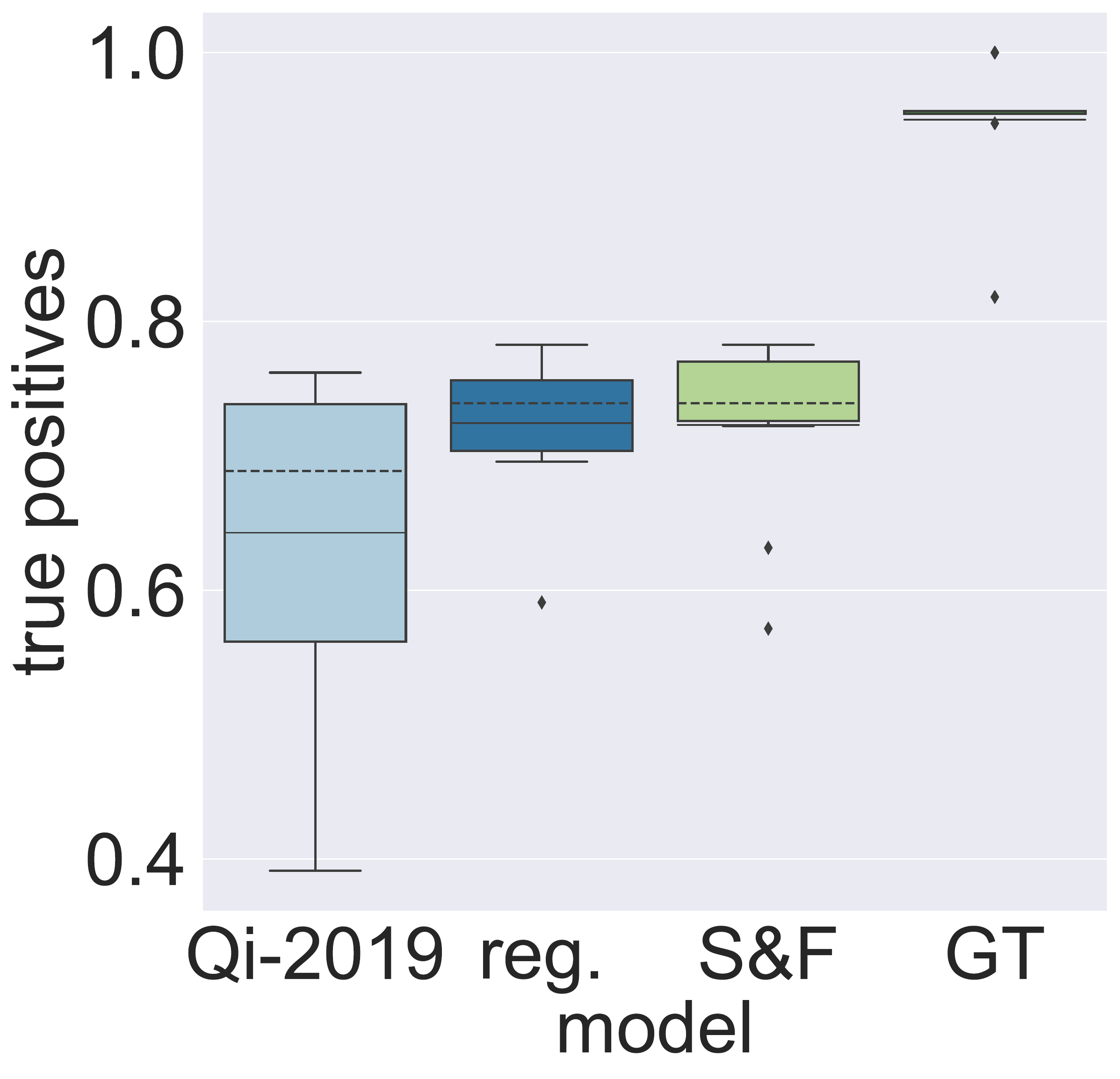}
        \caption{True positive ratio.}
    \end{subfigure}%
    \begin{subfigure}[t]{.25\textwidth}
    \centering
    \includegraphics[width=\textwidth]{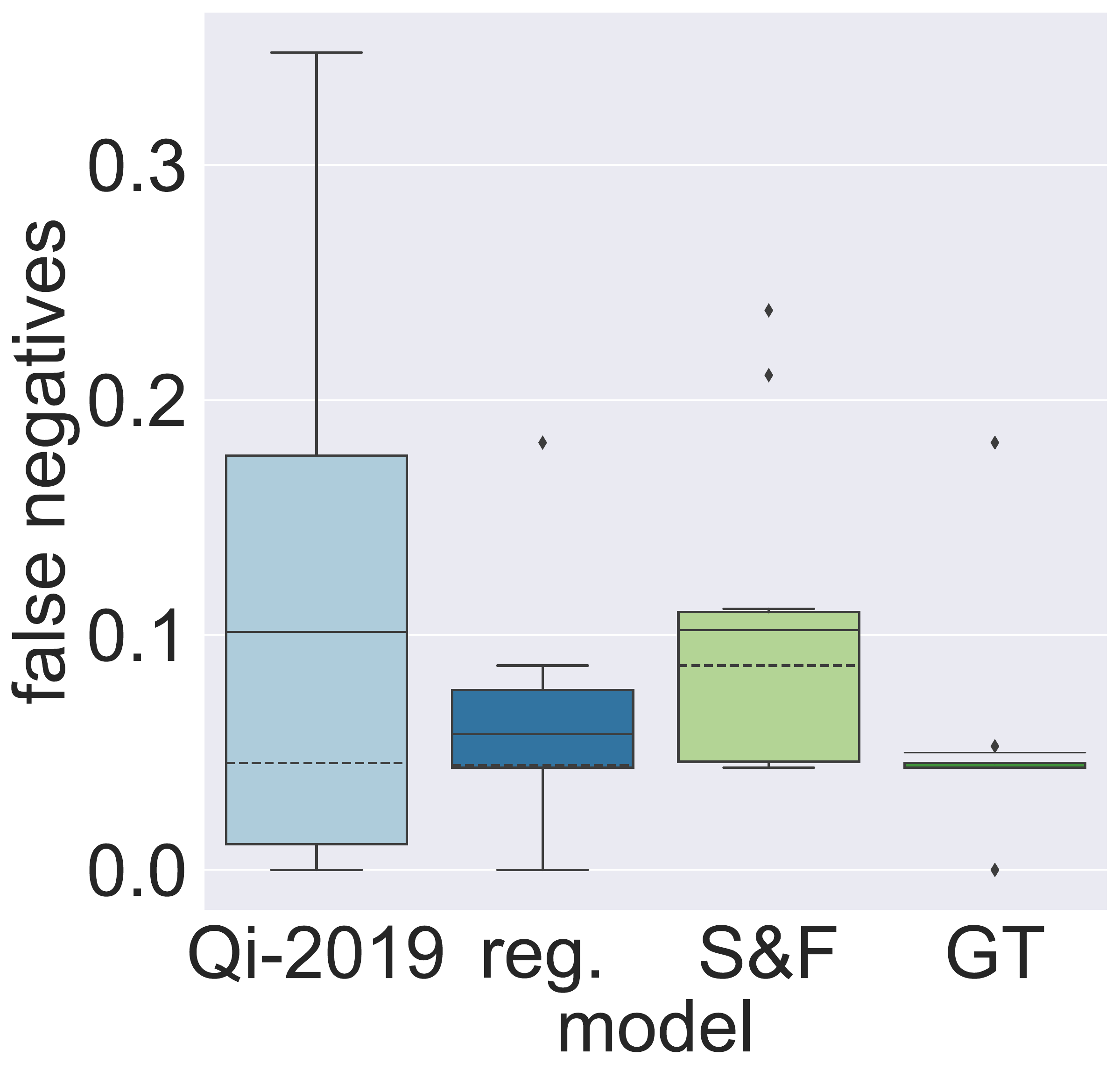}
    \caption{False negative ratio.}
    \end{subfigure}%
    \begin{subfigure}[t]{.25\textwidth}
    \centering
    \includegraphics[width=\textwidth]{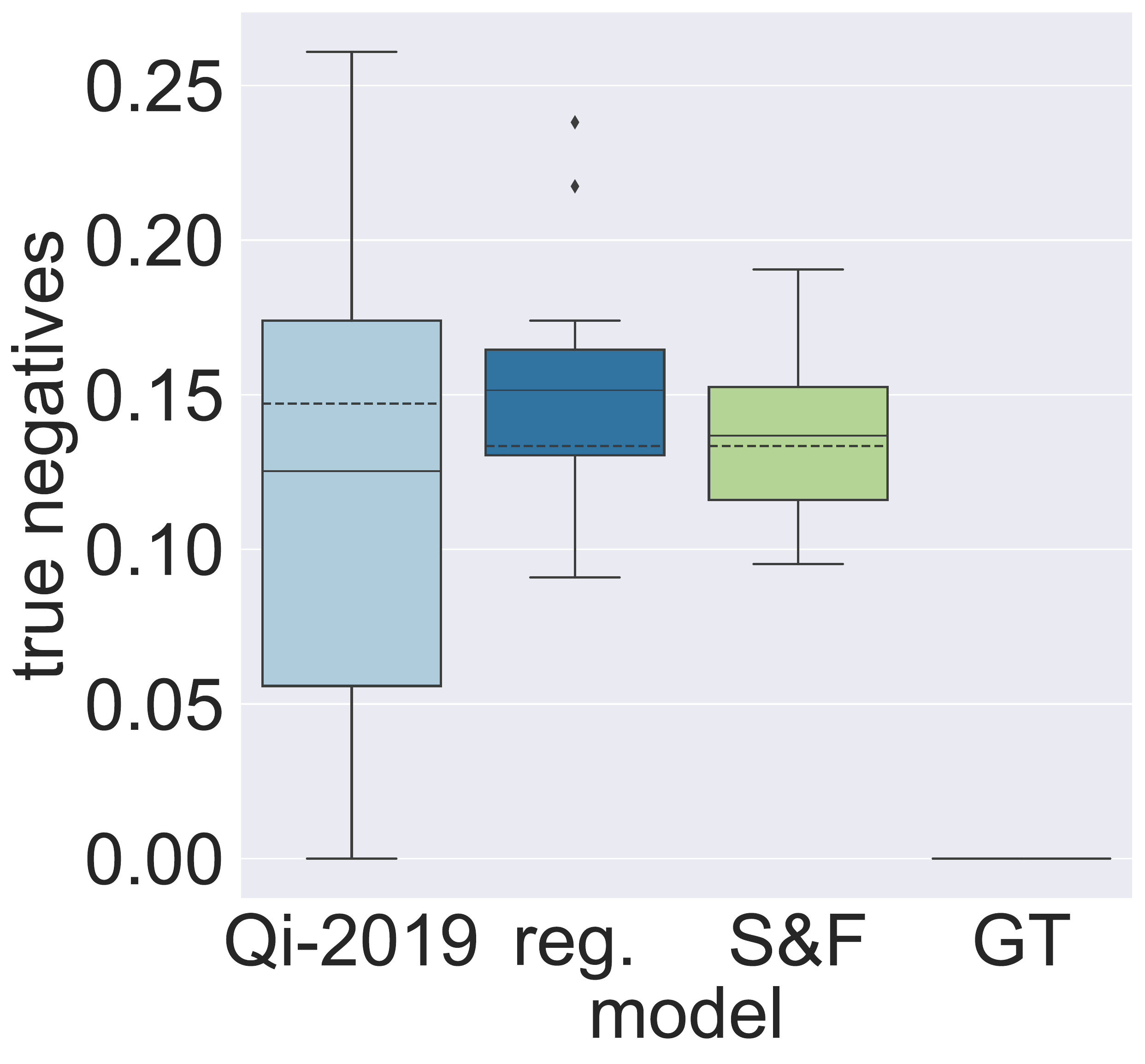}
        \caption{True negative ratio.}
    \end{subfigure}
    ~
    \begin{subfigure}[t]{.25\textwidth}
    \centering
    \includegraphics[width=\textwidth]{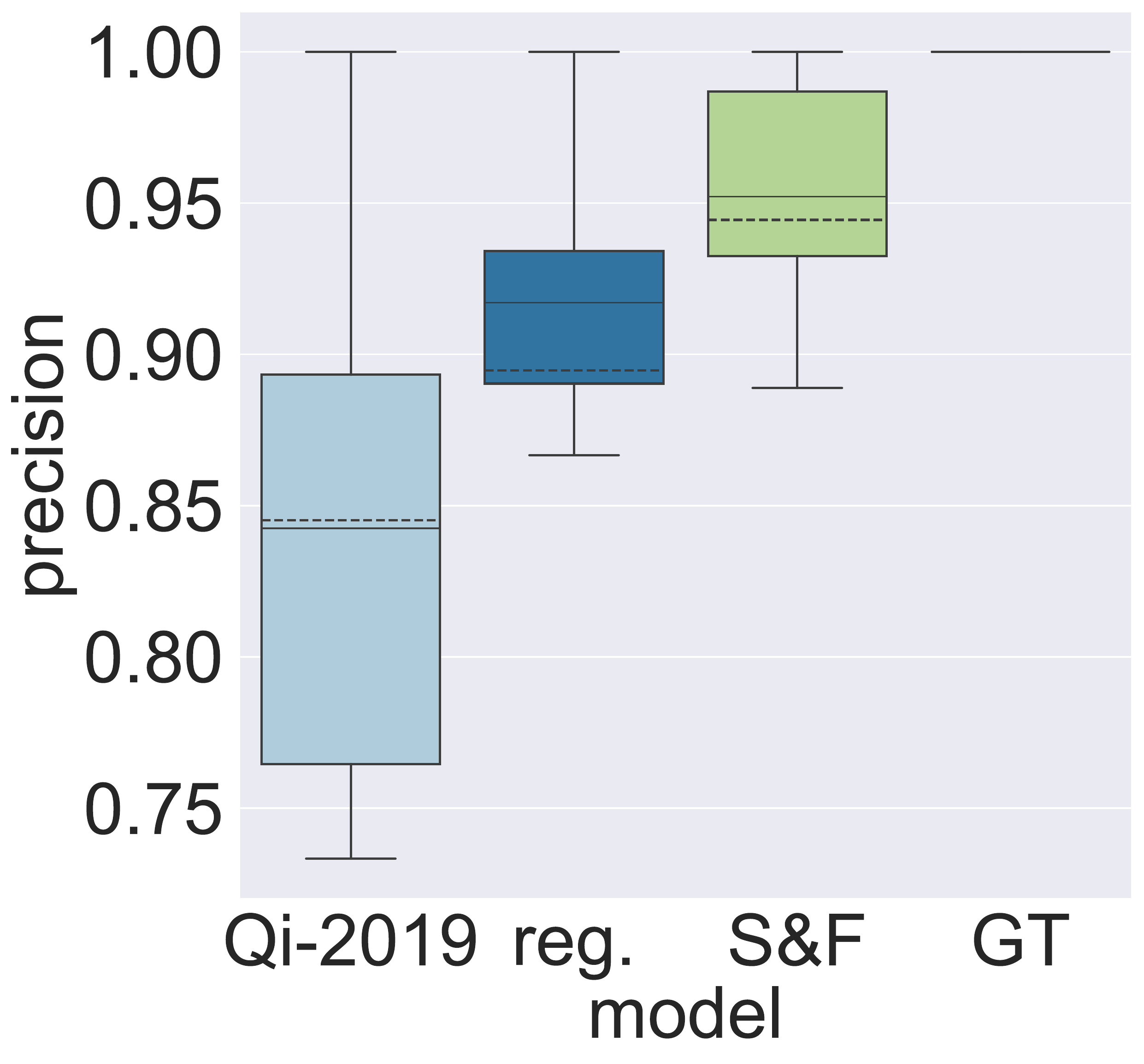}
        \caption{Precision of user ratings.}
    \end{subfigure}%
    \begin{subfigure}[t]{.25\textwidth}
    \centering
    \includegraphics[width=\textwidth]{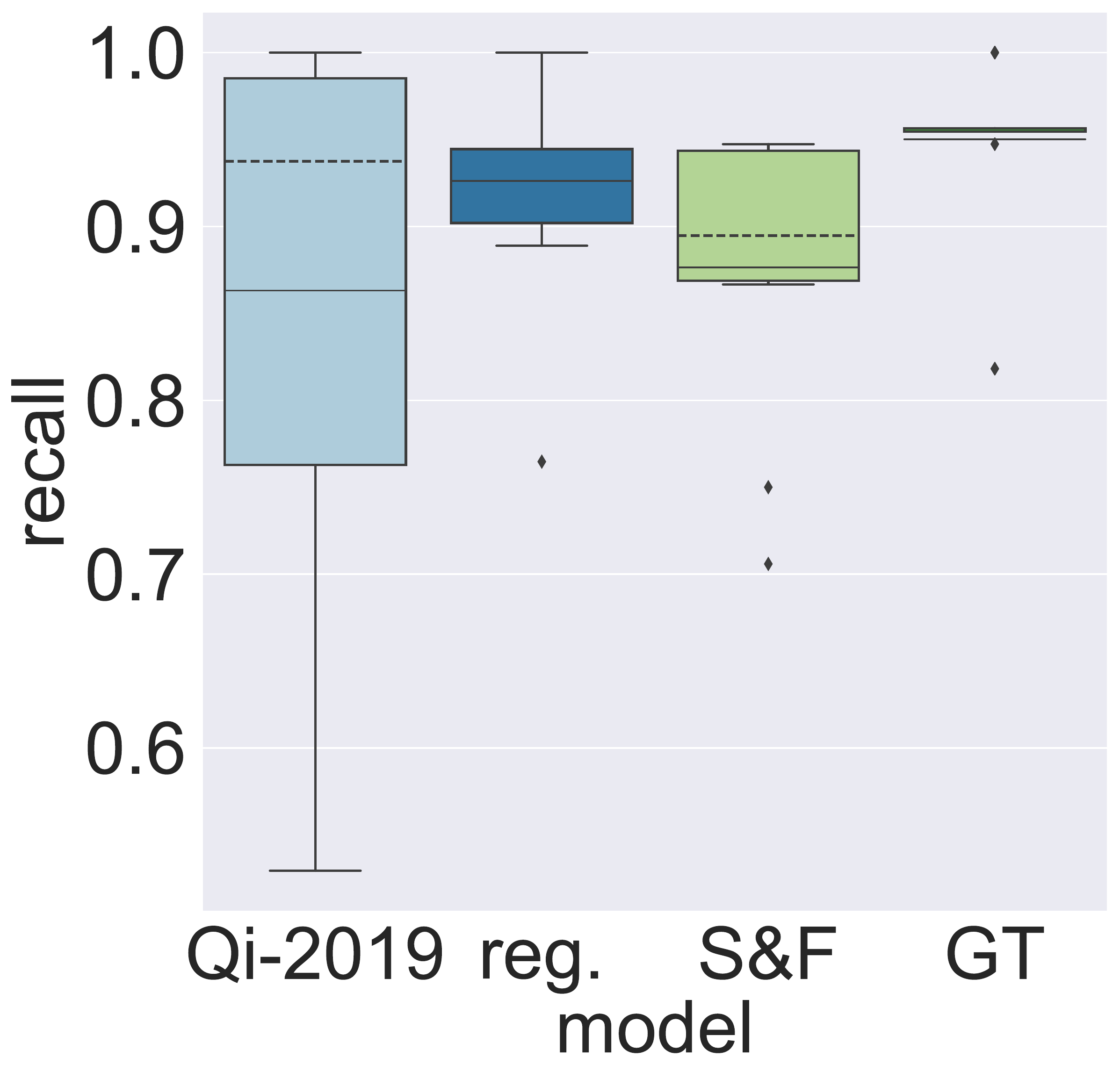}
        \caption{Recall of user ratings.}
    \end{subfigure}%
    \begin{subfigure}[t]{.25\textwidth}
    \centering
    \includegraphics[width=\textwidth]{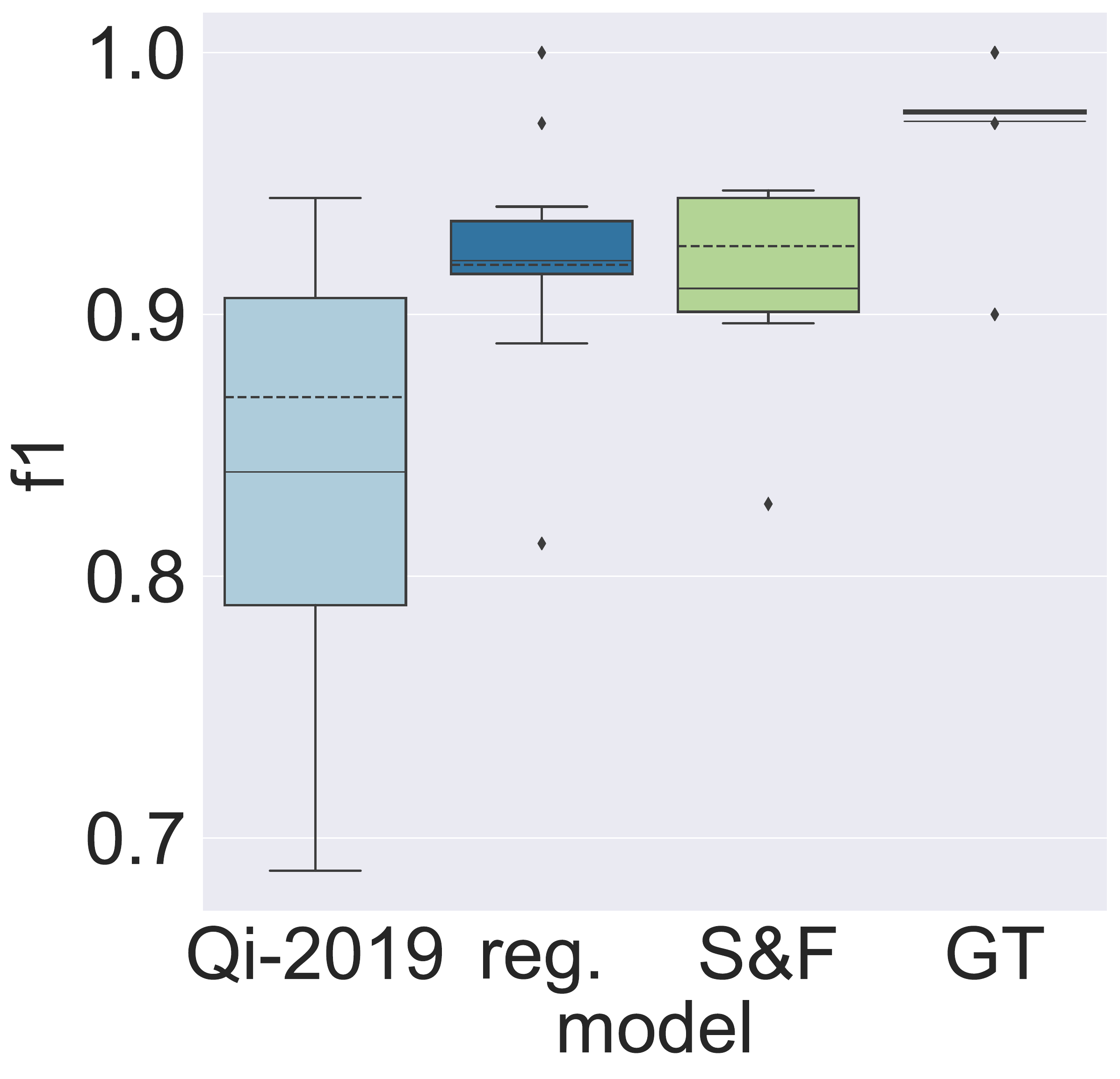}
        \caption{\F\ of user ratings.}
    \end{subfigure}%
    \caption{Boxplots for all continuous dependent variables. Boxes mark quartiles, whiskers mark 1.5 inter-quartile ranges, outliers are plotted separately. Vertical solid lines within boxes mark means, vertical dashed lines mark medians.}\label{fig:all_boxplots}
\end{figure*}

\begin{figure*}
    \centering
    \begin{subfigure}[t]{.33\textwidth}
    \centering
    \includegraphics[width=\textwidth]{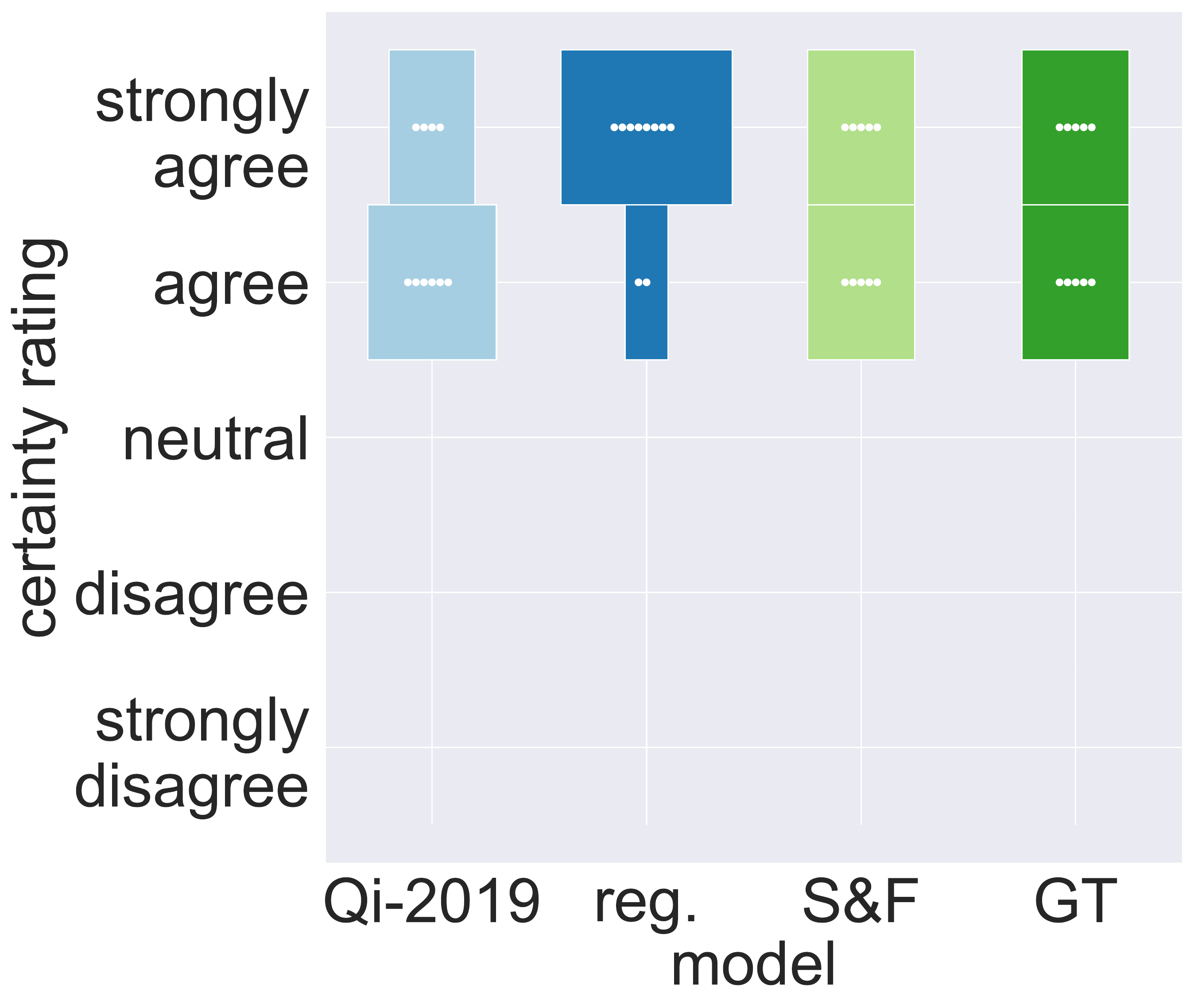}
    \caption{Certainty.}
    \end{subfigure}%
    \begin{subfigure}[t]{.33\textwidth}
    \centering
    \includegraphics[width=\textwidth]{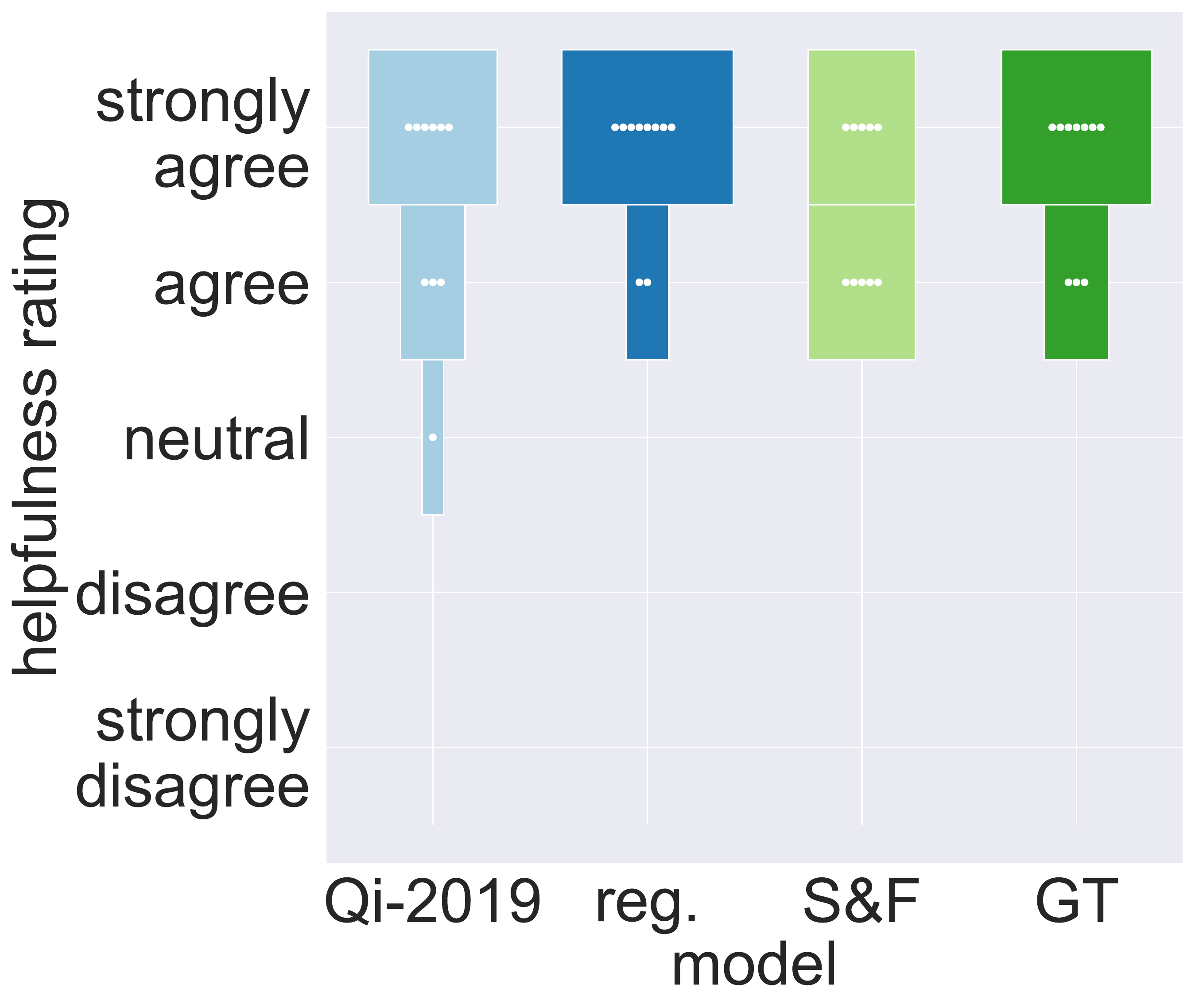}
    \caption{Helpfulness.}
    \end{subfigure}%
    \begin{subfigure}[t]{.33\textwidth}
    \centering
    \includegraphics[width=\textwidth]{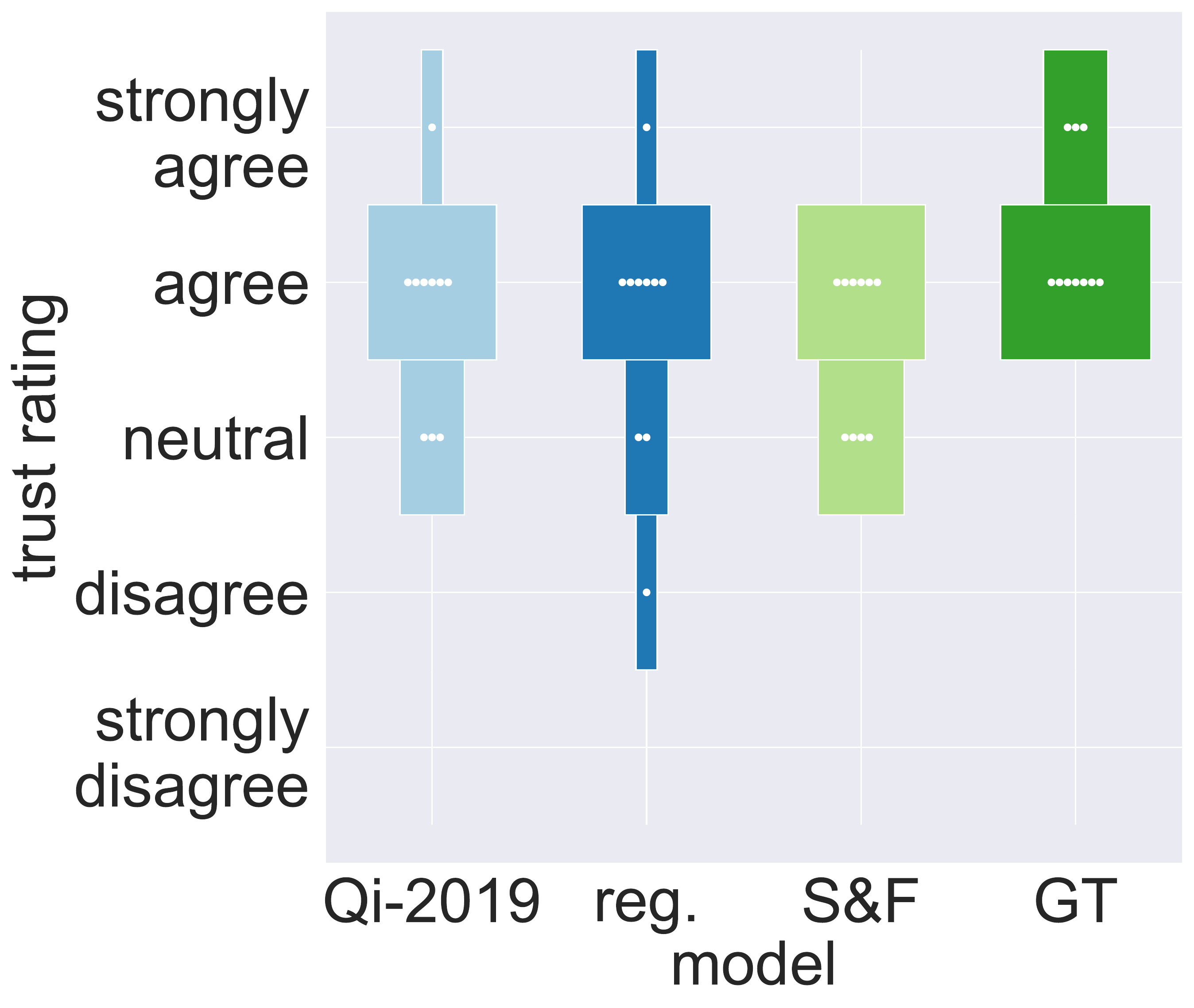}
    \caption{Trust.}
    \end{subfigure}
    ~
    \begin{subfigure}[t]{.33\textwidth}
    \centering
    \includegraphics[width=\textwidth]{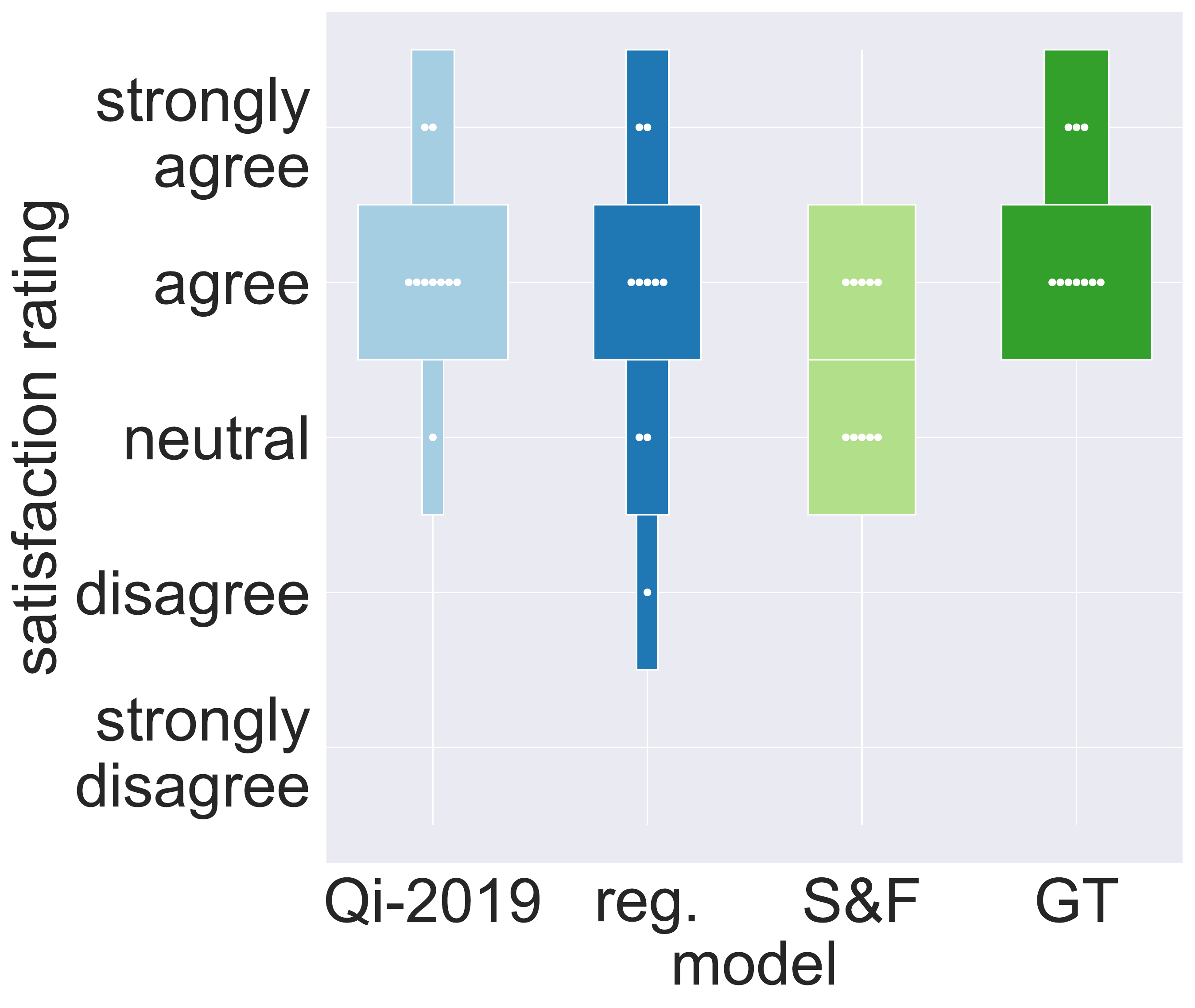}
    \caption{Satisfaction.}
    \end{subfigure}%
    \begin{subfigure}[t]{.33\textwidth}
    \centering
    \includegraphics[width=\textwidth]{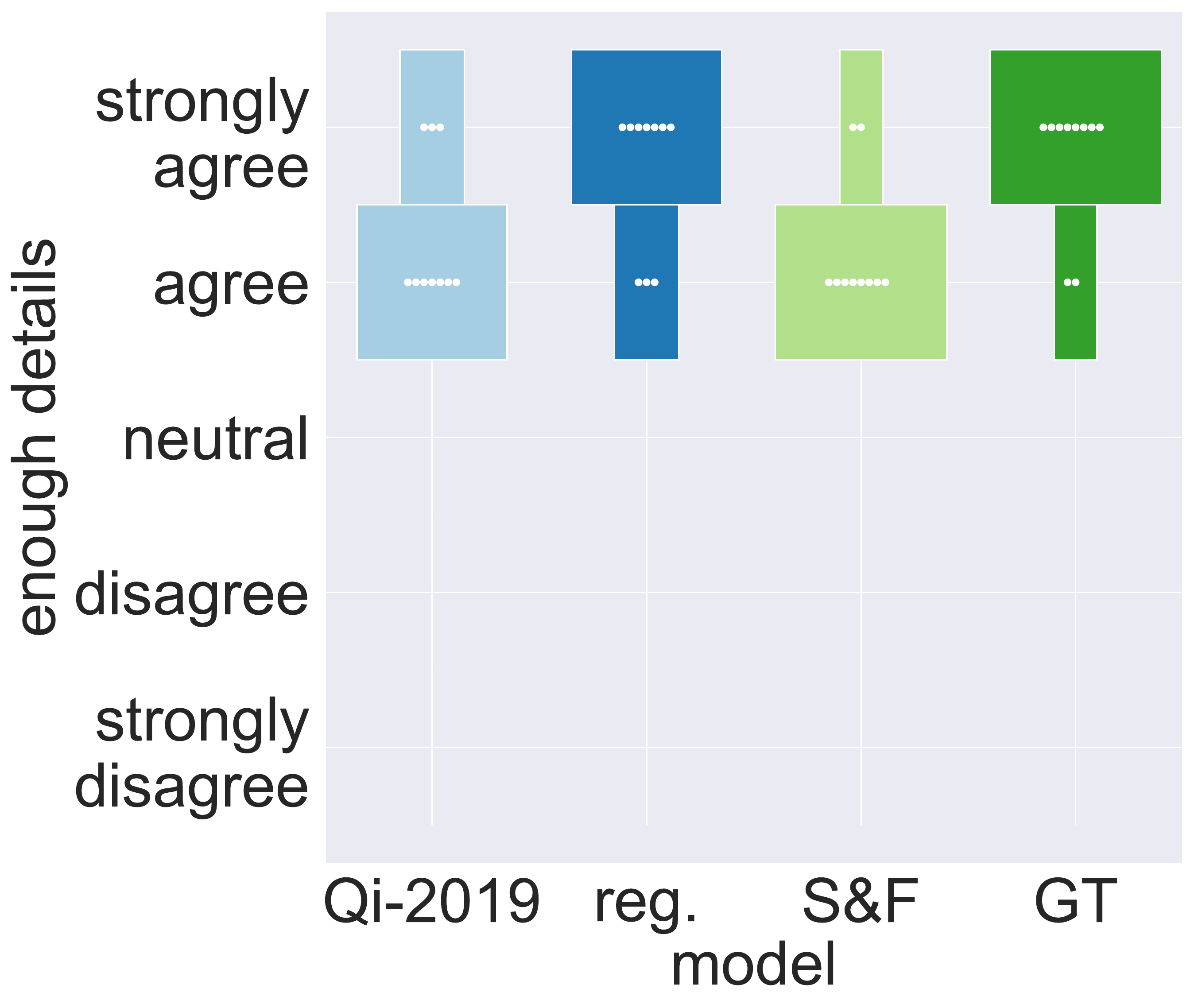}
    \caption{Enough details.}
    \end{subfigure}%
    \begin{subfigure}[t]{.33\textwidth}
    \centering
    \includegraphics[width=\textwidth]{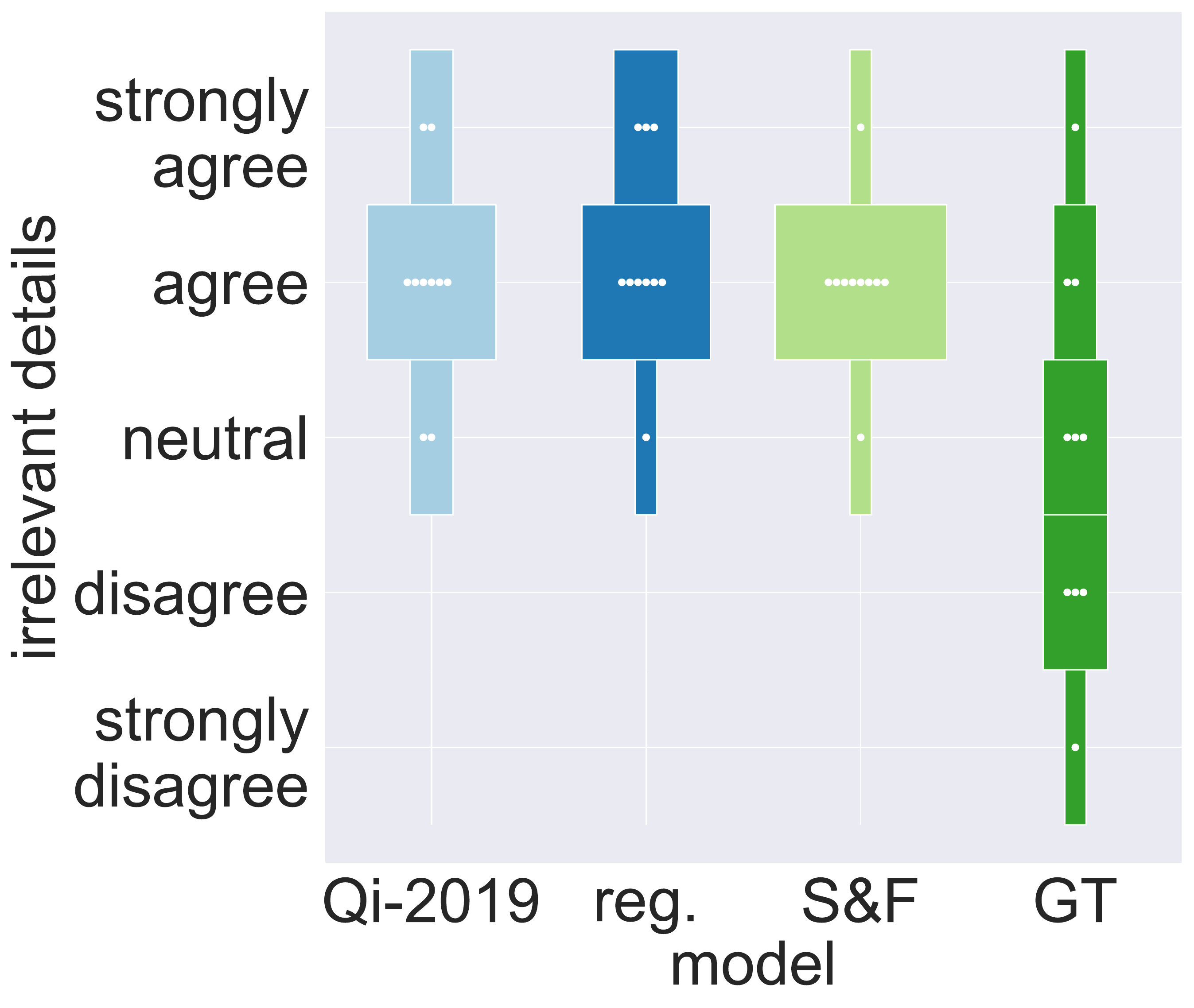}
    \caption{Irrelevant details.}
    \end{subfigure}%
    \caption{Distribution of Likert scale ratings. White dots mark number of participants, bar widths correspond to normalized frequency counts. Certainty and helpfulness ratings are aggregated per participant using the participant ratings' mode.}\label{fig:all_distribution}
\end{figure*}

\begin{figure*}
	    \centering
	    \F-scores\\
	    \includegraphics[width=.33\textwidth]{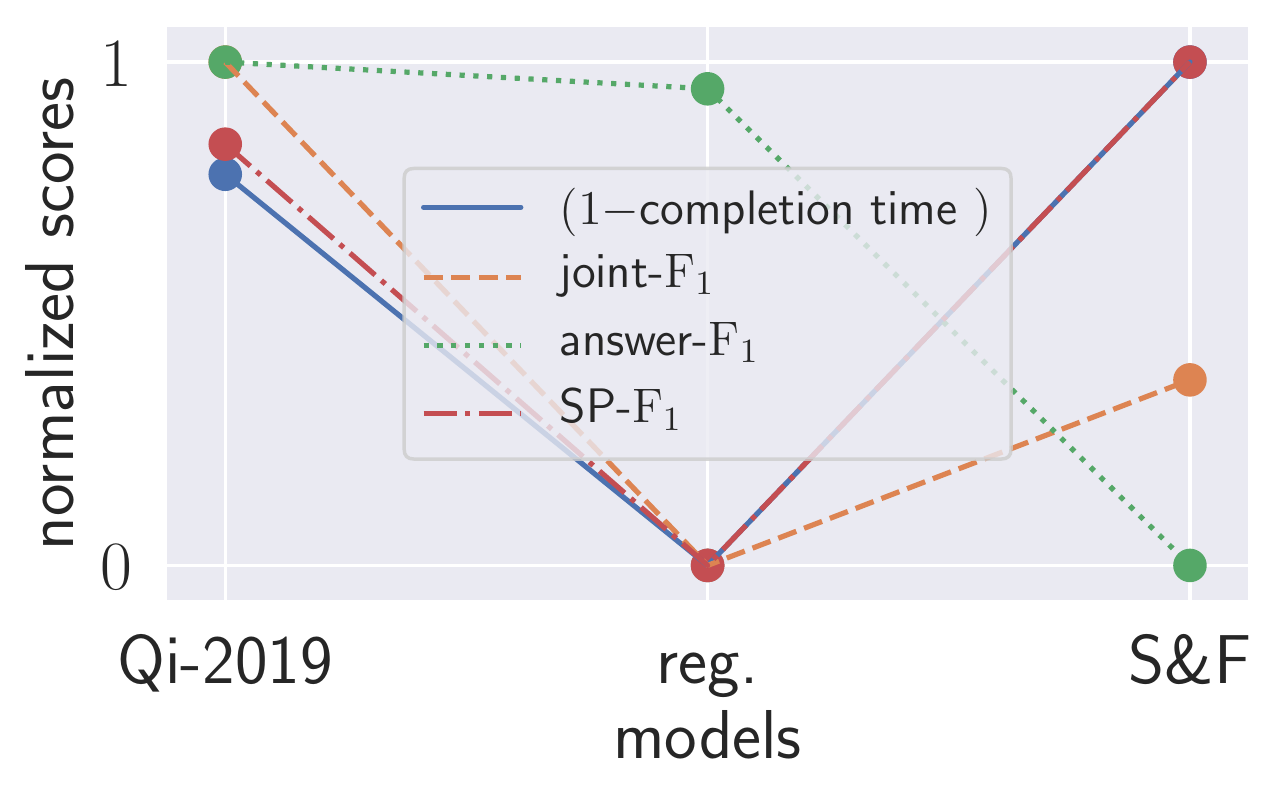}%
        \includegraphics[width=.33\textwidth]{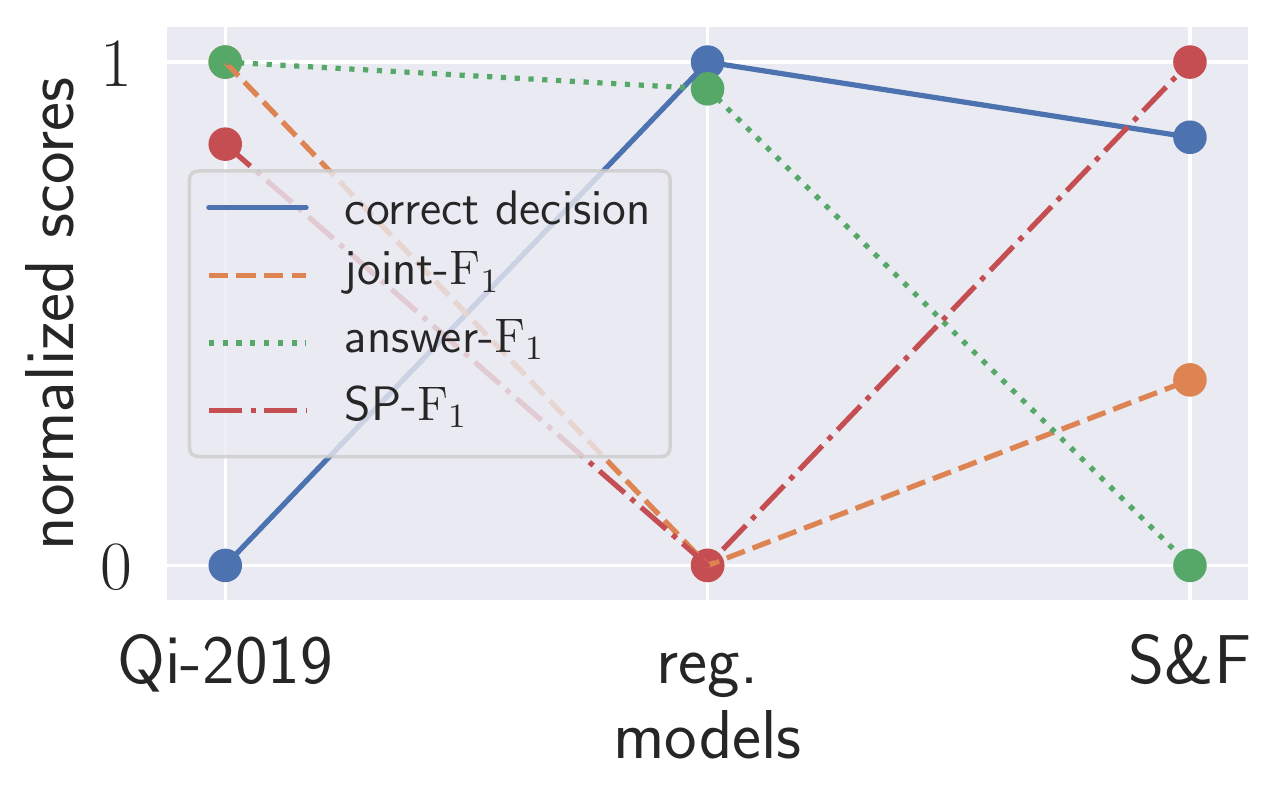}%
        \includegraphics[width=.33\textwidth]{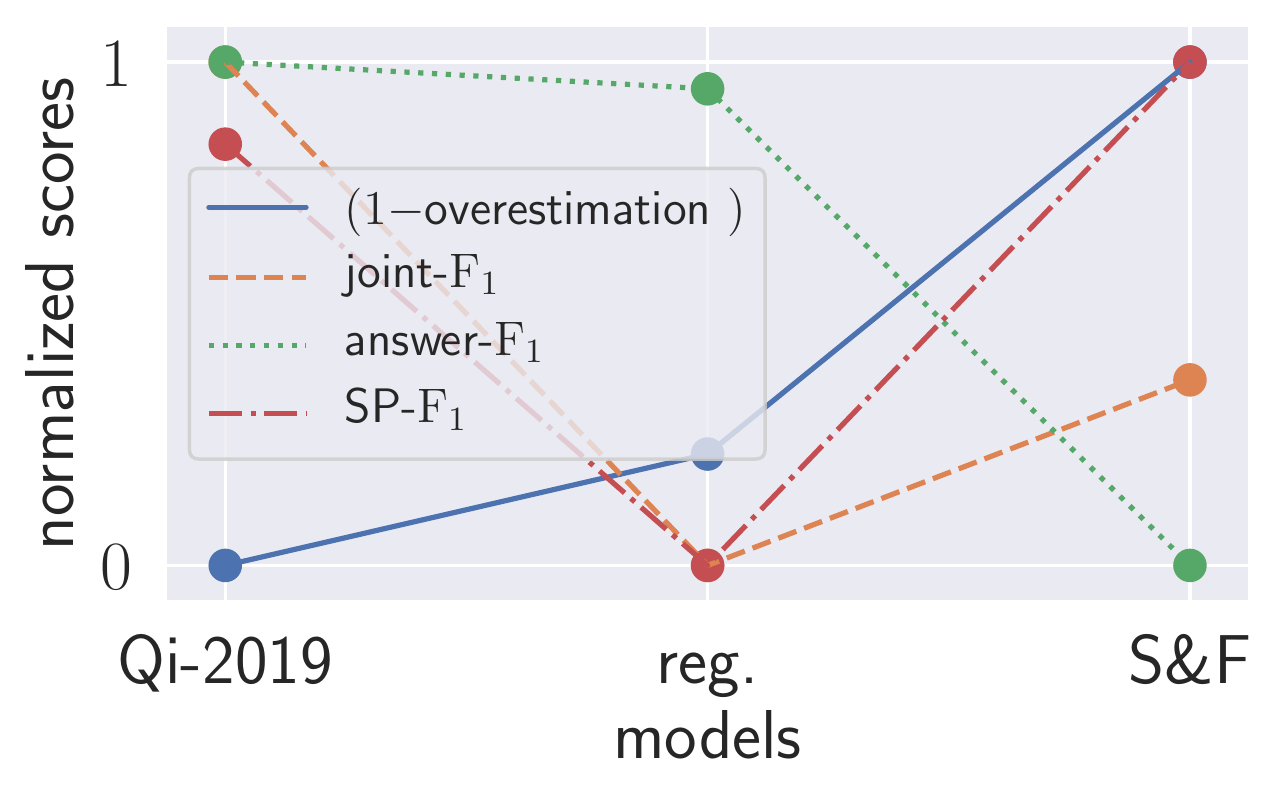}
        $\textsc{FaRM}(4)$ and \textsc{LocA} scores\\
        \includegraphics[width=.33\textwidth]{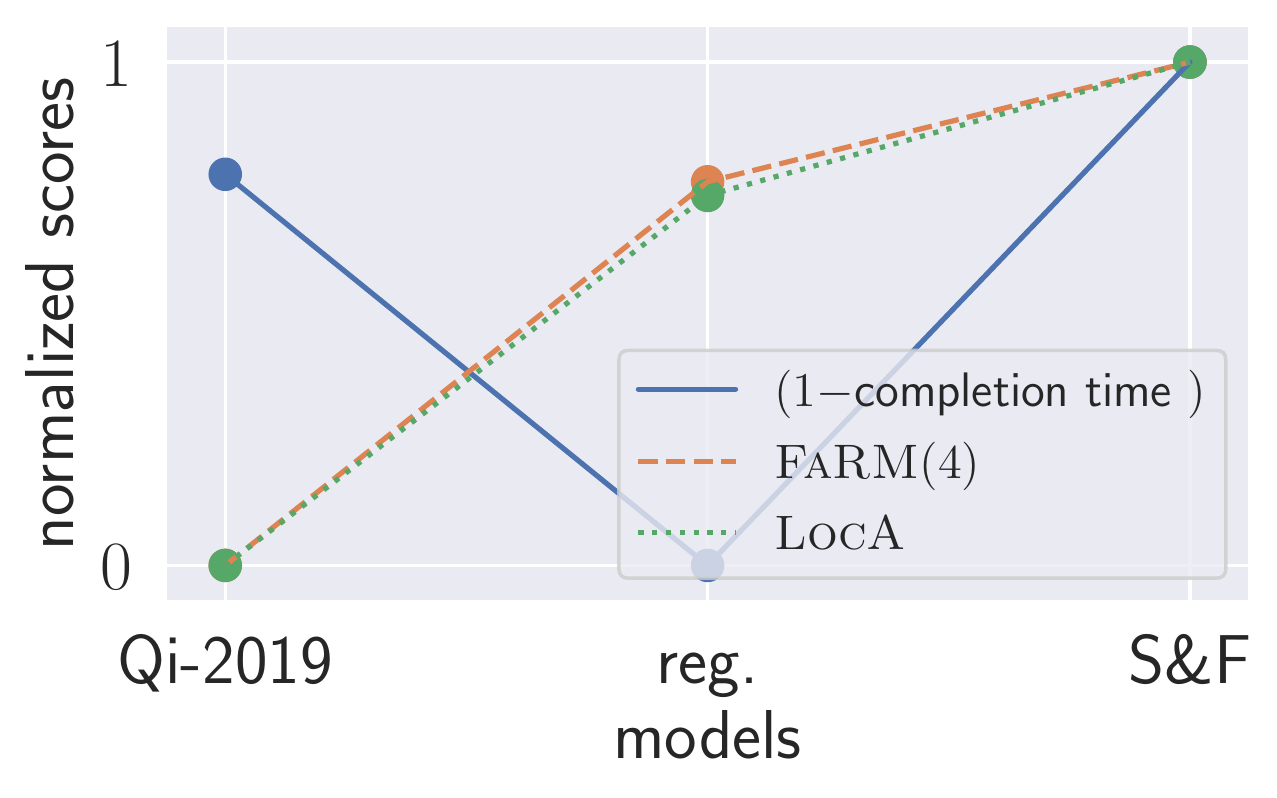}%
        \includegraphics[width=.33\textwidth]{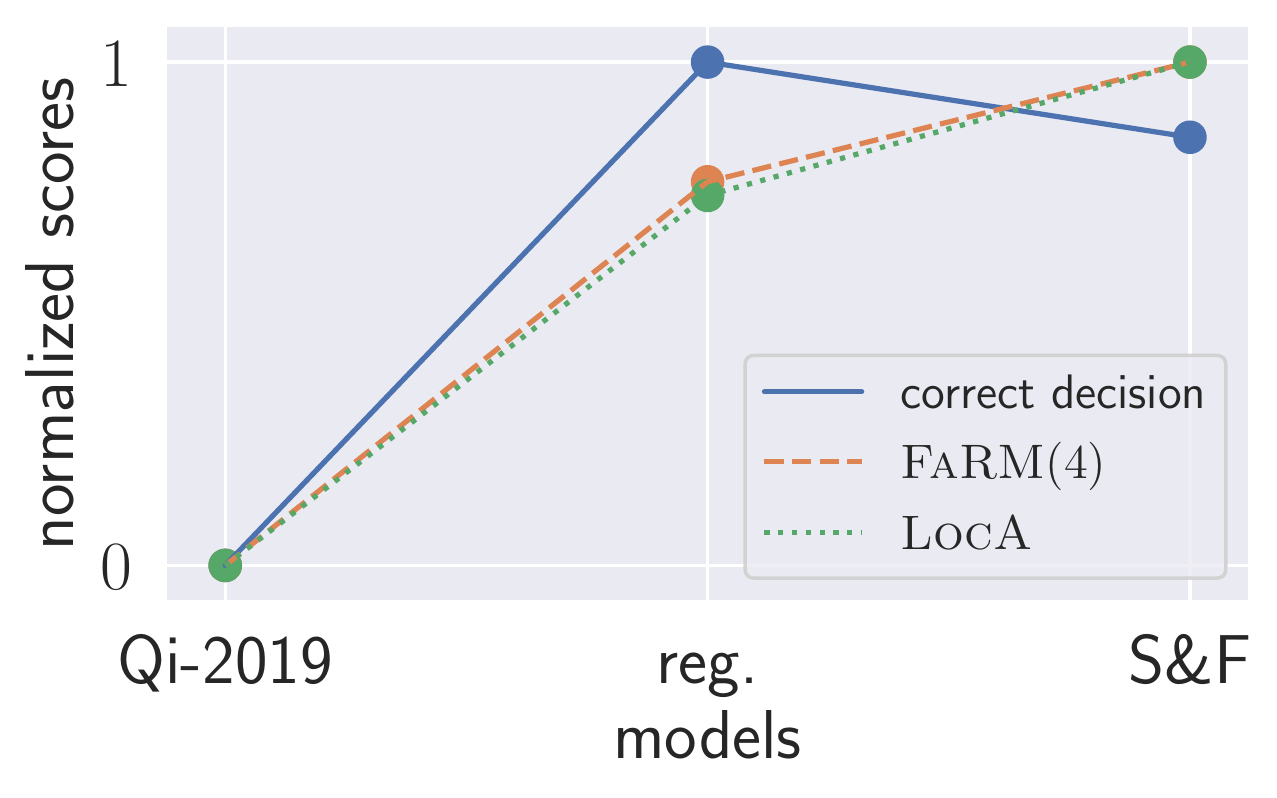}%
        \includegraphics[width=.33\textwidth]{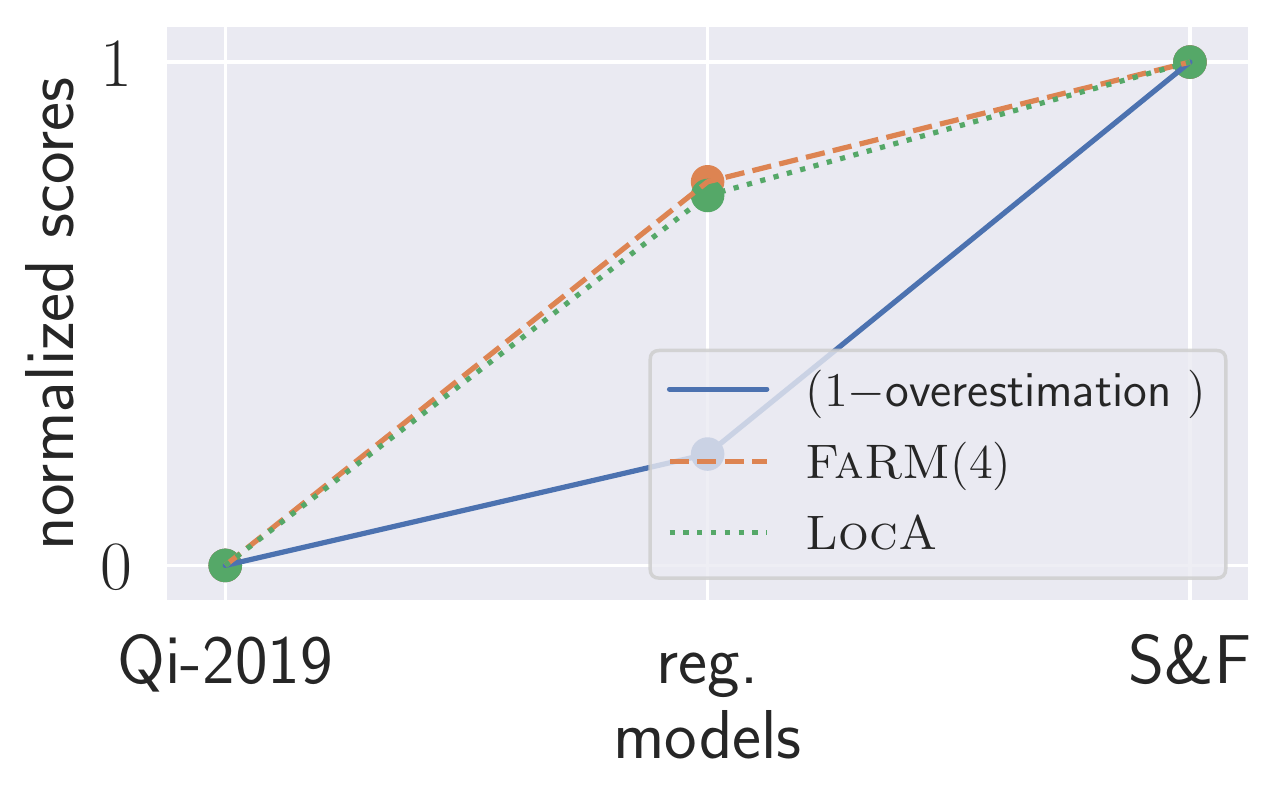}%
        \hrule
        ~\\
        \F-scores\\
        \includegraphics[width=.33\textwidth]{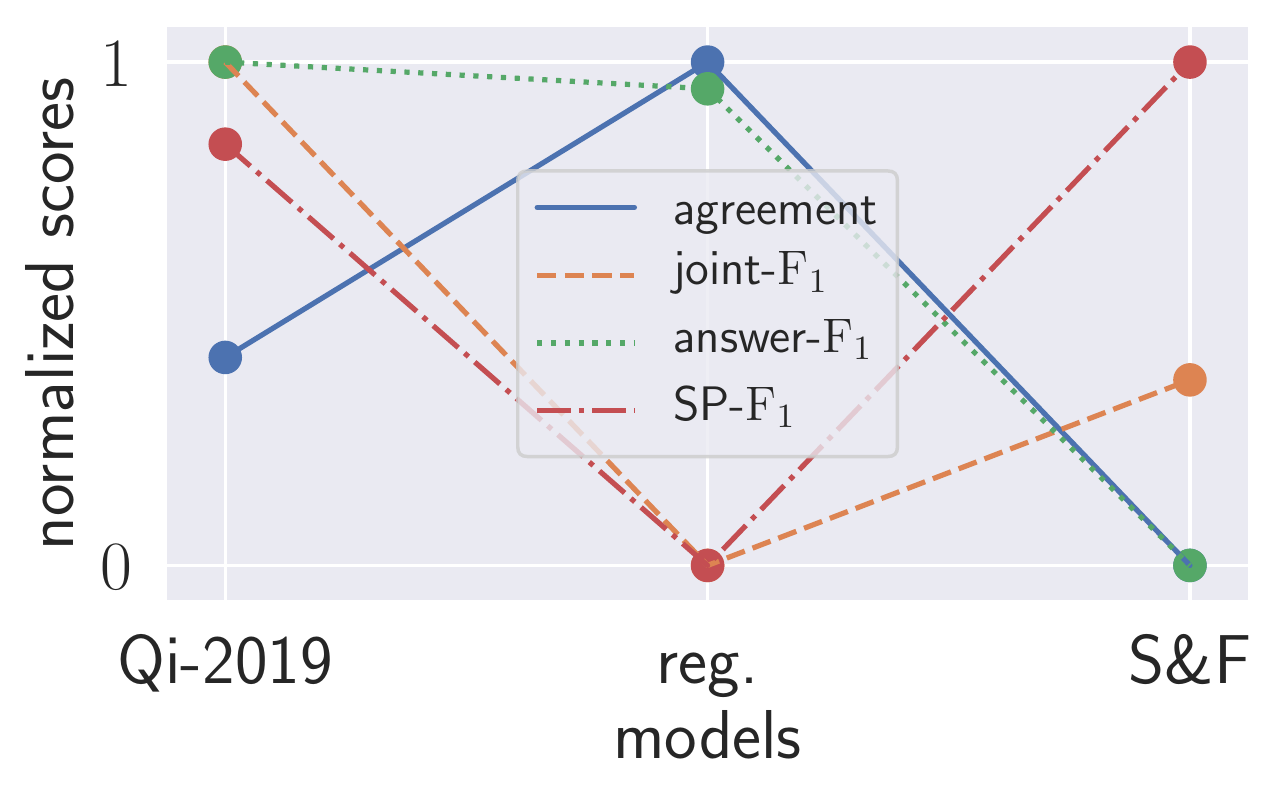}%
        \includegraphics[width=.33\textwidth]{figures/FP_f1.pdf}%
        \includegraphics[width=.33\textwidth]{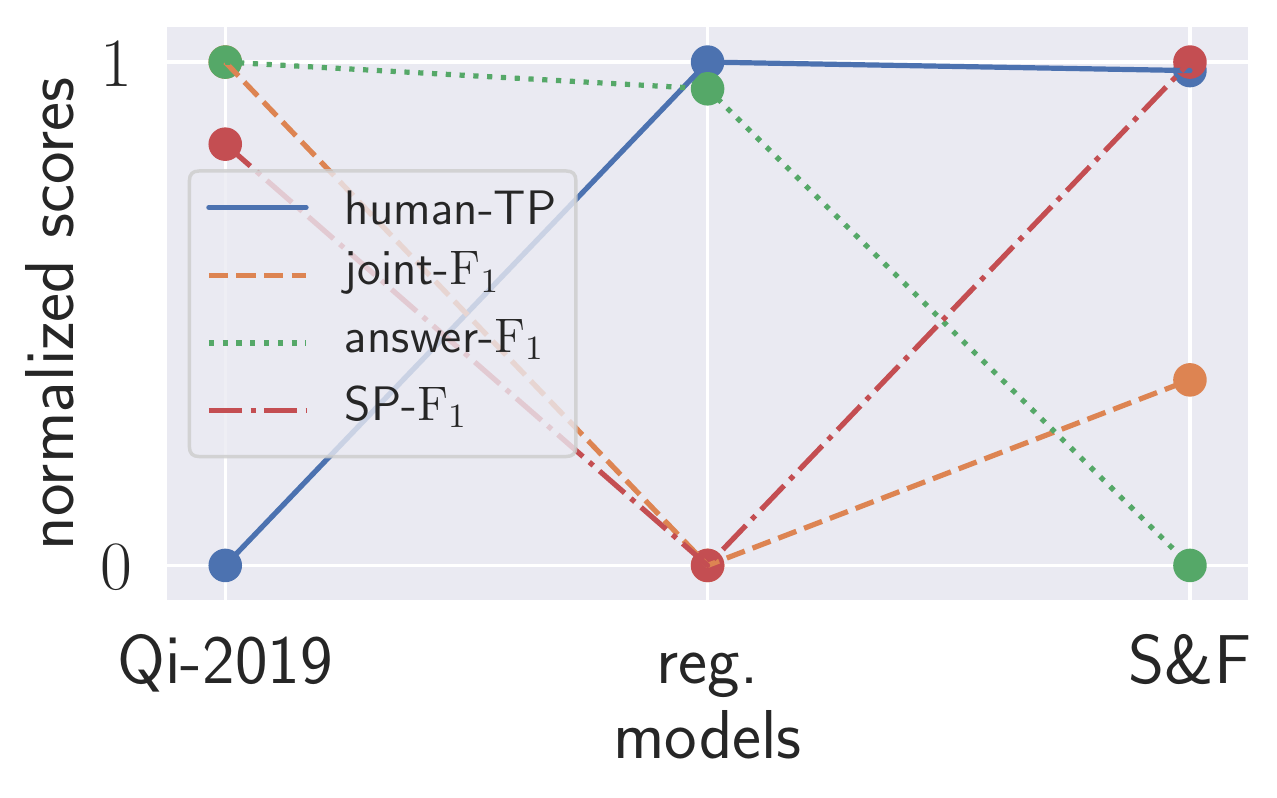}
        $\textsc{FaRM}(4)$ and \textsc{LocA} scores\\
        \includegraphics[width=.33\textwidth]{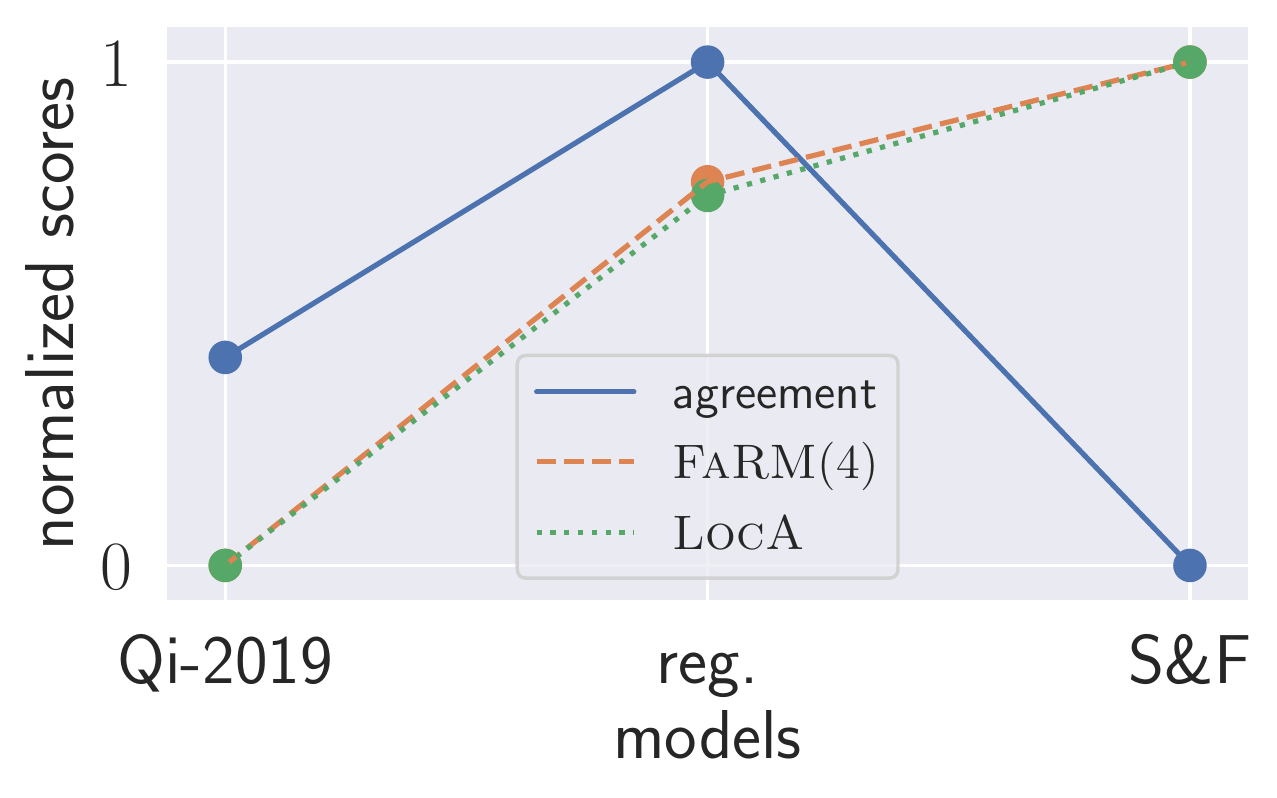}%
        \includegraphics[width=.33\textwidth]{figures/FP_ours.pdf}
        \includegraphics[width=.33\textwidth]{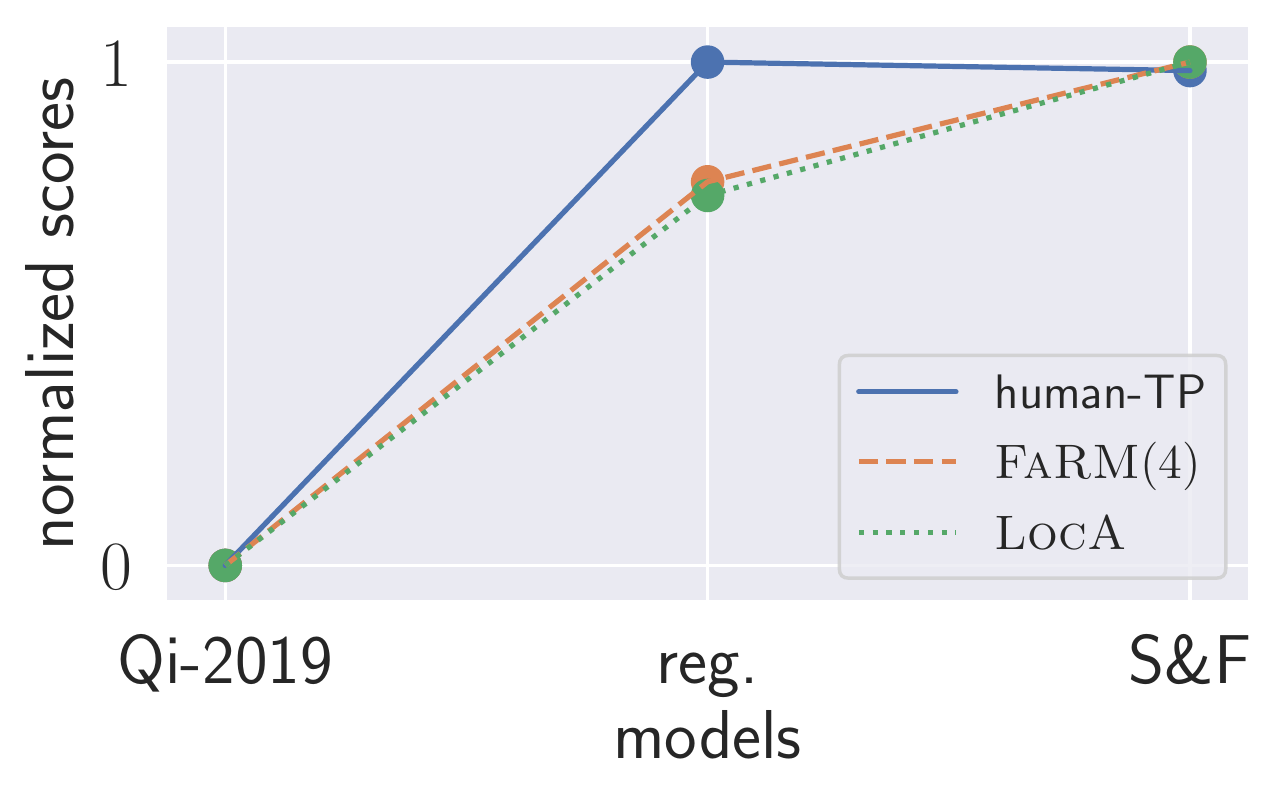}
        \caption{Comparisons between human measures and model scores. All scores are normalized before plotting by subtracting the minimum score and re-scaling the score span to $[0,1]$. Human measures for which lower values correspond to better performance are plotted as $(1-$score$)$ for convenience of the reader. The figure shows scores for completion time, fraction of correct user decisions, overestimation, agreement, false positives and true positives.}\label{fig:contrast_f1_ours_appendix_1}
\end{figure*}

\begin{figure*}
    \centering
    \F-scores\\
        \includegraphics[width=.33\textwidth]{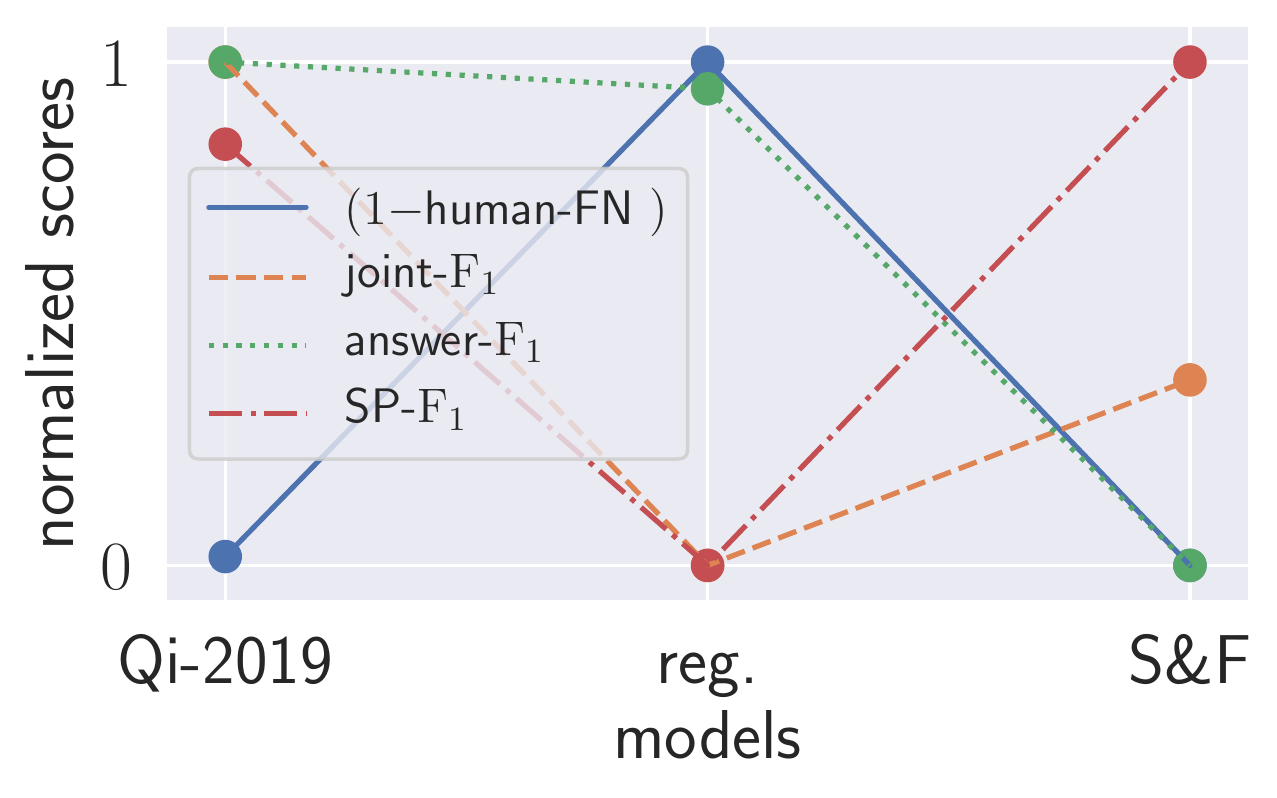}%
        \includegraphics[width=.33\textwidth]{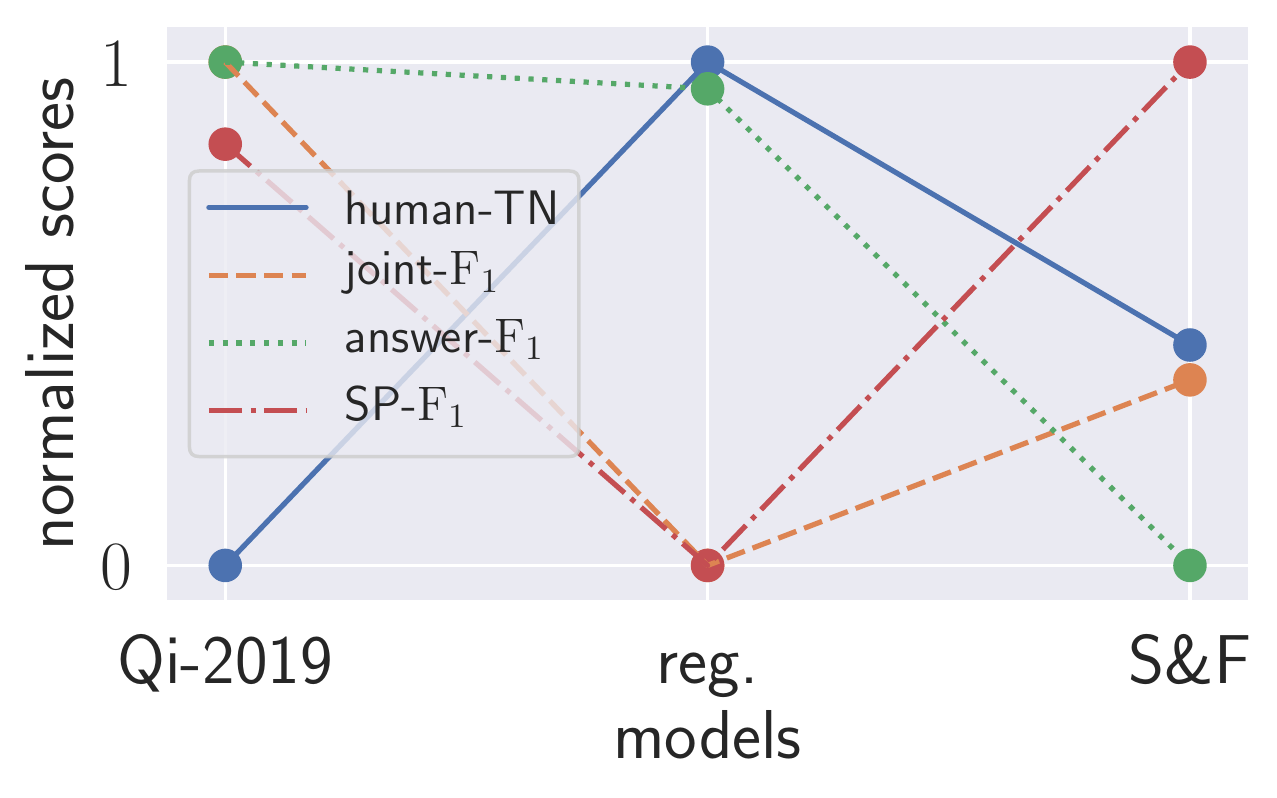}%
        \includegraphics[width=.33\textwidth]{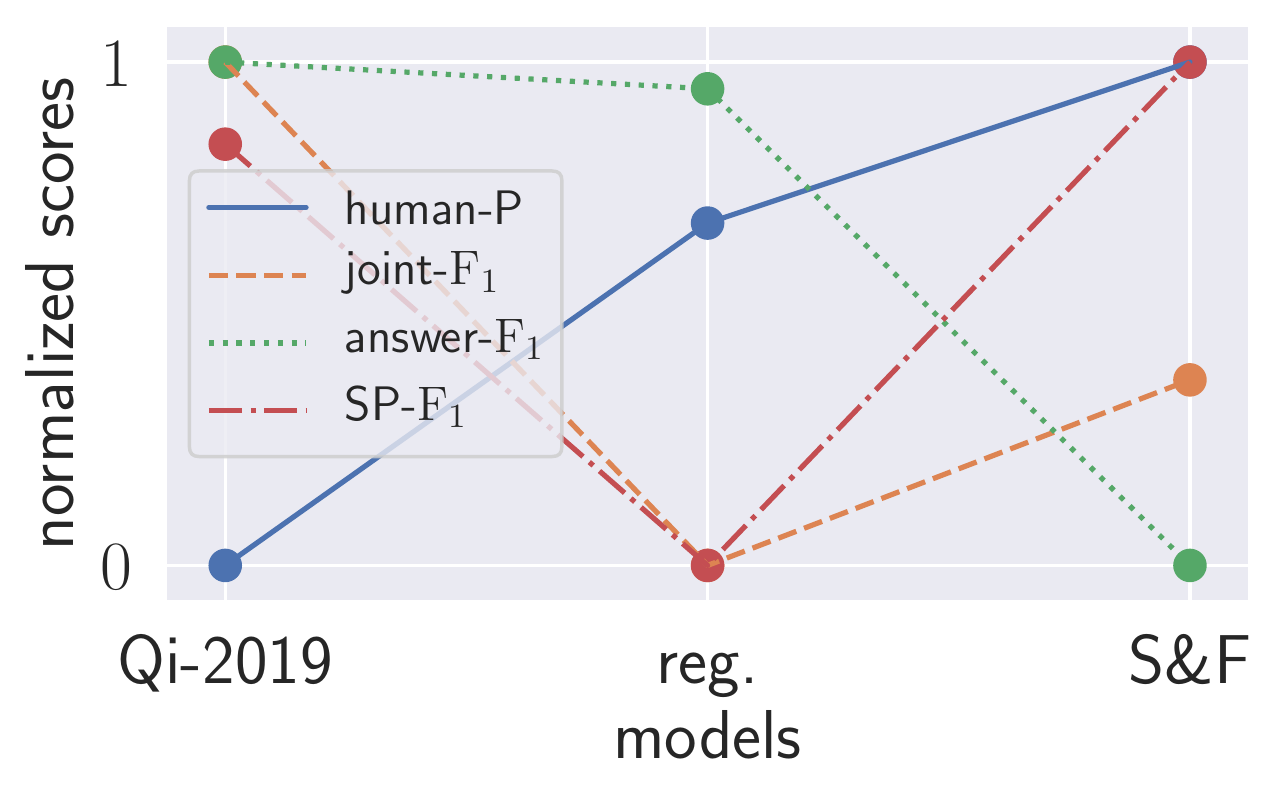}
        $\textsc{FaRM}(4)$ and \textsc{LocA} scores\\
        \includegraphics[width=.33\textwidth]{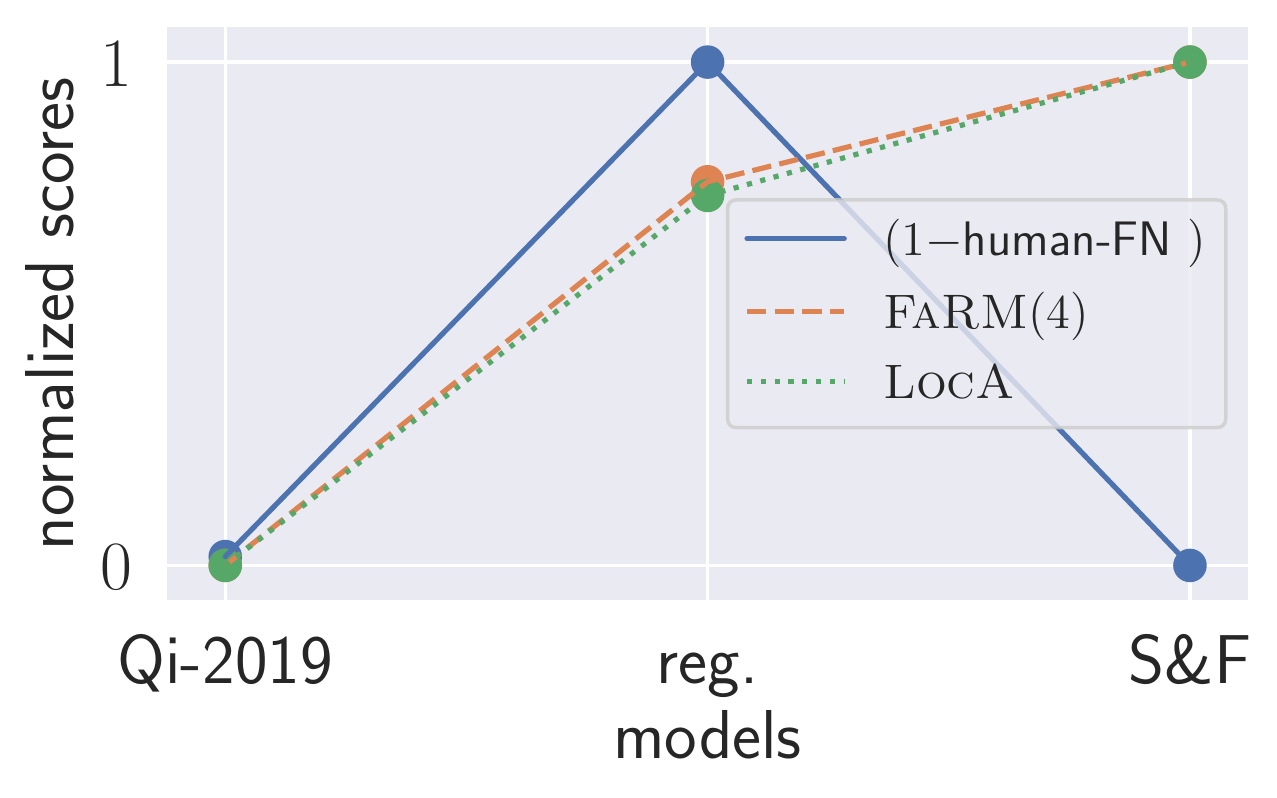}%
        \includegraphics[width=.33\textwidth]{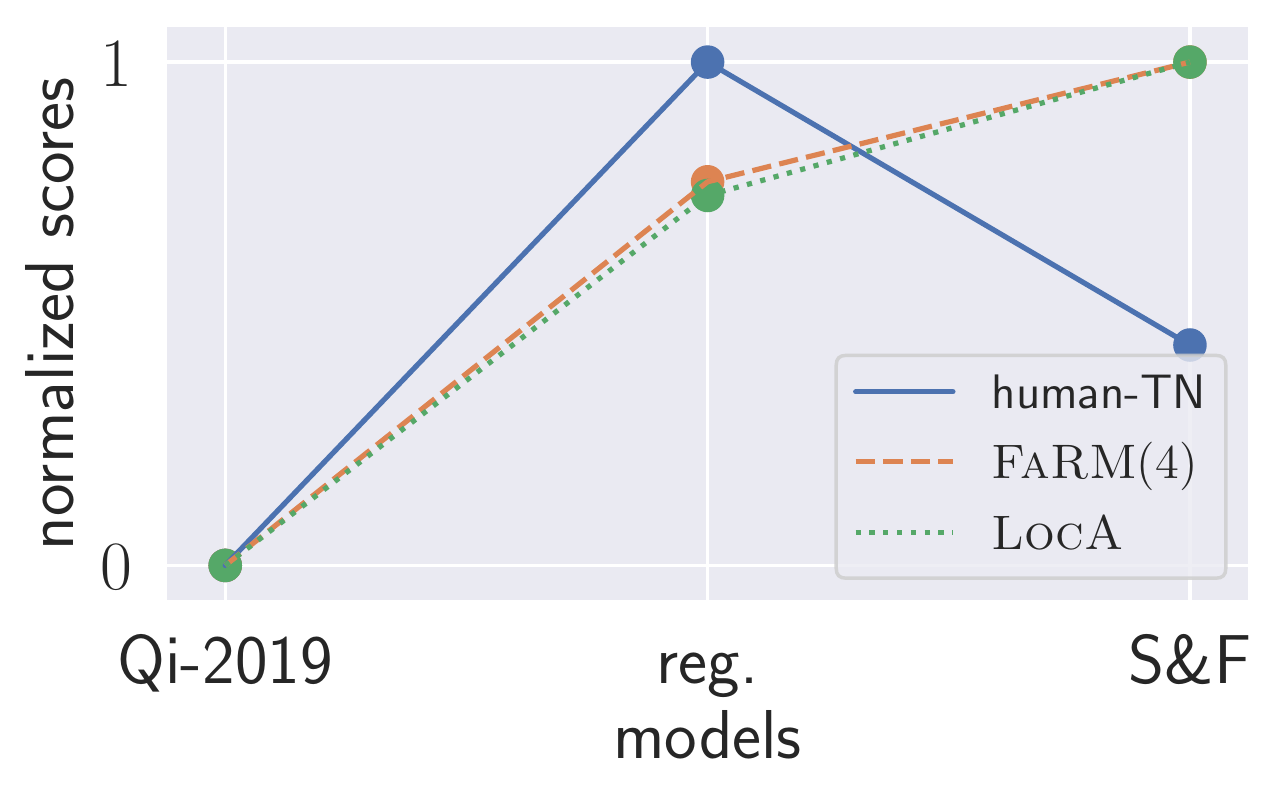}
        \includegraphics[width=.33\textwidth]{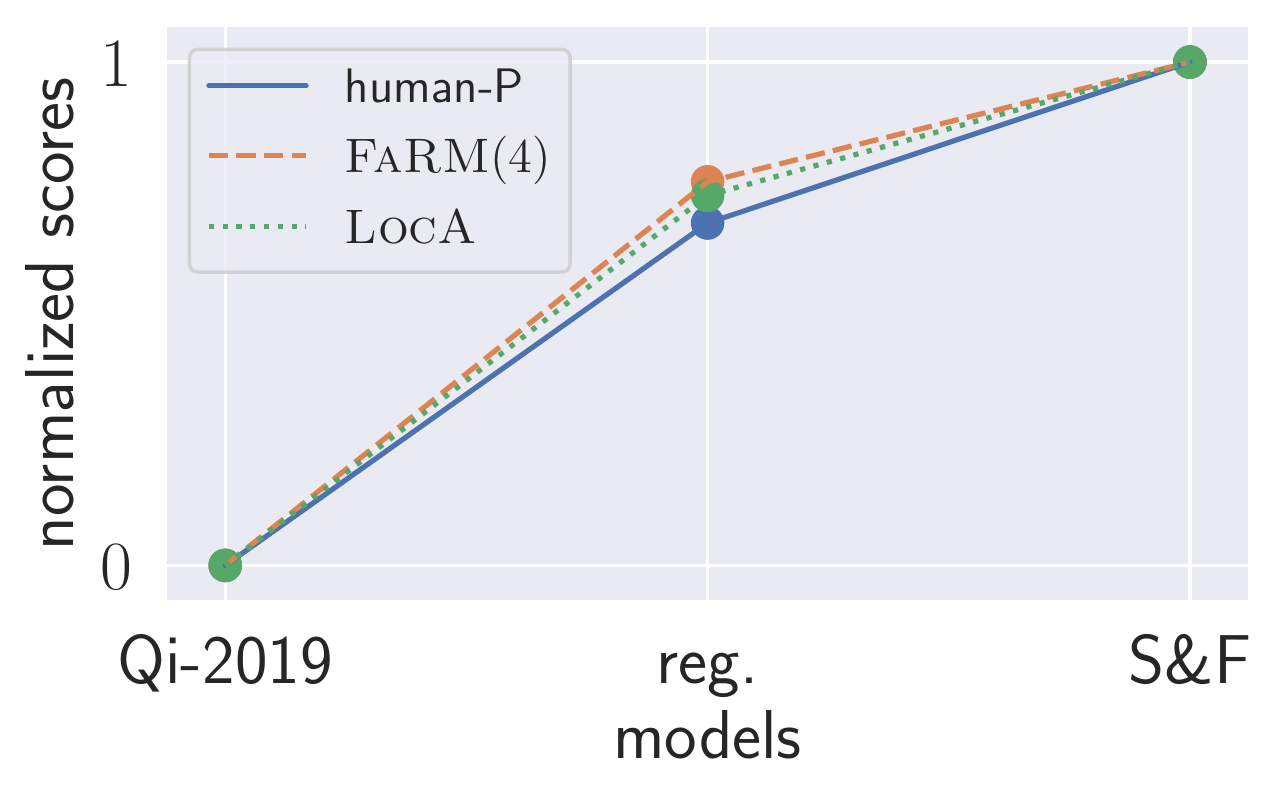}
        \hrule
        ~\\
        \F-scores\\
        \includegraphics[width=.33\textwidth]{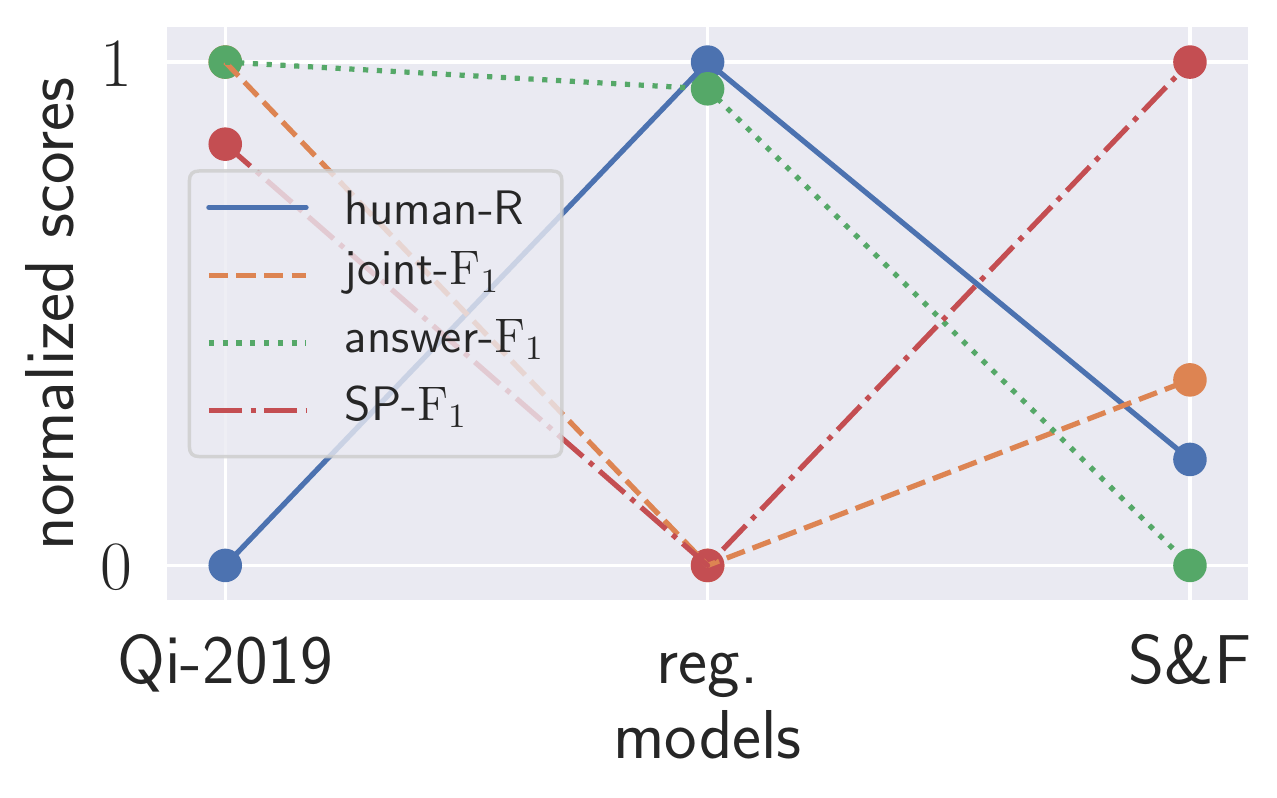}%
        \includegraphics[width=.33\textwidth]{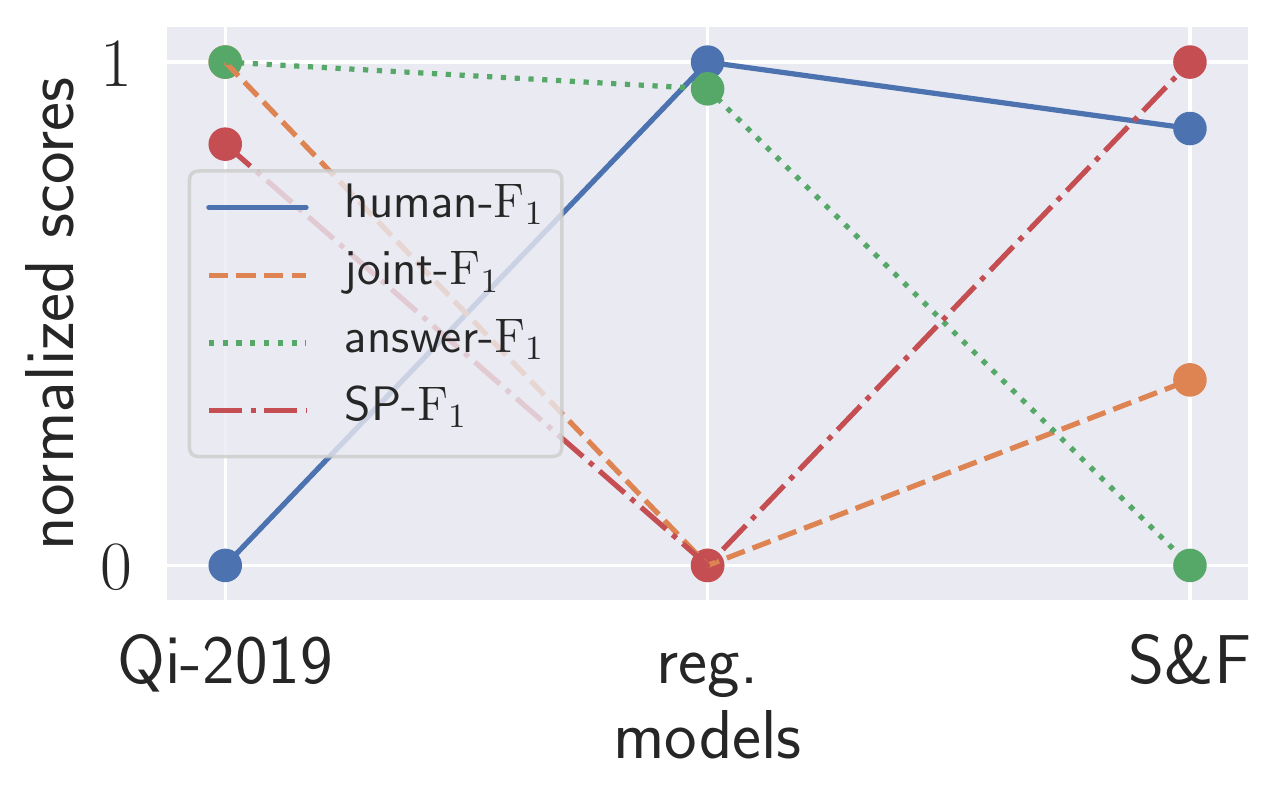}
        ~\\
        $\textsc{FaRM}(4)$ and \textsc{LocA} scores\\
        \includegraphics[width=.33\textwidth]{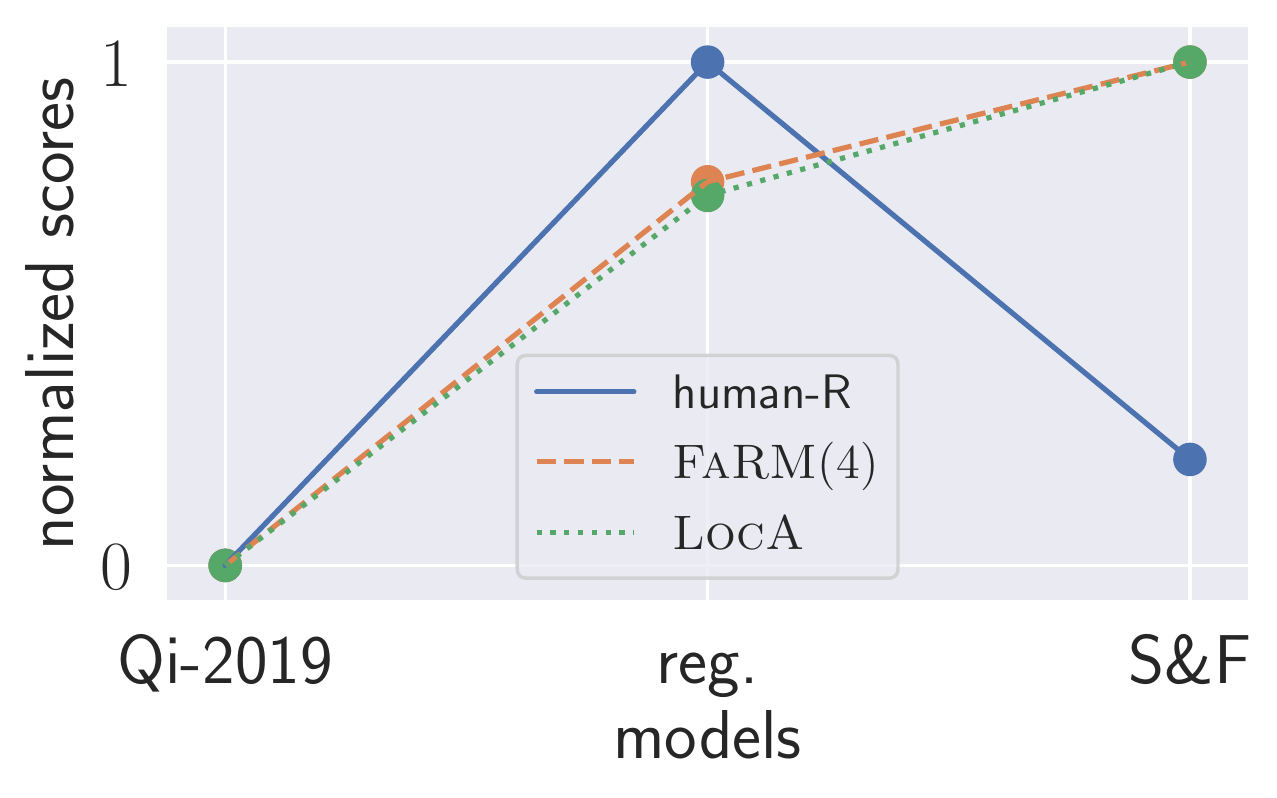}%
        \includegraphics[width=.33\textwidth]{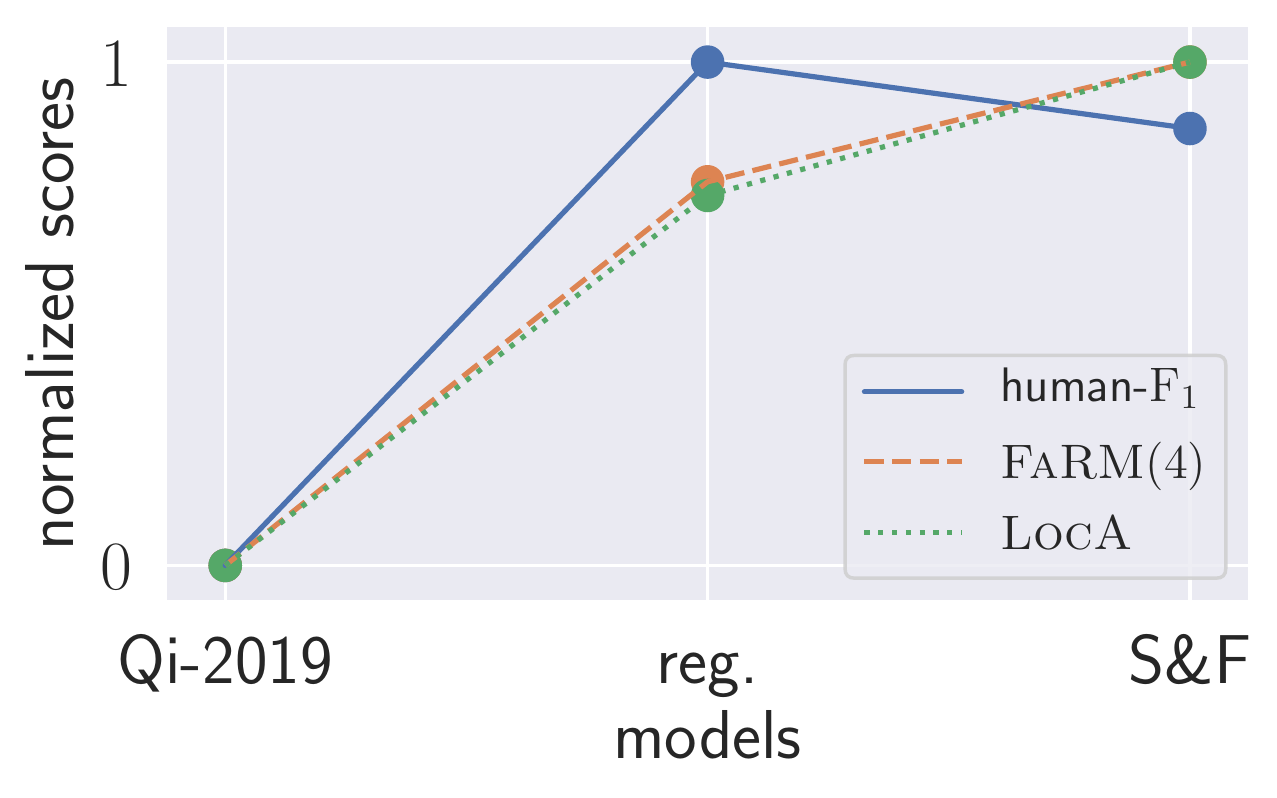}
        \caption{Comparisons between human measures and model scores. All scores are normalized before plotting by subtracting the minimum score and re-scaling the score span to $[0,1]$. Human measures for which lower values correspond to better performance are plotted as $(1-$score$)$ for convenience of the reader. The figure shows scores for false negatives, true negatives, precision, recall and user \F.}\label{fig:contrast_f1_ours_appendix_2}
\end{figure*}

Figure~\ref{fig:all_boxplots} shows boxplots per condition for all continuous dependent variables.
Figure~\ref{fig:all_distribution} shows rating distributions per condition for all ordinal dependent variables.
Figures~\ref{fig:contrast_f1_ours_appendix_1}~and~\ref{fig:contrast_f1_ours_appendix_2} compare model scores and human measures grouped into \F-scores and our proposed $\textsc{FaRM}(4)$ and \textsc{LocA} scores.
Rows alternate between \F-scores and our scores.
Table~\ref{tab:ranks_pearson} shows pairwise Pearson correlation coefficients between human and automatized scores.

\clearpage
\begin{table*}
\centering
\resizebox{\textwidth}{!}{%
\begin{tabular}{l!{\vrule width 1pt}cccccccccccc!{\vrule width 1pt}cccccccc}
\toprule
		& \multicolumn{12}{c}{Standard Scores} &  \multicolumn{7}{c}{Proposed Scores}    \\
		\cmidrule(lr){2-13}
		\cmidrule(lr){14-20} 
		& \multicolumn{4}{c}{Answer} &  \multicolumn{4}{c}{Supporting Facts}  &  \multicolumn{4}{c}{Joint} & \multicolumn{2}{c}{Answer Changes} & \multicolumn{2}{c}{\textsc{FaRM}} & \multicolumn{2}{c}{\%-in-fact} & \textsc{LocA} \\
		\cmidrule(lr){2-5}
		\cmidrule(lr){6-9}
		\cmidrule(lr){10-13}
		\cmidrule(lr){14-15}
		\cmidrule(lr){16-17}
		\cmidrule(lr){18-19}
		\cmidrule(lr){20-20}
Human eval. & EM & \F & P & R & EM & \F & P & R & EM & \F & P & R & rel. & irrel. & $\textsc{FaRM}(1)$ & $\textsc{FaRM}(4)$ & rel. & irrel. & $\textsc{LocA}$\\\midrule 
\rowcolor{lightgray} correct decision  &  -0.33 &  -0.42 &  -0.37 &  -0.47 &  -0.50 &  -0.49 &  -0.57 & 0.63 &  -0.57 & -0.97 & -0.79 &  0.30 &  0.80 & -0.93 & 0.99 &  0.93 & 1.00 &  -0.76 &  0.92\\ \midrule 
overestimation  &  0.97 & 0.99 &  0.98 & 0.99 &  -0.43 &  -0.45 &  -0.36 &  0.29 &  -0.36 &  0.35 &  -0.07 &  0.62 & 0.96 &  -0.74 &  0.98 & 1.00 &  0.93 & -0.94 & 1.00\\ \midrule 
\rowcolor{lightgray} completion time  & 0.71 &  0.64 & 0.67 &  0.59 & -1.00 & -1.00 & -0.99 &  0.97 & -0.99 &  -0.63 &  -0.90 & 0.99 &  0.23 &  -0.95 &  0.67 &  0.49 &  0.79 &  -0.17 &  0.46\\ \midrule 
human-FP  &  0.70 & 0.77 &  0.73 & 0.80 &  0.09 &  0.07 &  0.16 &  -0.24 &  0.17 &  0.78 &  0.45 &  0.14 & 0.97 &  -0.30 &  0.73 & 0.87 &  0.61 & -0.98 & 0.88\\ \midrule 
\rowcolor{lightgray} human-TP  &  -0.45 &  -0.53 &  -0.49 &  -0.58 &  -0.39 &  -0.38 &  -0.46 & 0.53 &  -0.47 & -0.94 & -0.70 &  0.18 &  0.87 & -0.87 & 1.00 &  0.97 & 0.99 &  -0.84 &  0.96\\ \midrule 
human-FN  & -0.55 &  -0.47 & -0.51 &  -0.42 & 0.99 & 0.99 & 1.00 &  -1.00 & 1.00 &  0.77 &  0.97 & -0.94 &  -0.03 &  0.87 &  -0.51 &  -0.30 &  -0.65 &  -0.03 &  -0.28\\ \midrule 
\rowcolor{lightgray} human-TN  &  0.12 &  0.02 &  0.07 &  -0.04 &  -0.83 &  -0.82 &  -0.87 & 0.91 &  -0.87 & -0.98 & -0.98 &  0.69 &  0.45 & -1.00 & 0.83 &  0.68 & 0.91 &  -0.40 &  0.66\\ \midrule 
human-P  &  -0.71 & -0.78 &  -0.75 & -0.81 &  -0.07 &  -0.05 &  -0.14 &  0.22 &  -0.15 &  -0.77 &  -0.43 &  -0.15 & 0.98 &  -0.66 &  0.95 & 1.00 &  0.88 & -0.97 & 1.00\\ \midrule 
\rowcolor{lightgray} human-R  &  0.36 &  0.27 &  0.32 &  0.21 &  -0.94 &  -0.94 &  -0.97 & 0.98 &  -0.97 & -0.89 & -1.00 &  0.85 &  0.22 & -0.95 & 0.66 &  0.48 & 0.78 &  -0.16 &  0.45\\ \midrule 
human-\F  &  -0.35 &  -0.43 &  -0.39 &  -0.49 &  -0.49 &  -0.47 &  -0.55 & 0.62 &  -0.56 & -0.97 & -0.78 &  0.28 &  0.81 & -0.92 & 0.99 &  0.94 & 1.00 &  -0.77 &  0.93\\ \midrule 
\end{tabular}
}
\caption{Pearson correlations between human and automatized scores.}\label{tab:ranks_pearson}
\end{table*}

\end{document}